%% file: main.tex
\documentclass{clv2025}
\jvol{vv}
\jnum{nn}
\jyear{2026}

\dochead{Long Paper} 

\pageonefooter{This is a pre-print under review by Computational Linguistics}

\usepackage{amsmath}
\usepackage{booktabs}
\usepackage[super]{nth}
\usepackage{physics} 
\usepackage{siunitx} 
\usepackage{csquotes} 
\usepackage{pdflscape} 
\usepackage{wasysym} 
\usepackage{soul} 

\runningtitle{Predicting Compositionality Over Time}
\runningauthor{Jenkins, Raimundo Schulz, Miletić, Schulte im Walde}

\begin{document}

\title{Losing My Composure: Predicting Compositionality Over Time}

\author{Chris Jenkins\thanks{Corresponding Author},
Emma Raimundo Schulz,
Filip Miletić,\\
Sabine Schulte im Walde}

\affilblock{
    \affil{University of Stuttgart\\\quad         \email{christopher.jenkins@ims.uni-stuttgart.de}}
}

\maketitle

\begin{abstract}

We explore the phenomenon of semantic change of German and English noun compounds, with the objective of investigating and modeling gradual changes of meanings and degrees of compositionality in the past and over time.
To do so, we introduce the Compositionality Trend Prediction task, which is evaluated against a novel dataset of in-context compositionality ratings sampled across several decades of diachronic corpora for 23 German and 26 English target compounds, uniquely providing per-decade ratings and corresponding trends over time.
These per-decade compositionality ratings allow us to investigate empirically untested hypotheses of generalized trends in compositionality over time, such as the idea that compounds should become less compositional (less transparent) over time. 
Beyond our empirical observations from the diachronic compositionality annotations, we perform experiments with semantic vector representations of varying complexity, as well as several temporal granularities for training these representations on diachronic data, resulting in about 100 models of each representation type, each covering a different 1--5 decade slice of a diachronic corpus.
Contrary to the decisive tendency posited in the literature, we find only a small negative trend in compositionality over time in our target compounds.
In our computational experiments, we find that using models trained on narrow time slices of diachronic data (single decades, or incrementally expanding temporal windows) align better with the per-decade compositionality ratings than those trained on an entire half-century window, the latter setting being an analog for the prevalent modeling approach of training representations on an entire half of a corpus' data.
Additionally, we find static representations to be competitive with contextual representations in the Compositionality Trend Prediction task.
\end{abstract}

\section{Motivation}\label{ch8:motivation}
Every word is, so to speak, on its own journey through the ages, and compounds have some company along the way:
\textit{Kaffeehaus} (coffee house) can be analyzed in terms of its use as a whole (e.g.
 whether it is synonymous with the French borrowing \textit{café} or as a consequence of how its constituents are used (e.g.\ German \textit{Kaffee} as a plant, a drink, or a meal).
\textit{Ice water} may have been used in differing contexts before, during, and after the era of global trade in naturally-occurring ice ($\approx$\nth{19} century), prior to the invention of electrical refrigeration, while uses of \textit{ice} or \textit{water} may not have changed as much.
\textit{Bosom friend} (a close friend) is less compositional with respect to its modifier \textit{bosom} than it is with respect to its head \textit{friend}. If it was ever used in a more compositional way (e.g.\ a friend that one is constantly hugging), such a meaning is not extant in the diachronic corpus that we examined.
Investigating the semantic change of noun compounds thus takes advantage of the ready-made comparison that each compound invites, as the overall meaning of a compound noun can be interpreted as a single --possibly idiomatic-- unit, or with respect to the meanings of its constituent nouns.
In this vein, we consider noun compounds as a specific instance of multiword expressions while investigating the tension between the potential for semantic change at any point in time, and the stability of many meanings over time~\citep{Blank-1999, Koch:16}.


While some changes have occurred dramatically at a known time, like the signification of \textit{atomic} after 1945\footnote{Having to do with nuclear weapons.}, later divergent uses as evidenced by the 1979 song "Atomic" by Blondie\footnote{Evoking a strong feeling.} are more difficult to attribute to a specific time.
Weinreich et al.\ argue that theories of language change could consider different temporal scales with their own descriptive machinery to explain changes unfolding over thousands of years down to those occurring within a single lifetime~\citep[p.103]{WeinreichLabovHerzog1968}.
Traugott emphasizes that language change requires both individual innovation on the part of producers and recipients of language, as well as evidence that innovative language use has become more widely adopted or conventionalized~\citep{SemanticChange}.
In this study, we take into account these insights from theoretical linguistic experts, and posit that there is no reason to expect that interesting changes in the uses of a word, the introduction of new senses, or the broadening or narrowing of existing ones, should occur neatly on either side of a temporal dividing line placed in the middle of a diachronic corpus.
While much of the work in Lexical Semantic Change Detection (LSCD)
(see below in Section~\ref{ch8:priors}) has used a two-era task formulation that economizes annotation effort by producing gold labels of semantic change across two time periods~\citep{periti-tahmasebi-2024-towards}, e.g.\ \citet{0f0fecbd3e8e4951acbd9411cd658d06,ling-etal-2023-construction,chen-etal-2023-chiwug-graph,d-zamora-reina-etal-2022-black}, we therefore go beyond relying on two-era comparisons by expanding the number of time periods that we investigate and operating at the granularity of single decades.
From a methodological perspective, this change in focus
aspires to measure or predict noun compound compositionality --as a special aspect of compound meaning-- over multiple time steps, to avoid over-aggregating variation where it occurs. Additionally, there is a practical benefit to working with compositionality within singular example contexts, as opposed to relying on relatedness judgments across a large number of example combinations, because each example use of a compound provides a direct comparison with its constituents.  

The specific type of semantic change that we focus on in this study is a change in the degree of compositionality.
Studying the semantic change of noun compound meanings (via changes in compositionality) differs from the study of changes in simplex words because --as outlined above-- it is not only the compound as a whole that can change over time, but also the individual constituent words.
One hypothesis is that compound nouns tend to become less compositional over time, tending to begin as  straightforward combinations of constituents \citep{Wisniewski:96, Libben:98} and acquiring other (perhaps metaphorical) uses over time~\citep{bybee2015language}.
This hypothesis is not uncontested.
~\citet{Bell2016} observe that the constituent \textit{web} in the compound \textit{web site}, as annotated for compositionality by~\citet{ReddyEtAl:11a},
was rated as relatively compositional (2.7 on a scale from 0--5, where 5 is most compositional),
in spite of it clearly being a metaphorical invocation of a structure like a spider web, used to describe the link structure between different pages of content hosted online.
~\citeauthor{Bell2016} speculate that the high frequency and salience of this newer sense of \textit{web}, used to talk about the \textit{World Wide Web}, results in it being perceived by the annotators as (relatively) compositional
\footnote{The metaphorical foundations of \textit{web} in \textit{web site} are perhaps even more obscure in common usage since these ratings were published in 2011 --- after all, the Facebook `timeline' debuted that same year, signaling a shift away from `homepages' to endlessly-scrollable `feeds'.}.
The non-diachronic approach of studies that have collected compositionality judgments like~\citet{ReddyEtAl:11a} leave us to speculate whether there was a time when \textit{web site} was used and perceived as a non-compositional usage\footnote{Perhaps by the much smaller population of pre-WWW internet-users.}, and if this, contrary to~\citet{bybee2015language}, preceded its use and perception as a more compositional compound.
At the same time, the idiosyncrasies of the annotation setup of~\citet{ReddyEtAl:11a} caution us from over-interpreting its implications for historical trends in (non)-compositionality. In that study, annotators were provided with specific definitions of the compounds, e.g.\ the single definition of \textit{web site}: ``a computer connected to the internet that maintains a series of web pages on the World Wide Web''~\citep{ReddyEtAl:11a}.
Given that annotators were later asked to judge the compositionality\footnote{They used the term `literal'} of \textit{web} with respect to \textit{web site}, \textbf{in the sense of the provided definition} it is less surprising that this sense of \textit{web} (perhaps with a capital W) seems to have been salient for the responses.
In order to investigate these perspectives on the diachronic development of noun compound compositionality, 
we rephrase Bybee's hypothesis more precisely as our research question \textbf{RQ1}: \textbf{do compounds maintain their degree of compositionality or become less compositional over time?}


In order to investigate these general rules and expectations about the trajectory for noun compound compositionality over time, we conduct a series of computational experiments on English and German compounds, to compare diachronic compositionality annotations against compositionality predictions made with distributional semantic models.
We do so in the following sequence: (1) we conduct annotations of compositionality for in-context examples over several consecutive time periods, (2) we examine what trends in compositionality are apparent in the small sample of annotated data, (3) we compare several computational models of predicting compositionality over time against the annotated data in order to select approaches that best correspond with the annotations, and finally (4), we use the best models to examine our target compounds, asking again whether we can detect a trend in compositionality over time.


Developing an approach enabling a more gradual analysis of semantic change is instantiated by introducing the \textbf{Compositionality Trend Prediction (CTP)} task: a relational comparison between the meanings of target compounds and the meanings of their constituents, assessed over time, thus allowing for a measure of the speed of semantic change, not merely its strength.
The CTP task and the per-decade annotated compositionality ratings\footnote{Code for experiments as well as the annotations are available at \url{https://gitlab.com/cjenk/composition-tendency}} provide a test-bed for evaluating the effects of using different amounts (by decade) of historical context when making semantic predictions.
Additionally, we experiment with models of varying complexity, and with three standard metrics for measuring the semantic relatedness of word embeddings, with the goal of producing modeling recommendations appropriate for historical, small-data, and other modeling scenarios where data sources should not be grouped together as an undifferentiated mass.
In summary, our second research question \textbf{RQ2} can be stated as: \textbf{do contextual models using fine-grained temporal data perform better on the CTP task?}
This question counterpoises the temporal and context-sensitivity of the CTP task against the potential benefit of using more numerous and/or more aggregated masses of training examples.


The remainder of the paper is structured as follows: in Section~\ref{ch8:priors}, we provide a literature review that situates our work among the wider lexical semantic change community, with its particular focus on the phenomenon of noun compounds. Section~\ref{ch8:annotation} describes the diachronic corpora that we use, our annotation process, and our analysis of compositionality trends within the data, pursuant of RQ1. Section~\ref{ch8:sec:exp-setup} details our computational experiments in the CTP task, our selection of models and of temporal windows of training data used to train the models, and our analyses of aligning model predictions with the annotated compositionality ratings, which serves to answer RQ2.
Our contributions are: (i) a challenging new semantic change dataset and task for an underexplored lexical phenomenon, and (ii) experiments that shed light on the contribution of contextual representations and varying temporal granularities of training data to this time-sensitive task.
\section{Prior Work}\label{ch8:priors}
This literature review points to exemplars in two formulations of the Lexical Semantic Change Detection task (\textbf{LSCD}): semantic change described between two eras, and change that is measured more continuously.
We contrast this work with that of synchronic compositionality prediction, and situate it among several precedents largely drawn from the our collective prior works.
Finally, we discuss precedents in the use of BERT-type models that were trained using historical texts.
\paragraph{Two-Era Semantic Change Detection}
The LSCD evaluation format established by the SemEval 2020 Task 1~\citep{schlechtweg-etal-2020-semeval} is a major point of departure for this study.
It consists of two time-delimited sub-corpora ($C_1$ and $C_2$), for which target terms are labeled to have changed or stayed the same (binary classification), or else ranked according to the strength of the change between the sub-corpora (ranking). Change for the binary task is defined as the addition or loss of senses, while change in the ranking task is defined as a difference in the distribution of senses of the target term between the sub-corpora.
The relative sparseness of noun compounds, when compared with the (majority) simplex words used as targets in SemEval 2020 Task 1 motivate the use of a different temporal paradigm, where each target is evaluated over its own sequence of decades, which may or may not overlap with other targets.

The two-era paradigm was used in our previous work~\citep{jenkins-etal-2025-multi} to compare clustering and contextualized embedding aggregation approaches for predicting semantic changes in German and English noun compounds.
A shortcoming of this approach was the scalability of the annotations used to produce its gold standard ratings of compounds that had undergone semantic change or remained stable, as it relied on pairwise relatedness ratings of contexts in which noun compounds were used.
This paradigm can be seen in a wide variety of language change papers, e.g.\ \citet{gulordava-baroni-2011-distributional,mitra-etal-2014-thats,montariol-etal-2021-scalable,giulianelli-etal-2022-fire,cassotti-etal-2023-xl,hoeken-etal-2023-towards-detecting}.

\paragraph{Gradual Semantic Change Detection}
Though much of the recent work in semantic change modeling uses the two-era paradigm, there are important precedents modeling change in a more continuous fashion.
~\citet{10.1162/tacl_a_00081} used a Bayesian model operating over linear context windows of target terms, drawing conclusions from the fluctuation of terms in these context windows, modeling a sense distribution for each target term (with a number of senses fixed in advance), with each sense in a distribution having its own distribution of context words. 
Probabilities for these distributions were conditional on the previous time spans, which ablation experiments showed to be beneficial.

~\citet{tahmasebi-risse-2017-finding} also considered changes in word senses over time, by first using a clustering algorithm to obtain clusters (represented by salient co-occurring words to the target word), which were then progressively merged into larger units that traversed multiple time slices, using a similarity measure derived from WordNet~\citep{miller1995wordnet} to determine which clusters to merge.
While this method allowed them to model the expansion and contraction of sense inventories for pre-selected words such as \textit{aeroplane} across a diachronic corpus, the authors acknowledged that its reliance on WordNet meant that older meanings of words were less well-represented.

~\citet{giulianelli-etal-2020-analysing} used a single representational space to model semantic change, but analyzed changes in their representations across a continuum of time-steps (one per decade).
They used the diachronic English COHA dataset (introduced below in Section~\ref{ch8:diachronic-corpora}), but limited themselves to the \nth{20}--\nth{21} century portions of the data, which is prudent, considering that they used an (unaltered) language model trained on modern texts --- BERT~\citep{devlin-etal-2019-bert}, and used semantic change annotations (from~\citet{gulordava-baroni-2011-distributional}) that were presented to annotators without textual context, only using their intuition about how much they thought the target words had changed in the preceding 40 years.

A major influence for this study is~\citet{periti-tahmasebi-2024-towards}, a position paper that advocates for extending the LSCD task from describing semantic changes between two time periods to multiple time periods.
This influence can be seen in the way that this study's experimental framework operationalizes change over time on a decade-by-decade basis, and in our experiments using incremental clustering.
The open issue that Periti and Tahmasebi identify, the challenge of creating diachronic sense inventories, is only indirectly addressed in this study. Creating such a resource is assuredly a good idea, but it is also by far more onerous an undertaking than the diachronic compositionality ratings used in this study.
Although the CTP task does not evaluate against diachronic sense inventories, we take an important step in this direction by measuring compositionality over time, telling us something about the trajectory of a noun compound's semantic changes without the ontological overhead of delineating each compound's possible set of
senses.
The meaning of each compound is evaluated in the CTP task with respect to the meanings of its constituents, which are assumed to be a more stable reference point, which is similar to the technique discussed in~\citet[p.65]{Tahmasebi2021}, and used in~\citet{asgari-etal-2020-emblexchange} to use more stable semantic reference points in order to measure changes in the use of other terms.


\paragraph{Diachronic Compositionality Prediction}
Compared with prior work in synchronic compositionality prediction (e.g.~\citet{ReddyEtAl:11a,schulte-im-walde-etal-2016-role,cordeiro-etal-2019-unsupervised,alipoor-schulte-im-walde-2020-variants,miletic-schulte-im-walde-2023-systematic}),
there are relatively few works with a diachronic component.
This is not surprising, owing to the added difficulty of establishing a ground-truth of compositionality ratings \textit{as used in the past}.
Works that bridge the gap between synchronic and diachronic compositionality prediction include~\citet{dhar-etal-2019-measuring}, who jointly trained Word2Vec~\citep{https://doi.org/10.48550/arxiv.1301.3781} embeddings for noun compounds and their constituents, and compared them using cosine similarity across different segments of a diachronic corpus in order to estimate the degree of compositionality at different points in time.
Other approaches include~\citet{mahdizadeh-sani-etal-2024-diachronic-contexts}, who applied topic models to co-occurrence vectors of noun compounds and their constituents, using similarity comparisons at different time slices to make binary high/low compositionality predictions.
Expanding on this work,~\citet{miletic-schulte-im-walde-2025-modeling} broadened the approach of using historical developments to predict present-day noun compound compositionality to compare cooccurrence vectors, Word2Vec representations, topic models, and pretrained Transformers/BERT models (including some pretrained on pre \nth{20} century English). 
Notably for the present study, they found that the decontextualized Word2Vec representations were comparable or in some cases better than the more complex BERT-based representations in terms of compositionality classification accuracy. 
The historical BERT models fared poorly in classification accuracy, possibly owing to the modern skew of the surveyed target compounds within the COHA corpus. 

All of these studies used diachronic data to make a prediction against synchronic compositionality ratings, not to predict how compositional any target compound might have been in the past.
In contrast, the present study explicitly incorporates change over time into its evaluation framework.

\paragraph{Historical BERTs}

We were encouraged to train all of the models in this study from scratch by prior research which had trained Transformer-based models exclusively on historical data. These include MacBERTh~\citep{manjavacas-arevalo-fonteyn-2021-macberth}, trained on English texts ranging from 1450 to 1950 (of which COHA is a subset), or the subsequent work in~\cite{jdmdh:9152} where the authors found that pre-training on the MacBERTh dataset, rather than adapting a modern model to the historical dataset had better downstream performance on tasks whose test sets were created from texts within the 1450--1950 time span (e.g.\ POS tagging, word-sense disambiguation tasks).

\section{Data and Annotation}\label{ch8:annotation}

\subsection{Diachronic Corpora}\label{ch8:diachronic-corpora}


The German diachronic corpus that we use is the Deutsches Textarchiv (\textbf{DTA}) \citep{dta}.
The DTA is a \textbf{reference corpus} of the German language, containing texts from 1472 to 1969, with a focus on the \nth{17} through \nth{19} centuries.
The deliberate curation of the DTA means that it is meant to be representative beyond its particular sample of texts.
Orthographic normalization and modernization was performed on the DTA texts, rendering, e.g.\ `\longs' as the modern `s'.
It is curated\footnote{\url{https://www.deutschestextarchiv.de/doku/ueberblick}} to balance between fiction, non-fiction (including newspapers and letters), and scientific writing, but not necessarily the relative amount of text per year.

Table~\ref{ch2:tab:dta-freq} contains per-decade token and document counts for the decades that we use here. We can see that most decades contain more than 5 million tokens, drawn from around 100 documents per decade (fewer in the early \nth{18} century). The decade with the most material is the 1840s with around 17 million tokens across 884 documents. Generally speaking, the \nth{19} century portion contains the most material.

\begin{table}
  \centering\footnotesize
  \begin{tabular}{lrr}\toprule
    Decade & Tokens & Documents \\ \midrule
1680 & 6,446,985 &  49 \\
1690 & 4,527,661 & 47 \\
1700 & 7,108,977 &  62 \\
1710 & 5,456,632 & 62 \\
1720 & 6,685,974 &  63 \\
1730 & 3,637,782 & 68 \\
1740 & 5,594,914 & 115 \\
1750 & 5,098,525 & 57 \\
1760 & 4,268,880 & 54 \\
1770 & 8,675,796 &  107 \\
1780 & 6,744,396 &  137 \\
1790 & 9,828,721 &  196 \\
1800 & 5,829,689 & 121 \\
\bottomrule
  \end{tabular}
\begin{tabular}{lrr}\toprule
  Decade & Tokens & Documents \\ \midrule
1810 & 5,462,406  & 102 \\
1820 & 4,975,142  & 105 \\
1830 & 10,403,118  &  136 \\
1840 & 17,012,091  &  884 \\
1850 & 14,292,125  &  215 \\
1860 & 12,548,429  &  100 \\
1870 & 11,874,905  &  196 \\
1880 & 10,192,408  &  183 \\
1890 & 13,450,505  &  219 \\
1900 & 3,970,209  & 130 \\
1910 & 5,716,294  & 203 \\
1920 & 447,954  & 7 \\
& & \\
    \bottomrule
  \end{tabular}
  \caption{Per decade token counts and document counts in DTA}\label{ch2:tab:dta-freq}
\end{table}


For English we use the Corpus of Historical American English (\textbf{COHA}) \citep{Davies2012ExpandingHI}.
Substantial cleaning was performed by \citet{alatrash-etal-2020-ccoha}, resulting in the Cleaned Corpus of Historical American English (\textbf{CCOHA}).
We use the cleaned version for all of our analysis and experiments, referring to it as `COHA' for brevity.
COHA contains texts dating from 1810 to the 2010s.
Table~\ref{ch2:tab:ccoha-freq} shows per-decade token and document counts in COHA.
Generally speaking, later decades contain more text than earlier ones, with the first few decades of the \nth{19} century falling short of the 20+ million words that later decades contain.
In terms of genre, there is balancing between works of fiction (approximately half of each decade) against roughly equal contributions of newspapers, magazines and non-fiction works.
Notably, there are no newspaper entries included in the corpus for the 1810s through 1850s.
Attempts were made to balance sub-genres within the fiction and non-fiction entries \citep{Davies2012ExpandingHI}.

Many texts in COHA (specifically newspapers from 1850--1980, and some scanned books from 1900--1990~\citep{Davies2012ExpandingHI}) are the result of OCR processes, which can be error-prone,  especially for older, more cheaply produced texts like newspapers. The creators of the COHA describe a filtering process by which they rejected scanned texts that contained less than 98\% of types (unique word forms) overlapping with the set of extant types in their Corpus of Contemporary American English (COCA)~\citep{Davies2008COCA}.
Though this filtering undoubtedly makes the corpus easier to use, since many poorly-digitized texts are excluded, it may also present an artificially uniform view of the past, by penalizing archaic orthography, the use of words from other languages, or discursive topics that happen to not occur in the more modern reference corpus.

The earlier, \nth{19} century texts are primarily sourced from Project Gutenberg\footnote{\url{https://www.gutenberg.org/}} and the University of Michigan's `Making of America' digital library\footnote{\url{https://quod.lib.umich.edu/m/moagrp/}}, and some additional non-fiction material was sourced from \url{archive.org}~\citep{Davies2012ExpandingHI}.
The 1990--2010 material is taken from the COCA corpus~(\textit{ibid.}).
Unlike the DTA, the COHA does not contain any orthographic normalization layer outside of the filtering process described above.
At the same time, the corpus covers a more modern time period than the DTA, which should bring it into closer alignment with contemporary (US-American) standardized orthography.

Another noteworthy aspect of the COHA is that it systematically masks 5\% of tokens from its source texts~\citep{alatrash-etal-2020-ccoha}, for legal reasons, to avoid distributing \textit{readable} copies of the texts contained in the corpus, while still retaining \textit{most} of the statistical, distributional information in the texts.
This means that the COHA cannot be reliably used for the analysis of long-running passages of text, and it complicates the provision of context around an example item of interest, because it may occur near a sequence of redacted tokens.
The system of systematic omissions applies to all decades of the corpus, including those that are old enough to fall into the public domain, unfortunately further limiting the amount of examples that we are able to draw on.

Table~\ref{ch2:tab:ccoha-freq} shows the per decade count of tokens and documents in the COHA (not including redacted tokens).
We can see that most decades have about 20M tokens, with the bulk of the \nth{19} century decades having a few million fewer tokens, and the first two decades, 1810 and 1820 containing only about a 1M and 5.5M tokens respectfully.

\begin{table}
  \centering\footnotesize
  \begin{tabular}{lrr}\toprule
    Decade & Tokens & Documents \\ \midrule
1810 & 1,066,214   & 63 \\
1820 & 5,756,424   & 330 \\
1830 & 11,502,456  & 612  \\
1840 & 13,667,399  & 616  \\
1850 & 14,157,989  & 648  \\
1860 & 14,819,856  & 1,083  \\
1870 & 16,524,529  & 1,885  \\
1880 & 17,955,653  & 2,709  \\
1890 & 18,200,736  & 2,934  \\
1900 & 20,012,509  & 2,981  \\
\bottomrule
\end{tabular}
  \begin{tabular}{lrr}\toprule
    Decade & Tokens & Documents \\ \midrule
1910 & 20,918,990  & 3,355 \\
1920 & 23,873,684  & 11,557 \\
1930 & 23,042,600  & 10,352 \\
1940 & 23,113,886  & 11,343 \\
1950 & 23,457,457  & 11,935 \\
1960 & 22,866,052  & 10,113 \\
1970 & 22,696,646  & 9,419 \\
1980 & 24,295,129  & 11,106 \\
1990 & 26,858,129  & 9,778 \\
2000 & 28,375,158  & 13,795 \\
\bottomrule
  \end{tabular}
  \caption{Per decade token counts (excluding redacted tokens) and document counts in COHA}\label{ch2:tab:ccoha-freq}
\end{table}

\subsection{Selecting Target Compounds}\label{ch8:sec:selecting-targets}

Our goal was to select roughly 20 target compounds per language that we could meaningfully examine across one or more contiguous intervals of time, so as to be able to rate and predict their compositionality across those intervals.
The need to have many annotated examples in contiguous time intervals motivated us to focus on a small number of carefully sampled targets.
In total, we selected 23 German target compounds and 26 English target compounds.

Previous datasets of noun-noun compounds annotated for compositionality (\citet{ReddyEtAl:11a,cordeiro-etal-2019-unsupervised,schulte-im-walde-etal-2016-ghost}) were too skewed toward late \nth{20} and early \nth{21} century uses. Filtering for compounds that occurred at a reasonable minimum frequency (10 for English) in a large number of decades of the COHA corpus left us with too few targets --- a mere 19 remaining from the initial set of 210 noun-noun compounds.

As will be further discussed below, we used several criteria to select the target compounds.
The initial filtering was done according to compounds' overall frequency and their minimum per-decade frequency across a range of consecutive decades.
Further filtering was done by calculating the relative change in frequency between each decade for each potential target compound, to sort the possible targets by the number of between-decade changes that were larger than a set threshold.
Only potential targets with 1--4 of these inflection points of relative frequency changes above the threshold were retained.
Finally, a target set from the remaining compounds was manually selected according to broad thematic (e.g.\ pertaining to economic topics) or structural (e.g.\ sharing the same constituent) criteria.

In the remainder of this section, we describe in greater detail each step of the process of selecting the target compounds.
To obtain the English candidate compounds, we searched the COHA\index{corpora!COHA} diachronic corpus for candidate noun compounds.
These candidates were found based on the criteria that they were a sequence of two words tagged as nouns in the COHA dataset, and were not preceded or followed by a noun.
Compounds that were searched for in the corpus were initially in the format of \textit{open compounds} (i.e.\ space-separated).
Once selected as a target compound, examples were compiled from open, closed, and hyphenated compound variants.

German candidate compounds were obtained using a different process.
All of the modifier and head constituents from the GhoSt-NN dataset~\citep{schulte-im-walde-etal-2016-ghost} were used as potential components of other noun-noun compounds.
The DTA corpus was searched for items tagged as nouns that contained additional material after a word used as a modifier in the GhoSt-NN dataset, or before a word used as a head in the GhoSt-NN dataset.
Examples for candidate compounds were also collected for hyphenated and open compound variants, in addition to the more typical closed compound form.

Our goal was to find decades where there was an absolute change (up or down) in relative frequency compared with the previous decade, assuming that such points where a term is being used relatively more or less may be interesting center points around which to focus our investigation.
We call these \textbf{inflection points}, and emphasize that while they are an inflection of frequency, they are only \emph{potentially} sites of semantic change as well.

We compiled per-decade absolute frequencies for each remaining candidate compound, and calculated a change in relative frequency (Equation~\ref{eq:delta-rel-freq}).
The relative frequency of a term at time $t$ is defined as that term's frequency at time $t$ divided by the total number of tokens in that time period.
A change in relative frequency ($\Delta$) is the difference between the relative frequency at time $t$ and the relative frequency at the previous time $t-1$.

\begin{equation}\label{eq:delta-rel-freq}
  \Delta = \frac{\text{absolute target frequency}_{t}}{|\text{tokens}_{t}|} - \frac{\text{absolute target frequency}_{t-1}}{|\text{tokens}_{t-1}|}
\end{equation}

\begin{equation}\label{eq:infl-pts}
  \text{inflection points} = \Delta \in \Delta_{t_{2}...t_{n}} \text{where} \, \abs{\Delta} > \text{threshold}
\end{equation}

%
Inflection points per target are calculated per Equation~\ref{eq:infl-pts}.
We compared each $\Delta$ relative frequency ($\Delta$ being defined in Equation~\ref{eq:delta-rel-freq}, calculated with respect to the previous decade) against a threshold that was arrived at by trial and error\footnote{This ended up being \num{5e-6} for the German candidate targets, and \num{7e-7} for the English candidate targets.} for each corpus and candidate target set, arriving at a threshold where roughly 40\% of targets had either 1, 2, or 3 inflection points.
This gave us a set of targets for which we could focus on a smaller number of time-slices of the corpus, avoiding potential targets that either maintained a steady-state of relative frequency, or else experienced large changes in frequency between most of the decades.
After setting the relative frequency threshold for each language, we selected compounds that had between 1 and 4 inflection points according to the thematic criteria detailed below.

Ultimately, to avoid a large imbalance in the amount of material annotated for any one target compound, we only retained two inflection points per target compound, from a maximum of four allowed from the above filtering procedure.
In all cases where a target compound had more than one inflection point, these points were consolidated according to the following procedure.
First, if three decades in a row were inflection points, we retained only the central one.
Second, if two remaining inflection points were in adjoining decades, we retained only the earlier one.
Finally, if more than two inflection points remained, we retained only the first and last points. Each inflection point defines a window of up to 5 decades, centered on that point (unless the window abuts the end of the dataset).

A visual representation of the distribution of inflection points can be seen in Tables~\ref{tab:de-target-freq-1} and~\ref{tab:de-target-freq-2} (German) and Tables~\ref{tab:en-target-freq-1} and~\ref{tab:en-target-freq-2} (English), which show the frequency distributions of the target compounds in either dataset, along with the decades that were selected as inflection points (per-decade frequencies that are underlined). For example, in the English tables, we see that \textit{family physician} has an inflection point at the 1870s, and another at the 1970s.



The final selection of compounds was done according to their membership in thematic categories that we thought could potentially be sites of societal and/or semantic change.
We chose the following categories (in bold).
All targets are listed in Table~\ref{tab:ch8-targets}, sometimes falling into more than one category.
The first category are compounds describing designations/descriptions of individual people (\textbf{kinds of people} such as a \textit{university student}.
A natural extension of this category is to also look at compounds describing \textbf{groups of people}, for example, a \textit{rebel army}.
As both diachronic corpora contain some portion of the (European/North American) industrial revolution (irrespective of any particular periodization), and the societal changes that went along with it, we include compounds falling into the category of \textbf{economy}, including \textit{business community} and \textit{freight train}.
Alongside organizational technologies like train scheduling came potentially changing concepts related to \textbf{time}, e.g.\ \textit{leisure time}. 
Due to the various advances in medicine during the time periods included in either corpus, we looked for \textbf{health} terms among the remaining compounds, but found only one for each language, \textit{Krankenhaus} (hospital) and \textit{family physician}.
Acting as a counterpart to these terms with a highly social dimension, compounds pertaining to various \textbf{mundane objects} were included as well.
Some of these objects (e.g.\ \textit{silk hat}, an expensive and formal article of clothing, possibly used metonymically to refer to a wealthy person wearing one) may have some associations that could place them into one of the other categories.
Several of the German targets do not fall into any of these thematic categories, but were instead chosen because they share a constituent (e.g.\ \textit{Karten\textbf{spiel}, Kirch\textbf{spiel}}) or have the inverse order of another target (i.e.\ \textit{\textbf{Brief}wechsel, Wechsel\textbf{brief}}).
Additionally, it can be observed that \textit{friend, girl, hour, Mann, Bild, Stand, Haus, and Spiel} appear as the head constituent of more than one target.
Finally, \textit{Donnerwetter} was chosen because it was used by~\citet{schlechtweg-etal-2018-diachronic} as an example of a compound that gained an additional sense (as an exclamation) in addition to its earlier sense (thunderstorm) sometime around the turn of the \nth{19} century.

\begin{table}
  \centering
  \resizebox*{!}{\textheight}{%
  \begin{tabular}{ll}\toprule
    Targets (DE) & Targets (EN) \\ \midrule
    \textbf{Kinds of People} & \\
     \textit{Edelmann} (nobleman)    & \textit{baby boy} \\
     \textit{Ehefrau} (wife)         & \textit{bosom friend} \\
     \textit{Handelsmann} (merchant) & \textit{country girl} \\
     \textit{Landsmann} (compatriot) & \textit{family physician} \\
     \textit{Steuermann} (steersman) & \textit{lady friend} \\
     \textit{Weibsbild} (woman\footnotemark) & \textit{peasant girl} \\
      & \textit{servant girl} \\
      & \textit{university student} \\\midrule
    \textbf{Groups of People}  & \\
     \textit{Bauernstand} (peasantry) & \textit{business community} \\
     \textit{Bürgerstand} (bourgeoisie / middle class) & \textit{reading public} \\
     \textit{Mittelstand} (middle class / small business) & \textit{rebel army} \\\midrule
    \textbf{Economy}  & \\
     \textit{Bauernstand} (peasantry) & \textit{business community} \\
     \textit{Bürgerstand} (bourgeoisie / middle class) & \textit{freight train} \\
     \textit{Wechselbrief} (bill of exchange) & \textit{peasant girl} \\
     \textit{Gasthaus} (inn / restaurant) & \textit{sell price} \\
     \textit{Handelsmann} (merchant) & \textit{servant girl} \\
     \textit{Kaffeehaus} (coffee house) & \textit{silk hat} \\
     \textit{Marktpreis} (market price)  & \textit{trust fund} \\
     \textit{Mittelstand} (middle class / small business) & \textit{work hour} \\\midrule
    \textbf{Conflict}  & \\
     \textit{Kreuzzug} (crusade)  & \\
     \textit{Kriegszeit} (wartime) & \\\midrule
    \textbf{Time}  & \\
     \textit{Kriegszeit} (wartime) & \textit{church bell} \\
    & \textit{leisure hour} \\
    & \textit{leisure time} \\
    & \textit{school day} \\
    & \textit{work hour} \\\midrule
    \textbf{Health}  & \\
     \textit{Krankenhaus} (hospital)  & \textit{family physician} \\\midrule
    \textbf{Mundane Objects}  & \\
     & \textit{cherry tree} \\
     & \textit{church bell} \\
     & \textit{coffee pot} \\
     & \textit{frame house} \\
     & \textit{gold ring} \\
     & \textit{ice water} \\
     & \textit{iron door} \\
     & \textit{silk hat} \\\midrule
    \textbf{Other (Structural)}  & \\
     \textit{Augenschein} (appearance) &  \\
     \textit{Briefwechsel} (correspondence) & \\
     \textit{Donnerwetter} (stormy weather / an exclamation) & \\
     \textit{Kartenspiel} (card game) & \\
     \textit{Kirchspiel} (church law / area of a church) & \\
     \textit{Sinnbild} (symbol) & \\
     \textit{Wortspiel} (wordplay) & \\
    \bottomrule
  \end{tabular}
  }
  \caption{German (DE) and English (EN) target compounds, grouped according to thematic category.
  English translations of German terms in parenthesis are not exhaustive.}\label{tab:ch8-targets}
\end{table}

\addtocounter{footnote}{-1} 
\stepcounter{footnote}\footnotetext{Currently very pejorative. We were influenced by~\citet{sokefeld2025semantic} in their work on the semantic change of (German) nouns referring to women.}

\subsection{Annotation Procedure}\label{ch8:annotation-procedure}

We commissioned a study to collect semantic ratings using participants hired from Prolific\footnote{\url{https://www.prolific.com/}} to annotate example contexts from the two diachronic corpora in terms of their degree of compositionality.
The annotations themselves were collected using Google Forms. 
Limitations of text style on the platform (bold font, etc) when creating forms programmatically were overcome by embedding images of the example contexts rendered using \LaTeX.

An example question can be seen in Figure~\ref{fig:annotation-en-standard-question} (its German equivalent is shown in Appendix~\ref{ch8:fig:example-question-de}).
Annotators were asked to read a context including a target compound (highlighted in bold font), and to rate the extent to which the overall meaning of the compound relates to each constituent of the compound.
The rating with respect to either constituent is given as a separate question, each requiring a rating from 0 (saying that the meaning of the compound overall is not related to the constituent) to 5 (saying that the meaning of the compound overall is highly related to the meaning of the constituent).

\begin{figure}
\includegraphics[width=\linewidth]{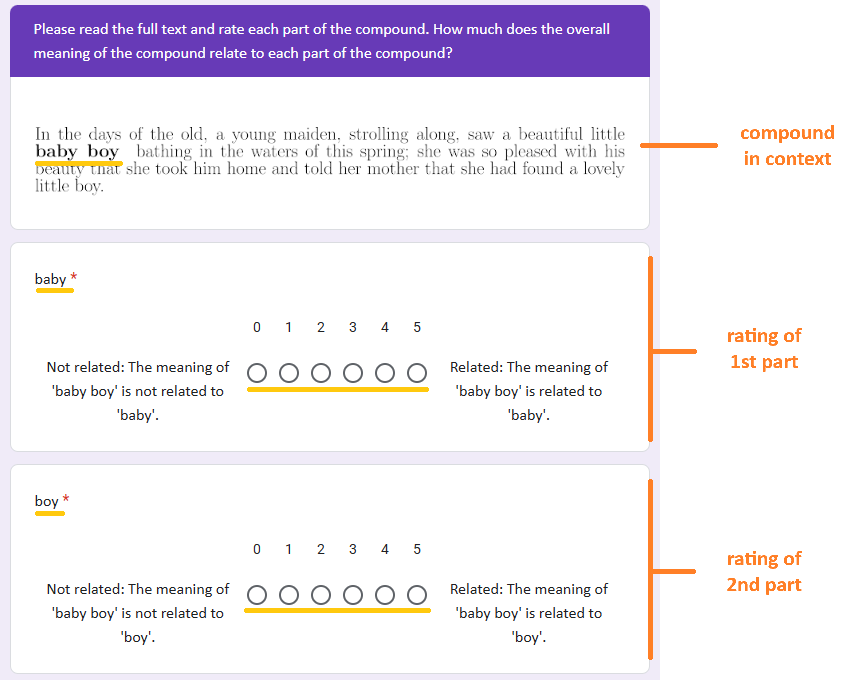}
\caption{Instructions for compositionality-in-context annotators with a sample question. }\label{fig:annotation-en-standard-question}
\end{figure}

\begin{table}[t!]
  \footnotesize
  \resizebox*{\textwidth}{!}{%
 \begin{tabular}{lrrrrrrrrrrrrr}\toprule
Target       & 1680 & 1690        & 1700         & 1710                 & 1720                 & 1730                 & 1740                 & 1750                 & 1760                 & 1770                 & 1780                 & 1790        & 1800        \\ \midrule
Augenschein  & 91   & \textbf{53} & \textbf{127} & \textul{\textbf{76}} & \textbf{86}          & \textbf{72}          & \textbf{107}         & \textul{\textbf{52}} & \textbf{58}          & \textbf{125}         & 83                   & 96          & 35          \\
Bauernstand  & 3    & 1           & 0            & 1                    & 3                    & 0                    & 9                    & 2                    & 1                    & 3                    & 16                   & 7           & 4           \\
Briefwechsel & 0    & 2           & 2            & 6                    & 4                    & \textbf{4}           & \textbf{148}         & \textul{\textbf{78}} & \textbf{12}          & \textbf{26}          & 51                   & 56          & 25          \\
Bürgerstand  & 2    & 7           & 3            & 0                    & 3                    & 3                    & 4                    & 2                    & 6                    & 3                    & 11                   & 34          & 17          \\
Donnerwetter & 22   & 4           & 21           & 13                   & \textbf{28}          & \textbf{59}          & \textul{\textbf{12}} & \textbf{11}          & \textbf{9}           & \textbf{17}          & \textul{\textbf{52}} & \textbf{36} & \textbf{12} \\
Edelmann     & 278  & 375         & 93           & \textbf{66}          & \textbf{80}          & \textul{\textbf{62}} & \textbf{80}          & \textbf{46}          & 56                   & 152                  & 110                  & 194         & 89          \\
Ehefrau      & 37   & 30          & 73           & 22                   & 60                   & 32                   & 65                   & \textbf{43}          & \textbf{53}          & \textul{\textbf{13}} & \textbf{13}          & \textbf{42} & 27          \\
Gasthaus     & 2    & 1           & 1            & 9                    & 2                    & 10                   & 3                    & 6                    & 1                    & 7                    & 4                    & 23          & 9           \\
Handelsmann  & 33   & 18          & 33           & \textbf{51}          & \textbf{8}           & \textul{\textbf{29}} & \textbf{18}          & \textbf{49}          & \textul{\textbf{11}} & \textbf{18}          & \textbf{10}          & 25          & 9           \\
Kaffeehaus   & 0    & 0           & 0            & 3                    & 0                    & 0                    & 4                    & 4                    & 6                    & 13                   & 23                   & \textbf{50} & \textbf{37} \\
Kartenspiel  & 5    & 10          & 18           & 3                    & 7                    & 2                    & 2                    & 2                    & 1                    & 8                    & 4                    & 39          & 8           \\
Kirchspiel   & 3    & 11          & 26           & 9                    & 17                   & 7                    & 39                   & 37                   & \textbf{25}          & \textbf{134}         & \textul{\textbf{18}} & \textbf{19} & \textbf{6}  \\
Krankenhaus  & 2    & 1           & 1            & 4                    & 1                    & 1                    & 2                    & 5                    & 4                    & 6                    & 6                    & 26          & 16          \\
Kreuzzug     & 1    & 0           & 1            & 0                    & 1                    & 6                    & 12                   & 15                   & 13                   & 58                   & 21                   & 28          & 14          \\
Kriegszeit   & 30   & 5           & 15           & 25                   & 12                   & 8                    & \textbf{13}          & \textbf{40}          & \textul{\textbf{3}}  & \textbf{34}          & \textbf{30}          & 31          & 5           \\
Landsmann    & 20   & 17          & 18           & 3                    & 7                    & \textbf{21}          & \textbf{33}          & \textul{\textbf{3}}  & \textbf{7}           & \textbf{25}          & 44                   & 45          & 31          \\
Marktpreis   & 0    & 0           & 0            & 0                    & 0                    & 0                    & 0                    & 7                    & 10                   & 2                    & 1                    & 3           & \textbf{18} \\
Mittelstand  & 0    & 4           & 0            & 1                    & 2                    & 2                    & 1                    & 3                    & 7                    & 10                   & 11                   & 50          & 5           \\
Sinnbild     & 52   & 28          & 24           & 18                   & 39                   & \textbf{15}          & \textbf{51}          & \textul{\textbf{18}} & \textbf{31}          & \textbf{45}          & 54                   & \textbf{45} & \textbf{59} \\
Steuermann   & 28   & 11          & 18           & 18                   & 11                   & 20                   & 10                   & 21                   & 5                    & 11                   & 14                   & 23          & 10          \\
Wechselbrief & 34   & 6           & \textbf{558} & \textbf{244}         & \textul{\textbf{29}} & \textbf{8}           & \textbf{4}           & \textbf{505}         & \textul{\textbf{22}} & \textbf{15}          & \textbf{2}           & 6           & 4           \\
Weibsbild    & 130  & 78          & 43           & 14                   & 44                   & 7                    & \textbf{12}          & \textbf{50}          & \textul{\textbf{9}}  & \textbf{28}          & \textbf{18}          & 6           & 1           \\
Wortspiel    & 0    & 0           & 0            & 0                    & \textbf{5}           & \textul{\textbf{22}} & \textbf{24}          & \textbf{10}          & 26                   & 19                   & 4                    & 12          & 25          \\
\bottomrule
 \end{tabular}
 }%
 \caption{German target frequencies in DTA (1680--1800), with inflection points underlined, decades covered by inflection points marked in bold.}\label{tab:de-target-freq-1}
\vspace{-2mm}
\end{table}

\begin{table}[h!]
  \footnotesize
  \resizebox*{\textwidth}{!}{%
 \begin{tabular}{lrrrrrrrrrrrrr}\toprule
Target       & 1810                 & 1820                 & 1830                 & 1840                  & 1850                 & 1860                 & 1870         & 1880                 & 1890                 & 1900                  & 1910         & 1920       & Total \\ \midrule
Augenschein  & 42                   & 23                   & 46                   & 95                    & 43                   & 52                   & 41           & 47                   & 53                   & 20                    & 23           & 0          & 1546  \\
Bauernstand  & 4                    & 8                    & \textbf{17}          & \textbf{176}          & \textul{\textbf{40}} & \textbf{89}          & \textbf{37}  & \textbf{54}          & \textbf{40}          & \textul{\textbf{34}}  & \textbf{10}  & 0          & 559   \\
Briefwechsel & 23                   & 33                   & 82                   & 104                   & 92                   & 31                   & 52           & 54                   & 86                   & 17                    & 62           & 0          & 1050  \\
Bürgerstand  & 14                   & \textbf{2}           & \textbf{17}          & \textul{\textbf{135}} & \textbf{47}          & \textbf{28}          & 25           & 10                   & 6                    & 5                     & 4            & 0          & 388   \\
Donnerwetter & 6                    & 12                   & 13                   & 33                    & 28                   & 20                   & 34           & 40                   & 19                   & 6                     & 20           & 0          & 527   \\
Edelmann     & 45                   & 30                   & 80                   & 140                   & 81                   & 134                  & \textbf{146} & \textbf{145}         & \textul{\textbf{65}} & \textbf{12}           & \textbf{53}  & 0          & 2612  \\
Ehefrau      & 17                   & 8                    & \textbf{9}           & \textbf{224}          & \textbf{41}          & \textul{\textbf{38}} & \textbf{66}  & 22                   & 49                   & 32                    & 96           & 1          & 1113  \\
Gasthaus     & 9                    & \textbf{7}           & \textbf{24}          & \textul{\textbf{137}} & \textbf{90}          & \textbf{48}          & \textbf{57}  & \textbf{64}          & \textbf{114}         & \textul{\textbf{114}} & \textbf{102} & \textbf{1} & 845   \\
Handelsmann  & 14                   & 6                    & 32                   & 72                    & 21                   & 16                   & 17           & 10                   & 15                   & 11                    & 14           & 0          & 540   \\
Kaffeehaus   & \textul{\textbf{6}}  & \textbf{7}           & \textbf{23}          & 123                   & 33                   & 18                   & 23           & \textbf{6}           & \textbf{19}          & \textul{\textbf{29}}  & \textbf{16}  & \textbf{6} & 449   \\
Kartenspiel  & 11                   & 4                    & 8                    & 20                    & 46                   & \textbf{31}          & \textbf{126} & \textul{\textbf{51}} & \textbf{15}          & \textbf{2}            & 7            & 1          & 431   \\
Kirchspiel   & 12                   & 4                    & 8                    & 46                    & 19                   & 33                   & 27           & 15                   & 22                   & 8                     & 16           & 0          & 561   \\
Krankenhaus  & 5                    & 4                    & 5                    & 54                    & 40                   & 82                   & 26           & \textbf{18}          & \textbf{72}          & \textul{\textbf{83}}  & \textbf{95}  & \textbf{6} & 565   \\
Kreuzzug     & 8                    & 9                    & 33                   & \textbf{87}           & \textbf{155}         & \textul{\textbf{34}} & \textbf{40}  & \textbf{47}          & 99                   & 41                    & 25           & 0          & 748   \\
Kriegszeit   & 5                    & 16                   & 14                   & 34                    & 32                   & 23                   & 33           & 30                   & 55                   & 6                     & 29           & 3          & 531   \\
Landsmann    & 27                   & 33                   & 68                   & 87                    & 88                   & 38                   & 54           & 81                   & 65                   & 9                     & 28           & 0          & 852   \\
Marktpreis   & \textbf{46}          & \textul{\textbf{15}} & \textbf{14}          & \textbf{11}           & 12                   & 20                   & 12           & \textbf{15}          & \textbf{192}         & \textul{\textbf{10}}  & \textbf{1}   & 0          & 389   \\
Mittelstand  & 17                   & 9                    & 30                   & 68                    & 38                   & 14                   & 62           & \textbf{15}          & \textbf{45}          & \textul{\textbf{47}}  & \textbf{27}  & 0          & 468   \\
Sinnbild     & \textul{\textbf{22}} & \textbf{25}          & \textbf{34}          & 29                    & 57                   & 84                   & 42           & 48                   & 17                   & 18                    & 10           & 2          & 867   \\
Steuermann   & \textbf{6}           & \textbf{56}          & \textul{\textbf{34}} & \textbf{37}           & \textbf{35}          & 47                   & 58           & 41                   & 19                   & 5                     & 28           & 0          & 566   \\
Wechselbrief & 8                    & 3                    & 1                    & 3                     & 4                    & 8                    & 0            & 0                    & 3                    & 0                     & 0            & 0          & 1467  \\
Weibsbild    & 3                    & 6                    & 9                    & 35                    & 30                   & 10                   & 17           & 22                   & 3                    & 2                     & 10           & 0          & 587   \\
Wortspiel    & 8                    & 4                    & 13                   & 15                    & 17                   & 61                   & \textbf{169} & \textbf{95}          & \textul{\textbf{12}} & \textbf{7}            & \textbf{1}   & 0          & 549   \\
\bottomrule
 \end{tabular}
 }%
 \caption{German target frequencies in DTA (1810--1920), with inflection points underlined, decades covered by inflection points marked in bold.}\label{tab:de-target-freq-2}
\vspace{-5mm}
\end{table}

Ten example contexts per target compound per decade (up to two decades before and after the inflection point)  were sampled and annotated.
In some cases, particularly in the first few decades of COHA and the last two decades of DTA, fewer than ten example contexts are available to be sampled from. In these cases
, we use all available examples.
In Tables~\ref{tab:de-target-freq-1} and~\ref{tab:de-target-freq-2}, we see the frequency of each of the German target compounds in the DTA, the inflection points per target compound (marked with an underline), as well as which decades were sampled from in the annotation (marked in bold).
Tables~\ref{tab:en-target-freq-1} and~\ref{tab:en-target-freq-2} show the equivalent for the English target compounds in COHA.

\begin{table}
  \footnotesize
  \resizebox*{\textwidth}{!}{%
 \begin{tabular}{lrrrrrrrrrr}\toprule
Target             & 1810       & 1820                 & 1830                 & 1840                 & 1850                 & 1860                 & 1870                 & 1880                 & 1890                 & 1900                 \\ \midrule
baby boy           & 0          & 0                    & 1                    & 0                    & 1                    & 2                    & 2                    & \textbf{16}          & \textbf{26}          & \textul{\textbf{9}}  \\
bosom friend       & \textbf{4} & \textul{\textbf{12}} & \textbf{14}          & \textbf{27}          & \textul{\textbf{13}} & \textbf{11}          & \textbf{24}          & 18                   & 7                    & 13                   \\
business community & 0          & 0                    & 2                    & 4                    & 4                    & 6                    & 20                   & 17                   & 12                   & 7                    \\
cherry tree        & 0          & 4                    & 5                    & 6                    & 10                   & 7                    & 7                    & 11                   & \textbf{14}          & \textbf{6}           \\
church bell        & 0          & 3                    & 10                   & 11                   & 16                   & 7                    & 11                   & 17                   & 16                   & 26                   \\
coffee pot         & 0          & 3                    & 0                    & 0                    & 2                    & 3                    & 2                    & \textbf{2}           & \textbf{3}           & \textul{\textbf{33}} \\
country girl       & \textbf{6} & \textul{\textbf{9}}  & \textbf{19}          & \textbf{16}          & 21                   & 23                   & 34                   & 23                   & 16                   & 29                   \\
family physician   & 0          & 1                    & 6                    & 5                    & \textbf{16}          & \textbf{10}          & \textul{\textbf{27}} & \textbf{22}          & \textbf{29}          & 21                   \\
frame house        & 0          & 2                    & 2                    & 2                    & 4                    & \textbf{4}           & \textbf{24}          & \textul{\textbf{13}} & \textbf{13}          & \textbf{9}           \\
freight train      & 0          & 0                    & 0                    & 1                    & 3                    & 4                    & 8                    & \textbf{28}          & \textbf{56}          & \textul{\textbf{25}} \\
gold ring          & 0          & 3                    & 3                    & \textbf{7}           & \textbf{6}           & \textul{\textbf{19}} & \textbf{24}          & \textbf{20}          & 16                   & 7                    \\
ice water          & 0          & 0                    & 0                    & 1                    & \textbf{3}           & \textbf{1}           & \textul{\textbf{1}}  & \textbf{23}          & \textul{\textbf{5}}  & \textbf{15}          \\
iron door          & 1          & \textbf{1}           & \textbf{7}           & \textul{\textbf{32}} & \textbf{28}          & \textul{\textbf{5}}  & \textbf{8}           & \textbf{12}          & 9                    & 24                   \\
lady friend        & 0          & 0                    & 0                    & \textbf{1}           & \textul{\textbf{14}} & \textbf{14}          & \textul{\textbf{37}} & \textbf{26}          & \textbf{15}          & 23                   \\
leisure hour       & \textbf{5} & \textbf{12}          & \textul{\textbf{38}} & \textbf{34}          & \textbf{41}          & 26                   & 24                   & 25                   & 20                   & 15                   \\
leisure time       & 0          & 1                    & 5                    & \textbf{14}          & \textbf{21}          & \textul{\textbf{7}}  & \textbf{21}          & \textbf{9}           & 13                   & 7                    \\
peasant girl       & 0          & 1                    & \textbf{10}          & \textbf{38}          & \textul{\textbf{10}} & \textbf{18}          & \textbf{13}          & 25                   & 19                   & 7                    \\
reading public     & 0          & \textbf{1}           & \textbf{21}          & \textul{\textbf{10}} & \textbf{13}          & \textbf{4}           & 14                   & 13                   & 28                   & 30                   \\
rebel army         & 0          & 3                    & 9                    & 1                    & \textbf{2}           & \textbf{66}          & \textul{\textbf{36}} & \textbf{46}          & \textul{\textbf{20}} & \textbf{9}           \\
school day         & 0          & 0                    & \textul{\textbf{3}}  & \textbf{2}           & \textbf{7}           & 10                   & 12                   & 19                   & 21                   & 23                   \\
sell price         & 0          & 0                    & 2                    & 1                    & 4                    & 3                    & 1                    & 12                   & 7                    & 14                   \\
servant girl       & 1          & 10                   & 12                   & \textbf{17}          & \textbf{20}          & \textul{\textbf{9}}  & \textbf{12}          & \textbf{20}          & 16                   & 30                   \\
silk hat           & 0          & 0                    & 2                    & 2                    & 1                    & 3                    & \textbf{11}          & \textbf{20}          & \textul{\textbf{41}} & \textbf{59}          \\
trust fund         & 0          & 0                    & 5                    & 1                    & 0                    & 3                    & 9                    & 7                    & 11                   & 25                   \\
university student & 0          & 0                    & 0                    & 1                    & 2                    & 1                    & 2                    & 3                    & 6                    & 12                   \\
work hour          & 0          & 1                    & 0                    & 3                    & 7                    & 6                    & 5                    & \textbf{13}          & \textbf{9}           & \textul{\textbf{33}} \\
\bottomrule
\end{tabular}
}%
\caption{English target frequencies in COHA (1810--1900), with inflection points underlined, decades covered by inflection points marked in bold.}\label{tab:en-target-freq-1}
\end{table}

\begin{table}
  \footnotesize
  \resizebox*{\textwidth}{!}{%
\begin{tabular}{lrrrrrrrrrrr}
Target             & 1910                 & 1920         & 1930                 & 1940                 & 1950                 & 1960                 & 1970                 & 1980                 & 1990                 & 2000        & Total \\ \midrule
baby boy           & \textbf{22}          & \textbf{9}   & 8                    & 8                    & 16                   & 14                   & \textbf{15}          & \textbf{18}          & \textul{\textbf{43}} & \textbf{59} & 269   \\
bosom friend       & 12                   & 14           & 10                   & 9                    & 4                    & 1                    & 1                    & 1                    & 2                    & 2           & 199   \\
business community & 9                    & 9            & 19                   & \textbf{17}          & \textbf{21}          & \textul{\textbf{94}} & \textbf{79}          & \textbf{73}          & 67                   & 44          & 504   \\
cherry tree        & \textul{\textbf{45}} & \textbf{39}  & \textbf{28}          & 27                   & 22                   & 14                   & 13                   & 16                   & 29                   & 21          & 324   \\
church bell        & 17                   & \textbf{22}  & \textbf{49}          & \textul{\textbf{66}} & \textbf{39}          & \textbf{24}          & 21                   & 26                   & 32                   & 26          & 439   \\
coffee pot         & \textbf{30}          & \textbf{31}  & 35                   & 25                   & 29                   & 34                   & 9                    & 15                   & 19                   & 20          & 295   \\
country girl       & 27                   & 21           & 15                   & 12                   & 29                   & 27                   & 13                   & 14                   & 18                   & 19          & 391   \\
family physician   & 27                   & 23           & 6                    & 14                   & \textbf{12}          & \textbf{36}          & \textul{\textbf{13}} & \textbf{12}          & \textbf{21}          & 18          & 319   \\
frame house        & 24                   & 46           & 59                   & 54                   & 38                   & \textbf{33}          & \textbf{43}          & \textul{\textbf{25}} & \textbf{36}          & \textbf{19} & 450   \\
freight train      & \textbf{29}          & \textbf{47}  & \textbf{39}          & \textbf{60}          & \textul{\textbf{31}} & \textbf{23}          & \textbf{24}          & 18                   & 31                   & 43          & 470   \\
gold ring          & 8                    & 19           & 16                   & 19                   & 12                   & 16                   & 11                   & 23                   & 29                   & 18          & 276   \\
ice water          & \textbf{11}          & 9            & 21                   & 28                   & 34                   & 24                   & 28                   & 28                   & 40                   & 61          & 333   \\
iron door          & 18                   & 20           & 16                   & 13                   & 28                   & 8                    & 8                    & 12                   & 13                   & 8           & 270   \\
lady friend        & 12                   & 21           & 14                   & 13                   & 11                   & 7                    & 13                   & 12                   & 14                   & 14          & 261   \\
leisure hour       & 17                   & 14           & 14                   & 11                   & 8                    & 4                    & 4                    & 4                    & 2                    & 1           & 319   \\
leisure time       & 7                    & 11           & 21                   & 26                   & 23                   & 29                   & \textbf{32}          & \textbf{9}           & \textul{\textbf{33}} & \textbf{31} & 320   \\
peasant girl       & 12                   & 4            & 11                   & 7                    & 10                   & 8                    & 11                   & 9                    & 6                    & 6           & 225   \\
reading public     & 34                   & 27           & 7                    & 8                    & 15                   & 11                   & 11                   & 8                    & 11                   & 6           & 272   \\
rebel army         & \textbf{6}           & 3            & 15                   & 24                   & 9                    & 1                    & 0                    & 5                    & 9                    & 9           & 273   \\
school day         & 32                   & 34           & 25                   & 24                   & 35                   & 33                   & \textbf{36}          & \textbf{21}          & \textul{\textbf{47}} & \textbf{79} & 463   \\
sell price         & \textbf{17}          & \textbf{109} & \textul{\textbf{61}} & \textbf{27}          & \textbf{15}          & 30                   & 16                   & 9                    & 5                    & 3           & 336   \\
servant girl       & 22                   & 11           & 16                   & 20                   & 11                   & 10                   & 10                   & 12                   & 7                    & 4           & 270   \\
silk hat           & \textbf{75}          & \textbf{70}  & \textul{\textbf{38}} & \textbf{25}          & \textbf{18}          & 12                   & 11                   & 16                   & 3                    & 1           & 408   \\
trust fund         & \textbf{11}          & \textbf{17}  & \textul{\textbf{49}} & \textbf{33}          & \textbf{30}          & \textbf{26}          & \textbf{42}          & \textul{\textbf{30}} & \textbf{47}          & \textbf{62} & 408   \\
university student & 9                    & 15           & 23                   & \textbf{18}          & \textbf{25}          & \textul{\textbf{39}} & \textbf{43}          & \textbf{29}          & 30                   & 39          & 297   \\
work hour          & \textbf{25}          & \textbf{36}  & \textul{\textbf{60}} & \textbf{48}          & \textbf{43}          & 36                   & 32                   & 33                   & 52                   & 33          & 475   \\
\bottomrule
\end{tabular}
}%
\caption{English target frequencies in COHA (1910--2000), with inflection points underlined, decades covered by inflection points marked in bold.}\label{tab:en-target-freq-2}
\end{table}

\begin{table}[t!]
    \centering\footnotesize
    \begin{tabular}{lrrrr} \toprule
        Corpus & Targets & Contexts & Avg.\ Num.\ Ratings  & Avg.\ $\alpha$ Agr.\   \\ \midrule
        COHA & 26  & 2012  &  7.63 ($\pm$1.52)  & 0.32 ($\pm$0.11) \\ 
        DTA & 23  & 1892  & 7.95 ($\pm$1.10)   & 0.41 ($\pm$0.09) \\ 
        \bottomrule
    \end{tabular}
    \caption{Annotated sentence pairs and agreement statistics using Krippendorff's $\alpha$.}
    \label{ch8:tab:annotations}
\end{table}

Summary statistics about the annotation are given in Table~\ref{ch8:tab:annotations}.
Each example context was rated by ten annotators who annotated the same batch of 30 contexts (60 questions in total), usually taking about 10--15 minutes to complete.
Annotators who worked on the English portion of the annotations were paid an average of 8.45GBP/h, annotators who worked on the German portion of the annotations were paid an average of 13.75GBP/h.
This difference is a consequence of more annotators on the German tasks having taken longer to complete the task.
Following procedures described by Prolific\index{Prolific} for so-called "Instructional Manipulation Checks"\footnote{\url{https://intercom-help.eu/prolific-research/en/articles/445153-prolific-s-attention-and-comprehension-check-policy} --- Last accessed 2025--10--28.}, we included additional questions designed to check that annotators were following the instructions. An example is given in Appendix~\ref{fig:annotation-en-check-question} (German translation in Appendix~\ref{ch8:fig:check-question-de}).
Annotators rejected via these criteria are not counted toward the ten total annotators who annotated each example.

Additional (post hoc) filtering of annotators was performed by calculating Krippendorff's alpha scores~\citep{krippendorff2013content,castro-2017-fast-krippendorff} to measure the agreement between each pair of annotators within a given batch.
Annotators were rejected if their average score between the other nine participants for their batch was less than $0.0$, a value representing a chance relation between ratings.
This tended to occur in cases where an annotator chose exclusively or nearly exclusively ratings of 0 or 5.
Of the 541 initial annotators for the English dataset, 119 were excluded by this criterion.
Of the 80 initial annotators for the German dataset (more German-speaking participants annotated more than one batch), 9 were rejected by this criterion.
The mean and standard deviation for the number of ratings reported in Table~\ref{ch8:tab:annotations} reflects this reduction from the initial 10 annotators per batch and by extension, per context to be annotated.
The mean and standard deviations of the agreement between annotators in each batch in Table~\ref{ch8:tab:annotations} also reflects the post-hoc filtering, making it much higher than the unfiltered agreement of $\alpha=0.16$ $(\pm 0.12)$ (EN) or $\alpha=0.30$ $(\pm 0.10)$.
All told, this points to (1) a task with some inherent difficulty (particularly the interpretation of the middle ranges of the scale), and (2) caution in over-interpreting the mean of the ratings for any item.

We also calculated inner-annotator agreement by the decade of the examples that were annotated. We drew inspiration from~\citet{cassotti-tahmasebi-2025-sense}, who noticed higher agreement on examples drawn from the year 1900 or later, compared with examples that were taken from older texts.
These results can be seen in Tables~\ref{tab:en-iaa-decades} (English) and~\ref{tab:de-iaa-decades} (German).
The per-decade agreement needs to be interpreted in the context of the temporal distribution of the targets and their inflection points (this can be seen by the position of underlined  decades (inflection point)\index{inflection point} and bold decades (included in window around an inflection point) in Tables~\ref{tab:de-target-freq-1},~\ref{tab:de-target-freq-2} and~\ref{tab:en-target-freq-1},~\ref{tab:en-target-freq-2}).
While the very earliest German examples show low agreement, there is no clear trend toward higher agreement as the examples become closer to the present day.

\begin{table}[t!]
  \centering\footnotesize
  \resizebox{\linewidth}{!}{
  \begin{tabular}{lrrrrrrrrrr} \toprule
    Decade &  & 1810   & 1820    & 1830    & 1840    & 1850    & 1860     & 1870    & 1880     & 1890\\
    IAA &        & 0.362  & 0.427 & 0.322 & 0.306 & 0.279 & 0.278  & 0.285 & 0.317  & 0.221 \\\midrule
    Decade & 1900  & 1910    & 1920    & 1930    & 1940    & 1950    & 1960    & 1970    & 1980    & 1990 \\
    IAA      & 0.224 & 0.317 & 0.259 & 0.264 & 0.336 & 0.313 & 0.296 & 0.266 & 0.316 & 0.346 \\\midrule
    Decade & 2000  & & & & & & & & & \\
    IAA      & 0.318 & & & & & & & & & \\
    \bottomrule
  \end{tabular}
  }%
  \caption{Per-decade inner annotator agreement scores calculated with Krippendorff's Alpha on the English / COHA dataset.}\label{tab:en-iaa-decades}
\end{table}

\begin{table}[t!]
  \centering\footnotesize
  \resizebox{\linewidth}{!}{
  \begin{tabular}{lrrrrrrrrrr} \toprule
    Decade & & & & & & & & & 1680  & 1690     \\
    IAA &  & & & & & & & &  0.181 & 0.154  \\\midrule
    Decade & 1700  & 1710    & 1720    & 1730    & 1740    & 1750    & 1760    & 1770    & 1780    & 1790 \\
    IAA    & 0.166 & 0.217 & 0.230 & 0.325 & 0.348 & 0.394 & 0.347 & 0.320 & 0.350 & 0.329 \\\midrule
    Decade & 1800  & 1810    & 1820    & 1830    & 1840    & 1850    & 1860    & 1870    & 1880    & 1890 \\
    IAA    & 0.237 & 0.294 & 0.247 & 0.301 & 0.327 & 0.317 & 0.365 & 0.309 & 0.288 & 0.240 \\\midrule
    Decade & 1900  & 1910  & 1920  & & & & & & & \\
    IAA    & 0.267 & 0.270 & 0.248 & & & & & & & \\
    \bottomrule
  \end{tabular}
  }%
  \caption{Per-decade inner annotator agreement scores calculated with Krippendorff's Alpha on the German / DTA dataset.}\label{tab:de-iaa-decades}
\end{table}

\subsection{Analysis of Annotation Results}\label{ch8:subsec:annotation-results}
Where in the previous section we focused on the meta-level details of collecting the in-context compositionality ratings, here we describe the results of this collection, and what they say about the tendency for noun compounds to become less compositional over time.


The first question to answer is whether there is an overall trend in compositionality across \textit{all} of the target compounds (\textbf{RQ1}). This means aggregating results across each constituent type, and over the one or two temporal ranges centered on an inflection point per target.
We measure the change across a temporal window simply by subtracting the first rating (average of annotators) from the final rating (also average of annotators).
A positive value indicates that the compound has become more compositional, while a negative value indicates that the compound has become less compositional.
This is represented by the $\Delta$ symbol in the tables that follow.
At the aggregate level, there is a very small negative trend in compositionality for both languages, a trend which is much smaller than the typical variation in ratings between one decade and the next.

For the German annotation, the mean $\Delta$ is $-0.0012$, for English it is $-0.0161$.
If we aggregate trends at the level of the constituent type (modifiers and constituents), we again see a very small average trend, but in the opposite direction for German: $\Delta$~modifiers: $-0.0036$, and $\Delta$~heads: $0.0012$.
In the English annotation, the trend direction is negative for both the average of all modifiers: $\Delta$ $-0.0217$, as well as the average of all heads: $\Delta$ $-0.0106$.
Our succinct answer to \textbf{RQ1} is that we cannot conclude that there is a generalized trend toward non-compositionality on this basis, as the magnitude of the aggregate changes is very low.

We can contextualize these average trends by looking at Tables~\ref{tab:de-anno-trends} and~\ref{tab:en-anno-trends} for German and English respectively.
These tables are sorted by the overall trend ($\Delta$) for each $\langle$compound, constituent, inflection point$\rangle$ triple, from the largest positive trend (more compositional at the end of the range) to the largest negative trend (less compositional at the end of the range).
The last column of either table shows the average absolute difference in compositionality ratings between each decade in the temporal range. This serves to summarize the variation between decades, regardless of the direction of the fluctuation.

We can see that the overwhelming majority of compositionality trends are smaller than $0.1$ (on a 6-point scale of compositionality), and that many of the larger average absolute differences across time-spans are also the triples that had the highest positive or negative trend.

\begin{figure}[h]
\centering\includegraphics[width=0.8\textwidth]{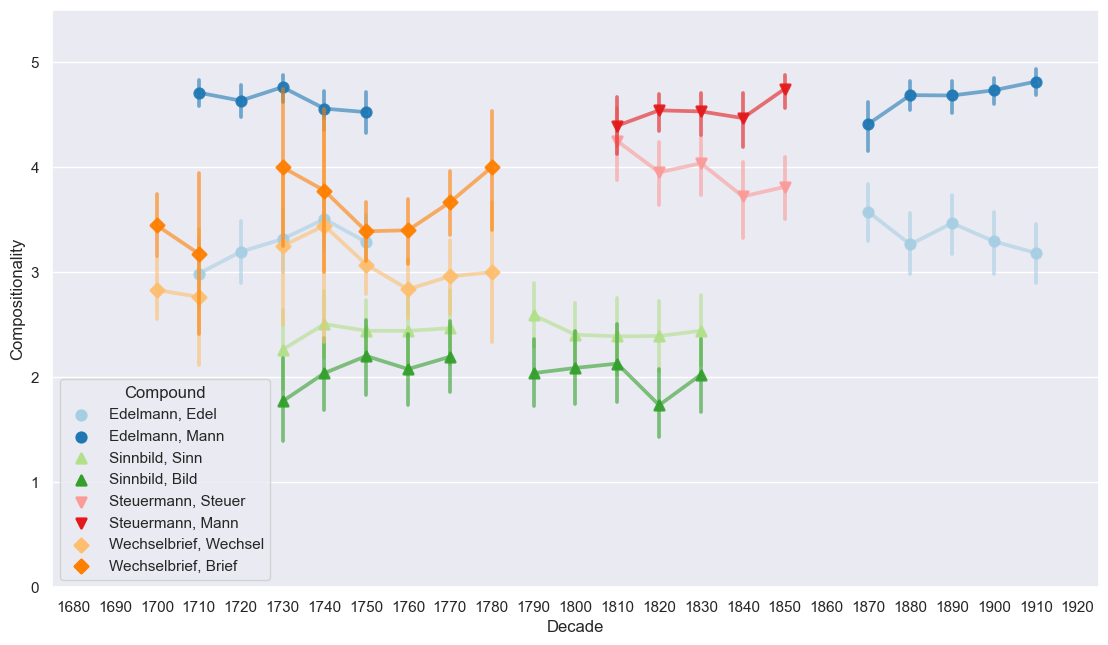}
\caption{Four German compounds with the largest positive trend in one of their $\langle$compound, constituent, inflection-point$\rangle$ triples.}\label{fig:de-most-pos}
\end{figure}

\begin{figure}
  \centering\includegraphics[width=0.8\textwidth]{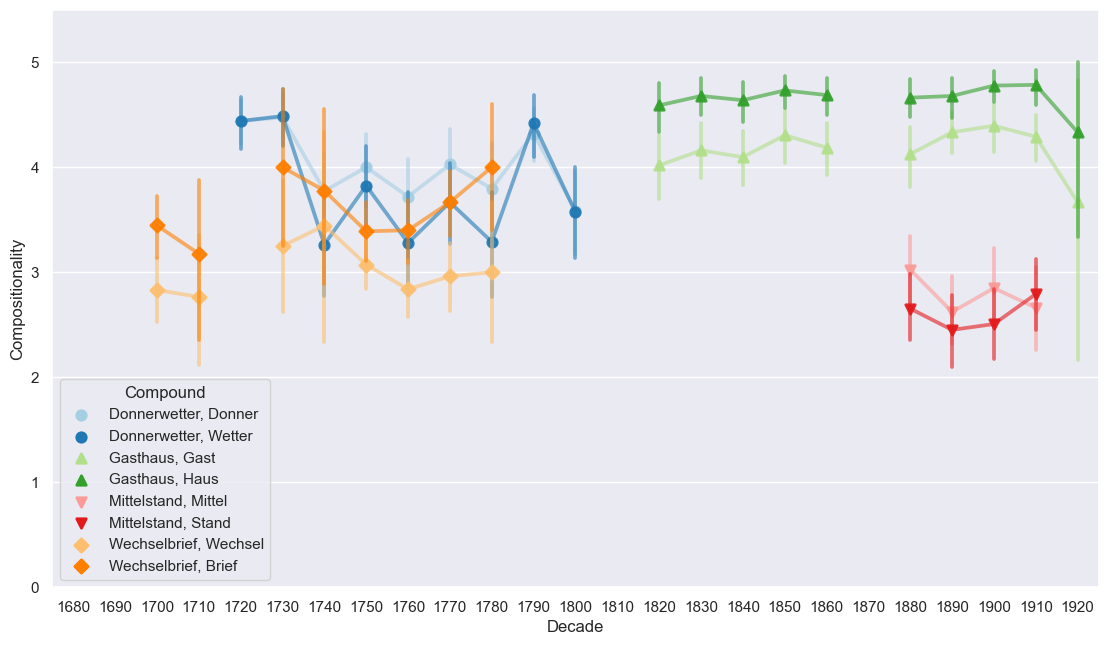}
  \caption{Four German compounds with the largest negative trend in one of their $\langle$compound, constituent, inflection-point$\rangle$ triples.}\label{fig:de-most-minus}
\end{figure}

To further emphasize the overall flatness of the trends, we can look to a pair of line charts for either language, where each pair shows the four compounds where at least one constituent at one inflection point had the largest positive or negative trend.
Each chart displays the per-decade average compositionality rating given by annotators, as well as 95\% confidence intervals which summarize the spread of annotations.
In Figure~\ref{fig:de-most-pos}, we see the German compounds where at least one constituent had the largest positive trend in Table~\ref{tab:de-anno-trends}: \textit{Edelmann}, \textit{Sinnbild}, \textit{Steuermann}, and \textit{Wechselbrief}.
We can see a marked divergence in the trends of the two constituents of \textit{Edelmann} and \textit{Steuermann}.
In Figure~\ref{fig:de-most-minus}, we see the German compounds where at least one constituent had the largest negative trend in Table~\ref{tab:de-anno-trends}: \textit{Donnerwetter}, \textit{Gasthaus}, \textit{Mittelstand}, and \textit{Wechselbrief}.
Two compounds stand out here: \textit{Donnerwetter}, which, as mentioned above, is heading in a non-compositional direction as we had expected from the prior work of~\citet{schlechtweg-etal-2018-diachronic}, but still remains at 1800 solidly above the middle range of the compositionality scale.
The second example to look more closely at is \textit{Gasthaus}, for which both constituents' average ratings drop in the final decade, but with variance stretching across much of the scale, particularly for the modifier \textit{Gast} (guest).
This is a consequence of this final decade (the 1920s) being represented by a single example
\footnote{Even out of context, it is recognizably from Kafka's \textit{Der Prozess} (The Trial), with its protagonist \textit{K.}: ``K. war sehr ermüdet, da er wegen einer Stammtischfeierlichkeit bis spät in die Nacht im \textbf{Gasthaus} geblieben war, er hätte fast verschlafen''. Annotators were also shown a sentence before and after this one.}
.
We can consider this second inflection point for \textit{Gasthaus} to be an outlier, but it should not generally affect our conclusions about the absence of a general trend in compositionality.

\begin{figure}
\centering\includegraphics[width=0.8\textwidth]{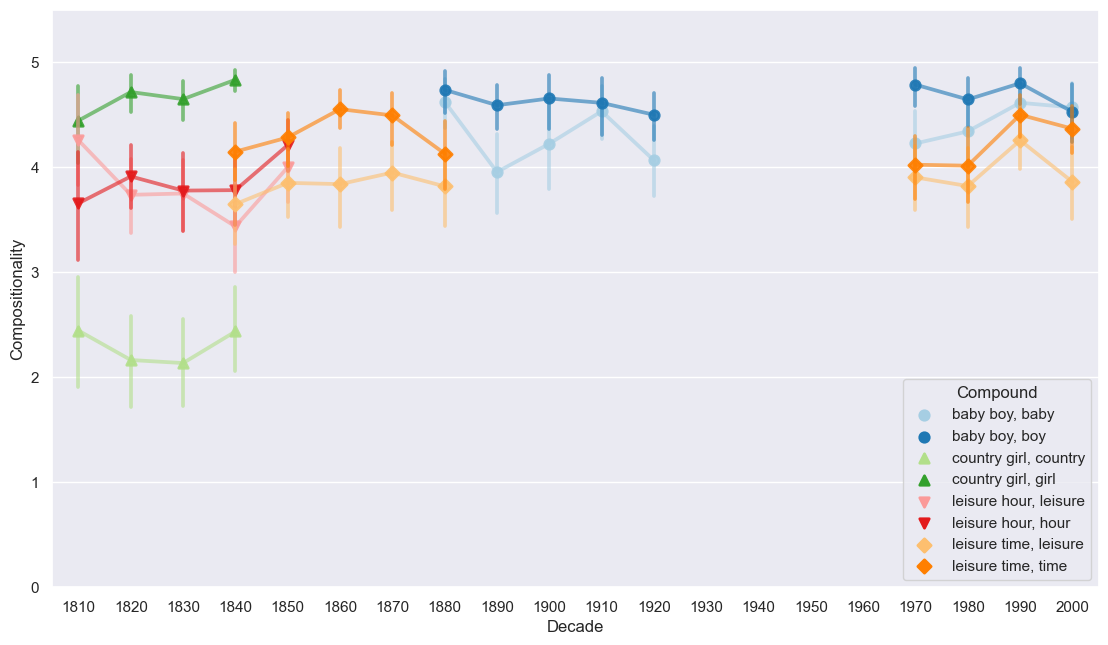}
\caption{Four English compounds with the largest positive trend in one of their $\langle$compound, constituent, inflection-point$\rangle$ triples.}\label{fig:en-most-pos}
\end{figure}

\begin{figure}
  \centering\includegraphics[width=0.8\textwidth]{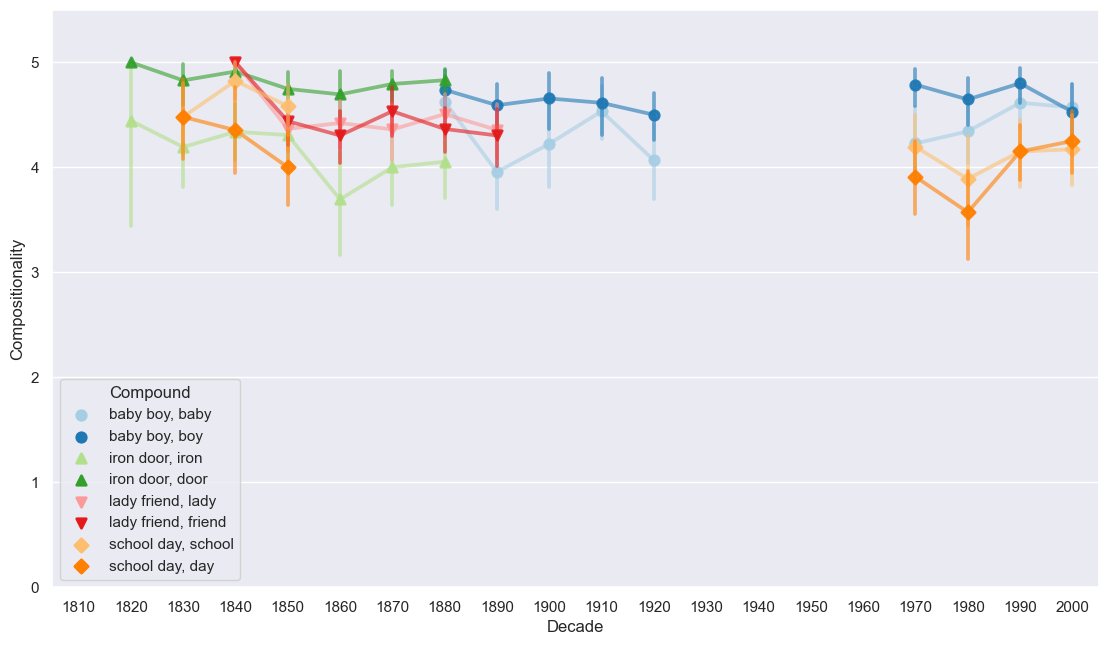}
  \caption{Four English compounds with the largest negative trend in one of their $\langle$compound, constituent, inflection-point$\rangle$ triples.}\label{fig:en-most-minus}
\end{figure}

Turning to the English chart showing the compounds where at least one constituent had the largest positive trend, we can refer to Figure~\ref{fig:en-most-pos}, which shows the targets \textit{baby boy} (the modifier centered on 1990), \textit{country girl}, \textit{leisure hour}, and \textit{leisure time}.
All trends are very slight. The variance in \textit{leisure hour} and \textit{leisure time} where they overlap temporally shows them as being rated similarly, generally with the modifier \textit{leisure} being less related to the overall meaning than the head \textit{time} or \textit{hour}. The greater specificity of \textit{hour} did not lead to it being rated as more compositional on average.

Finally, we come to the English chart showing the compounds where at least one constituent had the largest negative trend in Figure~\ref{fig:en-most-minus}. These compounds are: \textit{baby boy} (the modifier centered on 1900), \textit{iron door}, \textit{lady friend}, and \textit{school day}.
Our attention should be drawn to the modifier \textit{iron}, which shows a sharp drop in compositionality between the 1850s and 1860s, with a wide variance in the 1860s.
As above with the German example of \textit{Gasthaus}, this compound occurs very few times in that decade, with only five contexts. Two of these examples had sharp disagreement between annotators when rating the compositionality with respect to the modifier (with ratings of 0 and of 5).
These examples may be instructive of the overall difficulty of this annotation task:
\begin{displayquote}
A safe in the room, which contained some of the hated records, was fallen upon by the men, who strove to wrench open its impregnable lock with their naked hands, and, baffled, beat them on its \textbf{iron doors} and sides till they were stained with blood, in a mad frenzy of senseless hate and fury.
\end{displayquote}
which had an average rating of 3.28, and
\begin{displayquote}
Every frequenter of the State-House may remember seeing him, after being pestered beyond endurance, hasten across the antechamber into this room, where he would bolt and bar out the waiting crowd until he could finish some urgent work demanding freedom from the interruptions to which he was subject in his own apartment.
Once behind that \textbf{iron door} he was free; and it was the only place in the whole building where he was secure from intrusion.
His patience, however, under all manners of interruption, // was marvellous.
\end{displayquote}
which had a still lower average rating of 2.66.

Our reading of these two examples above would rather assign them a higher compositionality rating than the annotators did, especially in comparison to the example below, which received a higher average rating of 4.33:
\begin{displayquote}
As one who in the night, passing a street Deserted, finds a lost key rusted and old, Yet knows that it will fit some great \textbf{iron door} Behind which countless treasures are concealed, So I, when first I came to Mesmer's works, Knew I had $\lbrack\ldots\rbrack$ problems.
\end{displayquote}

Some of this inconsistency could be attributed to different groups of annotators having rated each of these contexts containing \textit{iron door} in the 1860s, or else to the basic difficulty of the annotation task due to the open-ended possibility of polysemy.


Although we did not find dramatic trends over time for any of our target compounds, and the overall temporal differences over time are slight, we can still draw useful distinctions between $\langle$compound, constituent, inflection-point$\rangle$ triples with the highest magnitude $\Delta$ compositionality, and those in the middle of the scale. Furthermore, the average change between decades shows that any particular pair of consecutive decades has some variance in compositionality ratings, and we can see experimentally to what extent different models for predicting compositionality align with this variance between decades~(\textbf{RQ2} asks whether contextualized models using fine-grained temporal data should perform better).
The experiments described below assume that the compositionality $\Delta$s and the per-decade average compositionality ratings (which collectively constitute the compositionality $\Delta$s) can be used as a gold standard.

\begin{table}[h]
  \footnotesize
  \begin{tabular}{lllrr} \toprule
    Compound & Const.\ & Infl.\ & $\Delta$ & avg.\ \\\midrule
 Wechselbrief &  mod & 1720 &  0.153 & 0.249 \\
     Sinnbild & head & 1750 &  0.085 & 0.168 \\
 Wechselbrief & head & 1720 &  0.083 & 0.438 \\
     Edelmann & head & 1890 &  0.081 & 0.102 \\
   Steuermann & head & 1830 &  0.070 & 0.125 \\
   Kaffeehaus &  mod & 1810 &  0.065 & 0.237 \\
     Edelmann &  mod & 1730 &  0.060 & 0.185 \\
 Donnerwetter & head & 1780 &  0.059 & 0.685 \\
    Wortspiel &  mod & 1890 &  0.053 & 0.206 \\
  Augenschein &  mod & 1710 &  0.047 & 0.059 \\
    Landsmann & head & 1750 &  0.045 & 0.114 \\
 Wechselbrief & head & 1760 &  0.044 & 0.249 \\
      Ehefrau &  mod & 1770 &  0.043 & 0.099 \\
  Kartenspiel & head & 1880 &  0.041 & 0.082 \\
     Sinnbild &  mod & 1750 &  0.041 & 0.084 \\
    Landsmann &  mod & 1750 &  0.040 & 0.249 \\
    Wortspiel & head & 1730 &  0.040 & 0.386 \\
  Mittelstand & head & 1900 &  0.035 & 0.181 \\
     Gasthaus &  mod & 1840 &  0.034 & 0.133 \\
   Kriegszeit &  mod & 1760 &  0.034 & 0.107 \\
    Wortspiel & head & 1890 &  0.033 & 0.171 \\
  Krankenhaus & head & 1900 &  0.031 & 0.051 \\
  Bauernstand & head & 1900 &  0.030 & 0.097 \\
     Kreuzzug &  mod & 1860 &  0.028 & 0.129 \\
  Bürgerstand &  mod & 1840 &  0.027 & 0.093 \\
  Bauernstand & head & 1850 &  0.022 & 0.152 \\
  Krankenhaus &  mod & 1900 &  0.021 & 0.099 \\
     Gasthaus & head & 1840 &  0.020 & 0.067 \\
      Ehefrau &  mod & 1850 &  0.018 & 0.091 \\
   Kaffeehaus & head & 1810 &  0.018 & 0.043 \\
   Kaffeehaus & head & 1900 &  0.018 & 0.057 \\
   Kaffeehaus &  mod & 1900 &  0.015 & 0.184 \\
     Kreuzzug & head & 1860 &  0.012 & 0.179 \\
   Marktpreis &  mod & 1900 &  0.010 & 0.282 \\
    Wortspiel &  mod & 1730 &  0.010 & 0.070 \\
    \bottomrule
  \end{tabular}
  \begin{tabular}{lllrr} \toprule
    Compound & Const.\ & Infl.\ & $\Delta$ & avg.\ \\\midrule
    Weibsbild &  mod & 1760 &  0.004 & 0.043 \\
  Augenschein &  mod & 1750 & -0.002 & 0.241 \\
  Augenschein & head & 1750 & -0.002 & 0.062 \\
  Bauernstand &  mod & 1850 & -0.002 & 0.092 \\
     Sinnbild & head & 1810 & -0.003 & 0.194 \\
      Ehefrau & head & 1850 & -0.006 & 0.017 \\
  Kartenspiel &  mod & 1880 & -0.009 & 0.094 \\
      Ehefrau & head & 1770 & -0.015 & 0.037 \\
  Handelsmann &  mod & 1760 & -0.015 & 0.081 \\
  Bauernstand &  mod & 1900 & -0.019 & 0.112 \\
  Handelsmann &  mod & 1730 & -0.021 & 0.105 \\
  Handelsmann & head & 1760 & -0.021 & 0.232 \\
   Marktpreis & head & 1820 & -0.021 & 0.058 \\
 Briefwechsel & head & 1750 & -0.022 & 0.261 \\
   Kriegszeit & head & 1760 & -0.022 & 0.270 \\
  Augenschein & head & 1710 & -0.023 & 0.204 \\
 Donnerwetter &  mod & 1780 & -0.027 & 0.448 \\
 Briefwechsel &  mod & 1750 & -0.029 & 0.128 \\
     Sinnbild &  mod & 1810 & -0.030 & 0.064 \\
  Bürgerstand & head & 1840 & -0.031 & 0.211 \\
  Handelsmann & head & 1730 & -0.036 & 0.124 \\
     Edelmann & head & 1730 & -0.037 & 0.111 \\
   Marktpreis &  mod & 1820 & -0.041 & 0.116 \\
   Marktpreis & head & 1900 & -0.044 & 0.107 \\
   Kirchspiel &  mod & 1780 & -0.054 & 0.204 \\
     Gasthaus & head & 1900 & -0.066 & 0.143 \\
   Kirchspiel & head & 1780 & -0.070 & 0.144 \\
    Weibsbild & head & 1760 & -0.074 & 0.287 \\
     Edelmann &  mod & 1890 & -0.078 & 0.199 \\
   Steuermann &  mod & 1830 & -0.087 & 0.198 \\
 Wechselbrief &  mod & 1760 & -0.089 & 0.192 \\
  Mittelstand &  mod & 1900 & -0.091 & 0.273 \\
     Gasthaus &  mod & 1900 & -0.092 & 0.250 \\
 Donnerwetter &  mod & 1740 & -0.144 & 0.310 \\
 Donnerwetter & head & 1740 & -0.232 & 0.594 \\
 \bottomrule
\end{tabular}
  \caption{Aggregate $\Delta$ annotated compositionality per decade covered by a $\langle$compound, constituent, inflection-point$\rangle$ triple. Avg.\ change between decades within each inflection-point. Sorted by $\Delta$, from largest increase in compositionality to largest decrease in compositionality. --- German annotation.}\label{tab:de-anno-trends}
\end{table}

\begin{table}[h]
  \footnotesize
  \begin{tabular}{lllrr} \toprule
    Compound & Const.\ & Infl.\ & $\Delta$ & avg.\  \\\midrule
       leisure hour & head & 1830 &  0.111 & 0.206 \\
       country girl & head & 1820 &  0.097 & 0.175 \\
           baby boy &  mod & 1990 &  0.086 & 0.144 \\
       leisure time & head & 1990 &  0.086 & 0.209 \\
         school day & head & 1990 &  0.085 & 0.339 \\
     reading public &  mod & 1840 &  0.081 & 0.285 \\
       fam.\ phys.\ &  mod & 1870 &  0.073 & 0.220 \\
         trust fund &  mod & 1930 &  0.072 & 0.361 \\
     reading public & head & 1840 &  0.060 & 0.230 \\
         trust fund & head & 1930 &  0.040 & 0.090 \\
          work hour &  mod & 1930 &  0.036 & 0.110 \\
       leisure time &  mod & 1860 &  0.034 & 0.114 \\
         rebel army & head & 1870 &  0.034 & 0.130 \\
         school day &  mod & 1830 &  0.034 & 0.293 \\
       bosom friend &  mod & 1850 &  0.031 & 0.113 \\
          work hour & head & 1930 &  0.029 & 0.116 \\
        frame house & head & 1980 &  0.022 & 0.136 \\
        church bell & head & 1940 &  0.015 & 0.112 \\
       servant girl &  mod & 1860 &  0.015 & 0.193 \\
        church bell &  mod & 1940 &  0.009 & 0.137 \\
       peasant girl & head & 1850 &  0.009 & 0.216 \\
       fam.\ phys.\ & head & 1970 &  0.005 & 0.100 \\
      uni.\ student & head & 1960 &  0.005 & 0.069 \\
           silk hat & head & 1890 &  0.003 & 0.093 \\
         coffee pot & head & 1900 &  0.002 & 0.134 \\
         rebel army &  mod & 1890 &  0.002 & 0.159 \\
         rebel army &  mod & 1870 &  0.001 & 0.194 \\
        cherry tree &  mod & 1910 &  0.000 & 0.120 \\
        cherry tree & head & 1910 &  0.000 & 0.098 \\
       bosom friend & head & 1850 & -0.001 & 0.101 \\
           silk hat &  mod & 1930 & -0.001 & 0.128 \\
       country girl &  mod & 1820 & -0.002 & 0.203 \\
        lady friend &  mod & 1870 & -0.002 & 0.103 \\
       leisure time & head & 1860 & -0.004 & 0.207 \\
      uni.\ student &  mod & 1960 & -0.004 & 0.151 \\
          gold ring & head & 1860 & -0.006 & 0.171 \\
         school day &  mod & 1990 & -0.006 & 0.196 \\
           silk hat & head & 1930 & -0.006 & 0.131 \\
          ice water & head & 1870 & -0.009 & 0.169 \\
    bus.\ community & head & 1960 & -0.010 & 0.092 \\
    \bottomrule
  \end{tabular}
  \begin{tabular}{lllrr} \toprule
    Compound & Const.\ & Infl.\ & $\Delta$ & avg.\ \\\midrule
leisure time &  mod & 1990 & -0.011 & 0.305 \\
      freight train &  mod & 1900 & -0.014 & 0.277 \\
       fam.\ phys.\ & head & 1870 & -0.017 & 0.029 \\
      freight train & head & 1900 & -0.017 & 0.069 \\
          iron door & head & 1860 & -0.017 & 0.088 \\
       peasant girl &  mod & 1850 & -0.017 & 0.182 \\
       servant girl & head & 1860 & -0.017 & 0.128 \\
    bus.\ community &  mod & 1960 & -0.020 & 0.148 \\
         trust fund &  mod & 1980 & -0.022 & 0.215 \\
          work hour &  mod & 1900 & -0.023 & 0.208 \\
        frame house & head & 1880 & -0.025 & 0.170 \\
        lady friend & head & 1870 & -0.027 & 0.149 \\
          gold ring &  mod & 1860 & -0.029 & 0.187 \\
         trust fund & head & 1980 & -0.029 & 0.090 \\
          work hour & head & 1900 & -0.029 & 0.213 \\
           silk hat &  mod & 1890 & -0.034 & 0.259 \\
          ice water &  mod & 1870 & -0.045 & 0.367 \\
           baby boy & head & 1900 & -0.047 & 0.092 \\
       bosom friend & head & 1820 & -0.049 & 0.074 \\
       leisure hour &  mod & 1830 & -0.051 & 0.353 \\
        frame house &  mod & 1880 & -0.053 & 0.215 \\
          ice water &  mod & 1890 & -0.054 & 0.439 \\
          iron door &  mod & 1860 & -0.057 & 0.251 \\
          ice water & head & 1890 & -0.058 & 0.144 \\
          iron door & head & 1840 & -0.061 & 0.119 \\
        frame house &  mod & 1980 & -0.063 & 0.587 \\
           baby boy & head & 1990 & -0.065 & 0.189 \\
       fam.\ phys.\ &  mod & 1970 & -0.068 & 0.187 \\
         sell price & head & 1930 & -0.068 & 0.105 \\
      freight train &  mod & 1950 & -0.070 & 0.297 \\
         rebel army & head & 1890 & -0.087 & 0.117 \\
       bosom friend &  mod & 1820 & -0.088 & 0.186 \\
         coffee pot &  mod & 1900 & -0.089 & 0.139 \\
         sell price &  mod & 1930 & -0.100 & 0.125 \\
      freight train & head & 1950 & -0.101 & 0.157 \\
           baby boy &  mod & 1900 & -0.112 & 0.429 \\
        lady friend & head & 1850 & -0.117 & 0.309 \\
          iron door &  mod & 1840 & -0.150 & 0.261 \\
        lady friend &  mod & 1850 & -0.160 & 0.250 \\
         school day & head & 1830 & -0.160 & 0.240 \\
         \bottomrule
  \end{tabular}
  \caption{Aggregate $\Delta$ annotated compositionality per decade covered by a $\langle$compound, constituent, inflection-point$\rangle$ triple. Avg.\ change between decades within each inflection-point. Sorted by $\Delta$, from largest increase in compositionality to largest decrease in compositionality. --- English annotation.}\label{tab:en-anno-trends}
\end{table}

\section{Experiments}\label{ch8:sec:exp-setup}
In this section, we introduce the mechanisms by which we predict per-decade compositionality for each target compound (for the decades included within its 1--2 inflection points), and combine those per-decade predictions into an overall estimate of the change in compositionality across the decades surrounding an inflection point.
The overall flatness of the trends found in our target set (described above in Section~\ref{ch8:subsec:annotation-results}) makes this a challenging experimental scenario.

First, in Subsection~\ref{ch8:subsec:representations}, we detail the five types of distributional representations (of varying construction and complexity): Modern BERT, second-order random indexing representations, Word2Vec models (continuous bag of words and skipgram variants), and ranked lists of words associated with each target according to Pointwise Mutual Information scores.
This final representation type is only used as a follow-up qualitative validation method.
Then, in Subsection~\ref{ch8:subsec:comp-measure}, we show how the various representation types are used to measure per-decade compositionality, detailing the experimental variables of which temporal granularity is afforded to the models, and which measures are used to compare representations to produce compositionality predictions. Finally, we show how these per-decade predictions are combined into an overall trend or change in compositionality over time.
These three experimental variables: the model type, the temporal window schedule, and the compositionality measure collectively allow us to address \textbf{RQ2}, asking whether narrow temporal granularity and contextual representations are beneficial in the CTP task.

Experimental results are reported in Section~\ref{ch8:results}, divided into three parts: Section~\ref{ch8:lr-scores} compares the alignment between predicted compositionality trends and those obtained by annotation, Section~\ref{ch8:sec:trend-prediction} compares rankings of targets by their degree of change between model predictions and the annotated rankings. Finally, Section~\ref{ch8:sec:pmi-results} presents a more qualitative analysis of word lists associated with each target in each decade surrounding its inflection point(s).

\subsection{Representations}\label{ch8:subsec:representations}

Here, we describe the five types of semantic representations used in this study.
Separate models are trained using each possible 1-decade up to 5-decade time slice of either of the two diachronic corpora.
In total, 115 models of each type were trained over the possible windows in the German data set (from 1680--1920), and 90 models of each type were trained over the windows in the English dataset (1810--2000).

The modern BERT and second-order random indexing representations are the prototypical representation types out of the five possible representations because they are both \textbf{contextualized} representations from which we can obtain a unique representation for each context that a target compound or constituent occurs in.
The two Word2Vec model types and the PMI representations, on the other hand are \textbf{static} representations --- each model produces a single representation for a particular target compound or constituent.
These model types were selected to enable a comparison between models of different complexity (in terms of the total number of parameters, as well as the way that local context from training examples is used) across many `small data' training scenarios.

\paragraph{Modern BERT}
We trained models from scratch, using unsupervised Masked-Language Modeling training over all possible 5-decade wide temporal windows of either diachronic corpus.
We used the Modern BERT architecture~\citep{modernbert} with the same configuration as~\citet{ehrmanntraut2025moderngbertgermanonly1bencoder} for both languages (the main effect of which was to use a smaller vocabulary size)\footnote{The reference model, downloaded from Huggingface (but only used for its configuration details) is available at \url{https://huggingface.co/LSX-UniWue/ModernGBERT_1B}}.
We decided to substitute this model architecture for the original BERT mainly to take advantage of its purported training and inference efficiency improvements (as we planned to train many models from scratch), and because it uses an encoding of relative position, rather than BERT's encoding of the absolute position of each token in its input (using a method from~\citet{SU2024127063}). One additional consideration was to use `whole-word masking'~\citep{Cui_2021}, meaning that in the random assignment of mask tokens during the training process, if any sub-tokenized token was selected to be masked, the entire sequence of sub-tokens would be masked.
Further details about Modern BERT model training are reported in Appendix~B.


\paragraph{2nd-order random indexing}

\begin{figure}
\centering\includegraphics[width=0.65\textwidth]{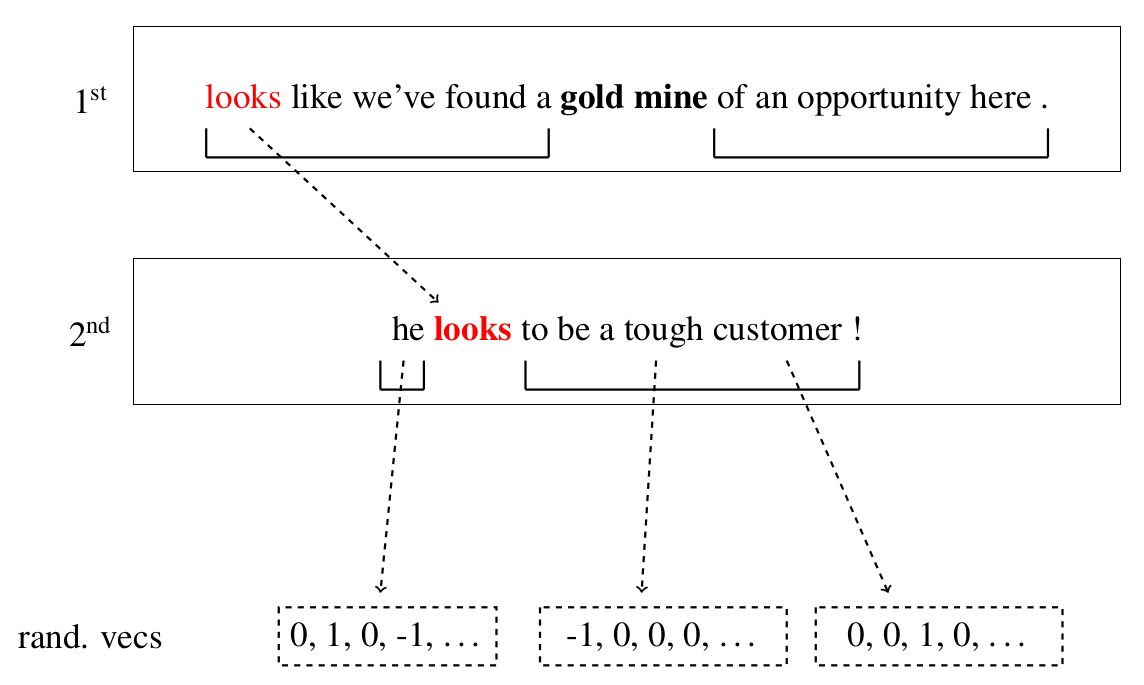}
\caption{Construction of second-order random-index vectors. First-order contexts of target `gold mine' are identified. Contexts of those contexts are represented by a vocabulary of random sparse vectors.}
\label{fig:2nd-order-repr}
\end{figure}

Having been used in our previous work~\citep{jenkins-etal-2025-multi}, derived from~\citet{basile-caputo-semeraro-2015}, 2nd-order random indexing vectors are high-dimensional latent representations, created from a single pass over a corpus, rather than as the result of the optimization of an artificial neural network's parameters through unsupervised learning.
The core idea is to represent every term in a vocabulary with a high dimensional, sparse  vector, where most dimensions are $0$, with a small random selection of dimensions are set to $1$ and $-1$ ($1\%$ each).
Next, we find the set of all words appearing in a context window of 5 space-delimited, lemmatized tokens around each target compound, as used anywhere in the corpus, we refer to these as \textit{first-order contexts} --- represented in the \nth{1} row of Figure~\ref{fig:2nd-order-repr}.
These first-order context lemmas are filtered using a method derived from~\citet{schutze-1998-automatic}\footnote{We first filter the context set using a per-language stopword list from NLTK~\citep{bird-loper-2004-nltk},
then select the $1,500$ contexts most dependent on the presence of a target compound, using a $\chi^{2}$ based criterion,
and another $1,500$ lemmas selected by (highest) frequency in the set of first-order contexts.}.
Each first-order context term from the filtered list has a representation created for it by again searching the full corpus to find a 5-lemma window on either side of each occurrence of the first-order term (\nth{2} row of Figure\ref{fig:2nd-order-repr}).
We then reach the terminal case of our recurrence, the \nth{3} row of Figure~\ref{fig:2nd-order-repr}, where each second-order context is represented by the random-index vector corresponding to it.
Having arrived at the last row of Figure~\ref{fig:2nd-order-repr}, we aggregate the results back to the \nth{1} row of the diagram.

For example, the sum of the context terms around `looks' in the \nth{2} row of Figure~\ref{fig:2nd-order-repr} yield one \nth{2}-order representation --- the average of all such representations of `looks' gives us its \nth{1}-order representation.
The average\footnote{Taking averages results in real-valued vectors, though the base vocabulary vectors are integers.}
of all of the \nth{1}-order representations in the window around `gold mine' in the \nth{1} row of Figure~\ref{fig:2nd-order-repr} (averaging the \nth{1}-order representations of `looks', `found', `opportunity', etc.) finally yields a contextualized representation of `gold mine'.
In rare cases where the target term (here, `gold mine') has no first-order contexts (since these are filtered to a maximum size, to economize on overall storage space), its representation is formed by averaging the random-indexing vectors corresponding to the target compound itself.
These representations are created with respect to each up-to 5-decade long segment of the corpora.

\paragraph{Word2Vec Representations --- continuous bag of words (cbow) and skipgram with negative sampling (skipgram)}

We also use Word2Vec distributional representations~\citep{https://doi.org/10.48550/arxiv.1301.3781} to provide a simpler comparison to the more complex and time-consuming modern BERT and 2nd-order random indexing representations.

Similar to the other representations, we created separate models for each up-to-5-decade span of the diachronic corpora.
We used default settings for the \texttt{gensim} library's implementation of Word2Vec~\citep{rehurek_lrec}, other than setting the minimum number of occurrences for a term to be included in the vocabulary at $1$ (by default, the dimensionality was set to 100, the models were trained for 5 epochs, with no maximum vocabulary size).
Separate models were created for the two Word2Vec architectures: cbow and skipgram.

All models were trained on the full contents of a particular time-slice of one of the diachronic corpora, and systematically converted all variant forms of the target compounds to a contiguous unit (for German: the closed compound form, for English, an artificial closed-compound form created by joining the two open compound halves with an underscore character\footnote{i.e.\ \textit{bosom friend} becomes \textit{bosom\_friend}.}).

\paragraph{PMI Associations}
Pointwise Mutual Information and related information theoretic measures have a long history of use~\citep{church-hanks-1990-word} for identifying terms in a corpus that are strongly associated.
Inspired by~\citet{regan2023semantic}, we create a set of statistical associations with our target noun compounds that do not rely on vector-space comparisons for their interpretation.
For each target's inflection points, we calculate a (modified) pointwise mutual information score (Equation~\ref{eq:mpmi}, per~\citet{regan2023semantic}) to produce a ranking of the top 20 most associated (unigram) terms to the target compound, given the frequency of the target in the temporal window used, along with unigram statistics for that window of the corpus.
We use a value of $\alpha$ of $0.75$, to balance between extreme high and low frequency events dominating the rankings, and additionally exclude items from the top 20 associated terms that occur fewer than 3 times in the given window of the corpus, which helps to remove noise artifacts that are highly associated with a target.
Context for these representations is defined as a linear symmetric window of 5 space-delimited tokens from the target compound.

\begin{equation}\label{eq:mpmi}
mPMI(x,y) = \frac{P(x,y)}{(P(x)P(y))^{\alpha}}
\end{equation}

\subsection{Measuring Compositionality Over Time}\label{ch8:subsec:comp-measure}
This subsection describes how we use the various representations introduced above to make predictions of the compositionality of a target noun compound and one of its constituents at a particular decade, and how multiple such predictions can be combined to produce an aggregate prediction of compositionality across a longer span of time.


\paragraph{Varying Window Sizes}
As mentioned above, all models are trained on a limited temporal window of the diachronic corpora. In total, we trained a model for each possible 1-decade, 2-decade, up to 5-decade window.
One of our goals is to be as faithful as possible with our models to the historical contexts that our examples occurred in.
As such, we experiment with variations in expanding or sliding a temporal window over the corpora, and using these shifting windows at inference time as we calculate trends in compositionality.
Trends are simply the slope of the line formed by several compositionality predictions plotted over time, representing the overall change in compositionality.

The three configurations of \textbf{temporal window schedules} are \texttt{full}, \texttt{decades}, and \texttt{sliding-full}.
These settings determine which time-delimited model(s) (always of the same representation type) is/are used in a given experimental run.
Figure~\ref{fig:window-schedules} shows an example of models (represented by temporal ranges of their training data in square brackets) that are utilized for an example inflection point, varying between the three temporal window schedules.

The first configuration, \texttt{full} is the simplest temporal window schedule. It consists of a single model or representational space, exposed to an entire 5-decade window of the corpus. Per-decade compositionality scores are calculated with each decade's examples being progressively added into the algorithm.
This temporal schedule is not compatible with static models, because they can only encode one representation for a term, regardless of how many decades were used in the training data. 
The second configuration, \texttt{decades} takes the opposite approach to \texttt{full}, in that each decade's examples are processed by a model that has been trained on only that decade.
The third configuration, \texttt{sliding-full} is a combination of the first two.
It begins with a model trained only on the first decade of the overall temporal window, and each successive decade makes use of a separate model, with each successive model and set of compositionality predictions incorporating data (to the model) and examples to predict from each decade up until that point, up until the final decade, which uses the same model as the \texttt{full} configuration.


\begin{figure}
\begin{center}  Example Inflection Point: \textbf{1960}
\begin{itemize}
  \item \texttt{full:}  [1940--1980] $\rightarrow$ trend
  \item \texttt{decades:}  [1940], [1950], [1960], [1970], [1980] $\rightarrow$ trend
  \item \texttt{sliding-full:}  [1940], [1940--50], [1940--60], [1940--70], [1940--80] $\rightarrow$ trend
\end{itemize}
\end{center}
\caption{The three configurations of temporal window schedules. Each time period in square brackets represents a separate model, and the training data range that it was exposed to.}
\label{fig:window-schedules}
\end{figure}

\paragraph{Embedding Similarity}

We use three `system-internal' measures to compare vector representations.
Here, we are always comparing representations of target compounds against representations of one of that compound's constituents, and we use the outcome of each as a measure of compositionality.
All three measures range from $0.0$ to $1.0$, where $0.0$ represents the most compositional prediction, and $1.0$ represents the least compositional prediction.

Two types of comparison operate directly on the vector representations: \texttt{PRT} (prototype) and \texttt{APD} (average pairwise difference)~\citep{giulianelli-etal-2020-analysing, martinc-etal-2020-leveraging}.
The items are compared either by first averaging all members of each comparison group and then taking the cosine distance between the two averages (PRT) or by averaging the cosine distances between all possible pairings between the two groups (APD).

The last measure: JSD (Jensen-Shannon Divergence) first requires a clustering step (detailed below), in which the representations (representations of both the target compound and one of its constituents) are grouped together.
After the clustering is performed, we calculate the JSD to summarize how similarly distributed among all the clusters the compound and the constituent examples are.
Because the Word2Vec representations are static, it is not possible to cluster them, so they are not evaluated with the JSD similarity measure.

\paragraph{Incremental Clustering}
We used the FISHDBC algorithm~\citep{dellamico2019fishdbcflexibleincrementalscalable} to cluster the modern BERT and 2nd-order random-indexing representations.\footnote{We used the following non-default parameters: setting the dissimilarity function to cosine distance, and setting the minimum number of samples (used by the underlying hdbscan algorithm) to 2.}
It is an extension of the HDBSCAN algorithm, which is itself an extension of the DBSCAN algorithm\footnote{See \url{https://scikit-learn.org/stable/modules/clustering.html\#hdbscan}}.
The clustering criterion shared by these algorithms is that they are density-based, which means that clusters should have a large number of items within a given amount of space, and be separated by spaces with a lower number of items per unit of space.
Hierarchical clustering finds clusters within other clusters, in this case, finding a relatively more dense region within a sparser region, which helps the algorithm to cope with clusters of varying densities.
The last part of FISHDBC to clarify is that it is incremental, meaning that additional items can be added to an existing cluster assignment, without re-calculating everything from scratch.
We took inspiration from the theoretical contribution of~\citet{periti-tahmasebi-2024-towards} to experiment with clustering over consecutive time periods --- who argue that the `faithfulness' property of incremental clustering algorithms may be more suitable in semantic change settings to avoid detecting real differences between corpus time-slices that are extraneous to semantic change detection (i.e.\ the idea of overfitting a dataset).

The way that incremental clustering is performed depends on the temporal window schedule (again, refer to Figure~\ref{fig:window-schedules}).
For \texttt{full}, one model is used, but successive predictions are obtained by adding examples from successive decades. 
In the example in Figure~\ref{fig:window-schedules}, we would first cluster examples from the 1940s, obtain a compositionality prediction for the 1940s from the JSD of that set of clusters, then we would add examples from the 1950s to the existing clusters, and obtain a prediction for the 1950s from the JSD of that larger set of clusters, continuing until we added examples from all five decades, and obtained a prediction for the final decade, the 1980s.
The \texttt{decades} temporal window schedule does not involve any incremental clustering, because representations for each decade are obtained from separate models.
The same FISHDBC algorithm is used, but for one round of clustering for each decade.
Finally, the \texttt{sliding-full} temporal window schedule uses representations from a different model for each decade's prediction, but incremental clustering is performed whenever a model was trained on multiple decades, in order to obtain a prediction for the final decade covered by the model.
In the example from Figure~\ref{fig:window-schedules}, this means first a single iteration of clustering would be run on the model trained only on the 1940s, to obtain a prediction (via JSD) of the compositionality for that decade, the prediction for the 1950s would be obtained by running two iterations of clustering on the model trained on the 1940s--1950s, first clustering examples from the 1940s, then adding examples from the 1950s to a second iteration of clustering, and making a compositionality prediction for the 1950s based on that, continuing in this way until the final model (the same one used in the \texttt{full} temporal window schedule) is reached.

\paragraph{Change Over Time}
We represent the overall change over the period of time surrounding an inflection point with the average gradient of the vector of each each individual compositionality prediction.
To compare each trend prediction against the per-decade compositionality scores obtained from the annotation, however, it is more convenient to fit a least squares linear regression model\footnote{Using the Scikit-learn implementation (version 1.2.2): \url{https://scikit-learn.org/stable/modules/generated/sklearn.linear_model.LinearRegression.html}} to obtain an $R^{2}$ score of how well the predicted sequence of per-decade compositionality scores fits the annotated compositionality ratings --- we refer to this method as an \textbf{alignment of compositionality prediction}.
This also facilitates the comparison of trends that operate on different scales: the predictions are always between $0.0$--$1.0$ and the gold scores are always from $0.0$--$5.0$.
Results for this evaluation are reported below in Section~\ref{ch8:lr-scores}.

The second method of evaluating each configuration's prediction of compounds that have changed over time, is to rank each $\langle$compound, constituent, inflection-point$\rangle$ triple by how much it was predicted to have changed, and to correlate this against the compositionality $\Delta$ values obtained in the annotation. We refer to this as \textbf{compositionality trend prediction}, with results reported below in Section~\ref{ch8:sec:trend-prediction}.

\paragraph{PMI-Ranked Association Lists}
Interpreting the contents of the PMI-ranked association lists is a qualitative exercise.
Some terms included in the list are obvious noise (numbers, single letters), but for others it is possible to see thematic coherence between the associated terms and their target compound.
There is always a risk with this kind of interpretation of telling too grandiose a story from the decontextualized association terms.
The method has the advantage that any such claims can be verified by searching a time slice of the corpora for a the presence of both the target compound and a particular association term.
The obvious downside is that such verification is highly labor-intensive, with the effort expended to check a list of associations greatly exceeding the effort to create the list.
We aim to highlight clear changes in the thematic contents of the presented word association lists over time without straining the credulity of the reader.

\subsection{Results}\label{ch8:results}

\paragraph{Compositionality Trend, All Items} 
We run experiments to calculate per-decade compositionality predictions (and the resulting trend over time) for each $\langle$target compound, constituent, inflection point$\rangle$ triple individually.
Each triple has per-decade compositionality calculated using one of three \textbf{model types}: \texttt{modern-bert}, \texttt{2nd-order}, and \texttt{word2vec}.
The second set of experimental variables is the \textbf{temporal window schedule}: \texttt{decades}, \texttt{sliding-full}, or \texttt{full} which determines the temporal granularity of the model or models used to calculate per-decade compositionality predictions.
The final set of experimental variables is the \textbf{similarity measure}: one of \texttt{PRT}, \texttt{APD}, or \texttt{JSD}, with \texttt{APD} and \texttt{JSD} not being used by the \texttt{word2vec} models, as there are no pairs of individual examples and no clustering is performed because the representations derived from the \texttt{word2vec} models are already aggregated at the type-level.

The first set of evaluations is in Section~\ref{ch8:lr-scores} presents each experimental configuration in terms of its alignment with decade-by-decade compositionality ratings from annotators (which, though relatively flat in the aggregate, show decade-by-decade changes).
The second evaluation is reported in Section~\ref{ch8:sec:trend-prediction}, where we rank the $\langle$compound, constituent, inflection point$\rangle$ triples by their predicted change in compositionality for each experimental configuration, and correlate these rankings with the ranked annotation results (as displayed in Tables~\ref{tab:de-anno-trends} and~\ref{tab:en-anno-trends}). This evaluation abstracts over all of the individual per-decade changes, and measures only the end result: an overall prediction of the change in compositionality.
The last evaluation is shown in Section~\ref{ch8:sec:pmi-results}, and provides an alternative, more qualitative perspective on the preceding two evaluations, using ranked word association lists from a simpler PMI model to characterize the semantic shifts predicted by the annotation results and by the strongest performing models according to the previous two evaluations.

\subsubsection{Alignment of Compositionality Predictions}\label{ch8:lr-scores}

The first way that we can compare the model configurations is by the extent to which they align with the decade-by-decade compositionality ratings of the annotators, as determined by fitting a linear regression model\footnote{Using the scikit-learn \texttt{LinearRegression} method. \url{https://scikit-learn.org/stable/modules/generated/sklearn.linear_model.LinearRegression.html}} on each model's compositionality predictions and the annotators' ratings (a separate fit for each $\langle$compound, constituent, inflection-point$\rangle$ triple), and obtaining the $R^2$ score for each fit, which serves as an estimate for the goodness of fit between the sequences of compositionality ratings and compositionality trend predictions.

Each configuration is listed in Tables~\ref{tab:ch8:de-avg-lr-repr-schedule} (German) and~\ref{tab:ch8:en-avg-lr-repr-schedule} (English), sorted in descending order by average $R^2$ score (from best to worst).
Recall that each target compound's inflection points are evaluated separately, which is why we aggregate the results here as average linear regression scores.
The maximum average $R^2$ score for the German results was $0.326$ $(\pm 0.304)$, and the minimum $0.145$ $(\pm 0.218)$.
The maximum average $R^2$ score for the English results was $0.363$ $(\pm 0.290)$, and the minimum was $0.098$ $(\pm 0.212)$.
The range of scores allows us to draw distinctions between the best and worst configurations.
Overall, across both languages, we can observe that the models that used incremental-clustering (indicated by the \texttt{JSD} measure) had lower scores than those that used the \texttt{PRT} or \texttt{APD} measures (the clustering of \texttt{modern-bert} embeddings constitute the worst configurations in both languages), and that the overall best performing configurations did not use the \texttt{full} temporal window schedule.
In all cases the magnitude of the standard deviation of $R^2$ scores is high.
Among the systems with the best average $R^2$ scores, large standard deviations show us that some targets' compositionality ratings were predicted rather closely to how they were annotated, and others were not well-predicted at all.

%
\begin{table}\centering\footnotesize
  \begin{tabular}{lllrr}\toprule
    Representation & Schedule & Measure &  $\mu$ $R^2$ & $\sigma$ $R^2$ \\ \midrule
    2nd-order &    decades      &   jsd    &        0.326   &     0.304 \\
    skipgram  &    decades      &   prt    &        0.324   &     0.292 \\
    2nd-order &    sliding-full &   apd    &        0.315   &     0.293 \\
    modern-bert      &    full         &   prt    &        0.313   &     0.247 \\
    modern-bert      &    full         &   apd    &        0.303   &     0.273 \\
    modern-bert      &    sliding-full &   prt    &        0.285   &     0.226 \\
    cbow      &    sliding-full &   prt    &        0.284   &     0.274 \\
    2nd-order &    sliding-full &   jsd    &        0.283   &     0.257 \\
    skipgram  &    sliding-full &   prt    &        0.280   &     0.264 \\
    2nd-order &    decades      &   apd    &        0.278   &     0.280 \\
    2nd-order &    decades      &   prt    &        0.264   &     0.279 \\
    modern-bert      &    sliding-full &   apd    &        0.264   &     0.227 \\
    2nd-order &    sliding-full &   prt    &        0.263   &     0.268 \\
    cbow      &    decades      &   prt    &        0.263   &     0.304 \\
    modern-bert      &    decades      &   prt    &        0.256   &     0.259 \\
    2nd-order &    full         &   apd    &        0.255   &     0.267 \\
    2nd-order &    full         &   jsd    &        0.254   &     0.237 \\
    modern-bert      &    decades      &   apd    &        0.246   &     0.262 \\
    modern-bert      &    decades      &   jsd    &        0.229   &     0.289 \\
    2nd-order &    full         &   prt    &        0.217   &     0.247 \\
    modern-bert      &    sliding-full &   jsd    &        0.156   &     0.239 \\
    modern-bert      &    full         &   jsd    &        0.145   &     0.218 \\\bottomrule
\end{tabular}
  \caption{Average and standard deviation $R^2$ scores of model compositionality predictions and annotated compositionality ratings: German / DTA dataset.}\label{tab:ch8:de-avg-lr-repr-schedule}
\end{table}

\begin{table}\centering\footnotesize
  \begin{tabular}{lllrr}\toprule
    Representation & Schedule & Measure & $\mu$ $R^2$ & $\sigma$ $R^2$ \\ \midrule
modern-bert       &  decades      &    apd &    0.363 &   0.290 \\
modern-bert       &  decades      &    prt &    0.363 &   0.279 \\
cbow       &  sliding-full &    prt &    0.336 &   0.282 \\
skipgram   &  sliding-full &    prt &    0.326 &   0.319 \\
modern-bert       &  sliding-full &    prt &    0.324 &   0.284 \\
modern-bert       &  sliding-full &    apd &    0.312 &   0.284 \\
modern-bert       &  full         &    prt &    0.310 &   0.279 \\
2nd-order  &  decades      &    prt &    0.295 &   0.268 \\
cbow       &  decades      &    prt &    0.294 &   0.312 \\
skipgram   &  decades      &    prt &    0.293 &   0.291 \\
2nd-order  &  decades      &    apd &    0.288 &   0.252 \\
modern-bert       &  full         &    apd &    0.282 &   0.276 \\
2nd-order  &  full         &    prt &    0.259 &   0.272 \\
2nd-order  &  sliding-full &    apd &    0.246 &   0.259 \\
2nd-order  &  sliding-full &    jsd &    0.243 &   0.244 \\
2nd-order  &  full         &    apd &    0.239 &   0.238 \\
2nd-order  &  full         &    jsd &    0.230 &   0.222 \\
2nd-order  &  decades      &    jsd &    0.229 &   0.236 \\
2nd-order  &  sliding-full &    prt &    0.204 &   0.229 \\
modern-bert       &  decades      &    jsd &    0.195 &   0.278 \\
modern-bert       &  sliding-full &    jsd &    0.114 &   0.218 \\
modern-bert       &  full         &    jsd &    0.098 &   0.212 \\\bottomrule
  \end{tabular}
  \caption{Average and standard deviation $R^2$ scores of model compositionality predictions and annotated compositionality ratings: English / COHA dataset.}\label{tab:ch8:en-avg-lr-repr-schedule}
\end{table}

\paragraph{German Results}
Looking specifically at the German models in Table~\ref{tab:ch8:de-avg-lr-repr-schedule}, we can identify a result particular to this language, which is that the best-scoring configuration used in incremental-clustering: $\langle$\texttt{2nd-order}, \texttt{decades}, \texttt{jsd}$\rangle$.
It is interesting that the clustering method was most successful along with the German 2nd-order random indexing representations, as neither configuration was successful in the English experiments.

We fit linear models\footnote{Using the \texttt{lm} method from the standard R stats package.} to the $R^2$ scores, with respect to either the representation, temporal window schedule, or measure, but these yielded no statistically significant differences.
This seems to reflect what we can see from looking at Table~\ref{tab:ch8:de-avg-lr-repr-schedule}, where all representation types, schedule types and measure types occur in both high and low places in the overall rankings, and is unlike the clear differences that we will see shortly when examining the corresponding English results.

As a consequence of not finding any overall effects on $R^2$ score and the model configurations, we performed a few additional Welch Two Sample t-tests between the top scoring configuration and configurations that differed by one setting (i.e.\ swapping \texttt{2nd-order} for \texttt{modern-bert}, or \texttt{jsd} for \texttt{prt}) with respect to their model---annotation $R^2$ scores.
We performed these tests using the next three highest scoring configurations: $\langle$\texttt{skipgram}, \texttt{decades}, \texttt{prt}$\rangle$, $\langle$\texttt{2nd-order}, \texttt{sliding-full}, \texttt{apd}$\rangle$, and $\langle$\texttt{modern-bert}, \texttt{full}, \texttt{prt}$\rangle$.
Of these, only $\langle$\texttt{modern-bert}, \texttt{full}, \texttt{prt}$\rangle$ contrasted significantly with two configurations differing by one setting: changing to \texttt{2nd-order}: ($t = 2.1393, df = 118, \text{p-value} = 0.0344$), and changing to \texttt{jsd}: ($t = 3.9422, df = 116.24, \text{p-value} = 0.0001$).

\paragraph{English Results}
Turning to the English model results in Table~\ref{tab:ch8:en-avg-lr-repr-schedule}, again we make broad observations before examining the top scoring models.
Overall, models using \texttt{jsd} score systematically more poorly.
Models using \texttt{full} schedules score lower than the other two schedule types.
The best two English models have the same mean model---annotation $R^2$ score of $0.363$ (somewhat higher than any of the best German model scores). These are \texttt{modern-bert} models, using \texttt{decades} and \texttt{prt} and \texttt{apd} respectively.
The third-place configuration is $\langle$\texttt{cbow, sliding-full, prt}$\rangle$, scoring $0.336$.

Here, there are more clearly observable differences with respect to the model configurations.
We fit linear models to the annotation/model $R^2$ scores with respect to either the representation, schedule or measure, shown in Tables~\ref{tab:lm-repr},~\ref{tab:lm-schedule}, and~\ref{tab:lm-measure} respectively.
In Table~\ref{tab:lm-repr}, which takes the \texttt{2nd-order} representation as a baseline, there is a significant (via t-test) improvement to switch to any other representation type. The improvement for the Word2Vec models is slightly larger than the \texttt{modern-bert} model, but with a higher standard errors and a less significant t-test statistic.
In Table~\ref{tab:lm-schedule}, which takes the \texttt{decades} schedule as a baseline, scores decrease by switching to either \texttt{full} or \texttt{sliding-full}, but only the difference between \texttt{decades} and \texttt{full} is significant.
This supports the initial intuition behind the \texttt{decades} schedule, to restrict the model to access examples from the same temporal granularity that is is evaluated at.
Finally, in Table~\ref{tab:lm-measure}, we have the \texttt{apd} measure as a baseline, and see a significant decrease in scores between \texttt{apd} and \texttt{jsd}, corresponding with the very low performance we had seen in Table~\ref{tab:ch8:en-avg-lr-repr-schedule}.

\begin{table}
  \centering\footnotesize
  \begin{tabular}{lrrrrl} \toprule 
    Representation & Estimate & Std.\ Error & $t$ value & $Pr(>|t)$ &     \\ \midrule
(Intercept)        & 0.24883  & 0.01032     & 24.111    & < \num{2e-16}   & *** \\
modern-bert               & 0.05997  & 0.01521     & 3.943     & \num{8.4e-05}   & *** \\
cbow               & 0.06615  & 0.02386     & 2.773     & 0.00562   & **  \\
skipgram           & 0.06043  & 0.02386     & 2.533     & 0.01141   & *   \\
\bottomrule
  \end{tabular}
  \caption{Linear model fit to English model---annotation $R^2$ scores, with respect to representation type. Intercept is \texttt{2nd-order}. Signif.\ codes:  0 ‘***’ 0.001 ‘**’ 0.01 ‘*’ 0.05}\label{tab:lm-repr}
\end{table}

\begin{table}
  \centering\footnotesize
  \begin{tabular}{lrrrrl} \toprule 
    Schedule & Estimate & Std.\ Error & $t$ value & $Pr(>|t)$ &     \\ \midrule
(Intercept)  & 0.30270  & 0.01103     & 27.437    & < \num{2e-16}    & *** \\
full         & -0.04224 & 0.01738     & -2.431    & 0.0152    & *   \\
sliding-full & -0.02291 & 0.01582     & -1.449    & 0.1476    &     \\
\bottomrule
  \end{tabular}
  \caption{Linear model fit to English model---annotation $R^2$ scores, with respect to schedule type. Intercept is \texttt{decades}. Signif.\ codes:  0 ‘***’ 0.001 ‘**’ 0.01 ‘*’ 0.05}\label{tab:lm-schedule}
\end{table}

\begin{table}
  \centering\footnotesize
  \begin{tabular}{lrrrrl} \toprule 
    Measure & Estimate & Std.\ Error & $t$ value & $Pr(>|t)$ & \\ \midrule
(Intercept) & 0.28910  & 0.01265     & 22.848    & < \num{2e-16}   & *** \\
jsd         & -0.05049 & 0.01914     & -2.637    & 0.00843   & ** \\
prt         & 0.01177  & 0.01596     & 0.738     & 0.46090   & \\
\bottomrule
  \end{tabular}
  \caption{Linear model fit to English model / annotation $R^2$ scores, with respect to measure type. Intercept is \texttt{apd}. Signif.\ codes:  0 ‘***’ 0.001 ‘**’ 0.01 ‘*’ 0.05}\label{tab:lm-measure}
\end{table}

\paragraph{Overall}
When comparing model trend predictions to annotated trends, there is scarce evidence for German to prefer one configuration over another (though we can say that clustering \texttt{modern-bert} embeddings is a bad idea).
This is in spite of the German annotation appearing to be stronger (with higher levels of agreement).
On the other hand, the larger size of many of the English sub-corpora may have made the \texttt{modern-bert} models more viable in general --- we see that the \texttt{modern-bert} + \texttt{decades} models scored higher for English than their German counterparts.
For English, the additional data made available by \texttt{full} schedules could be detrimental, whereas for German \texttt{modern-bert} models it seems to have been helpful.
In terms of \textbf{RQ2}, across both languages, configurations using narrower temporal window schedules (\texttt{decades} first, then \texttt{sliding-full}) performed somewhat better than configurations with \texttt{full}, generally favoring narrower temporal granularities.
There was less of a clear result favoring contextualized over static representations, with Word2Vec representations scoring highly in the results of both languages.

\subsubsection{Compositionality Trend Prediction Ranking}\label{ch8:sec:trend-prediction}
It is also worth examining, especially without overwhelming evidence to prefer one model over the other from Section~\ref{ch8:lr-scores}, which target compounds are predicted as having the largest changes in compositionality over time.
Rather than measuring the fidelity to each decade's annotated compositionality rating in sequence, we look at each model's ranked predictions for which compounds have changed the most (to be more or less compositional).

We can take each model's set of compositionality trend predictions (see above in Section~\ref{ch8:subsec:comp-measure}, under \textbf{Change Over Time}), one for each $\langle$compound, constituent, inflection point$\rangle$ triple, and correlate them against the ranked list of compositionality $\Delta$s from the annotation results in Section~\ref{ch8:subsec:annotation-results}.
To do so, we use Spearman's $\rho$ ranked correlations (with the alternative hypothesis being a negative correlation).
Recall that the embedding similarity measures (discussed in Section~\ref{ch8:subsec:comp-measure}) range from $0.0$ to $1.0$, where 0 represents the greatest similarity between embeddings, and 1 the greatest difference.
This is the opposite polarity of the compositionality ratings, for which the rating of 0 indicates non-compositionality (a lack of a transparent association between compound and constituent) and 5 represents maximum compositionality, where there is a strong association between the meanings of the compound and the constituent.

\paragraph{German Results}
In Table~\ref{tab:ch8:corr-de} we see the result of the correlation analysis for the German models.
Only three configurations' correlations with the annotation $\Delta$s were significant (with a p-value $< 0.05$): $\langle$\texttt{skipgram}, \texttt{sliding-full}, \texttt{prt}$\rangle$,   $\langle$\texttt{2nd-order}, \texttt{sliding-full}, \texttt{prt}$\rangle$ and $\langle$\texttt{2nd-order}, \texttt{full}, \texttt{prt}$\rangle$.
None of these correlations were strong (all smaller than $\rho = -0.3$).
No configuration scored well in the analysis of decade-by-decade linear regression scores (in Section~\ref{ch8:lr-scores}, Table~\ref{tab:ch8:de-avg-lr-repr-schedule}).
This gives us some additional evidence to prefer the \texttt{skipgram} and \texttt{2nd-order} representations, but makes the relative benefit of any particular temporal window schedule or compositionality measure less clear.

Reasoning that the middle range of annotated compositionality $\Delta$s are likely very difficult to successfully rank,
we repeated the correlation analysis only on the triples with the twenty highest or twenty lowest compositionality $\Delta$s in Table~\ref{tab:de-anno-trends}.
We report these results in Table~\ref{tab:ch8:corr-extremes-de}.
Unsurprisingly, this strengthens the correlations for the configurations with the highest correlations in the analysis over all triples, but now the $\langle$\texttt{2nd-order}, \texttt{decades}, \texttt{prt}$\rangle$ configuration, previously at the threshold of significance, has by a small margin the strongest correlation at $\rho = -0.41$ and a high degree of significance.


\begin{table}
  \centering\footnotesize
\begin{tabular}{lllrr}\toprule
  Representation & Schedule     & Measure & Spearman's $\rho$ & p-value  \\ \midrule
skipgram         & sliding-full & prt     & \textbf{-0.2927}           & 0.0069 \\
2nd-order        & sliding-full & prt     & \textbf{-0.2241}           & 0.0351  \\
2nd-order        & full         & prt     & \textbf{-0.2199}           & 0.0456 \\
2nd-order        & decades      & prt     & -0.2002           & 0.0508 \\
modern-bert             & full         & apd     & -0.1853           & 0.0781 \\
modern-bert             & full         & prt     & -0.1580           & 0.1139  \\
2nd-order        & full         & apd     & -0.1578           & 0.1141 \\
2nd-order        & full         & jsd     & -0.1555           & 0.1177   \\
2nd-order        & sliding-full & apd     & -0.1431           & 0.1257 \\
2nd-order        & decades      & apd     & -0.1287           & 0.1476 \\
modern-bert             & sliding-full & prt     & -0.0976           & 0.2176 \\
modern-bert             & full         & jsd     & -0.0877           & 0.2525 \\
2nd-order        & sliding-full & jsd     & -0.0870           & 0.2435 \\
modern-bert             & sliding-full & apd     & -0.0762           & 0.2714 \\
modern-bert             & decades      & prt     & -0.0513           & 0.3388 \\
modern-bert             & decades      & apd     & -0.0496           & 0.3438 \\
modern-bert             & decades      & jsd     & -0.0288           & 0.4077 \\
modern-bert             & sliding-full & jsd     & 0.0037            & 0.5120 \\
cbow             & sliding-full & prt     & 0.0041            & 0.5134 \\
skipgram         & decades      & prt     & 0.0183            & 0.5600 \\
cbow             & decades      & prt     & 0.0193            & 0.5631 \\
2nd-order        & decades      & jsd     & 0.0600            & 0.6867 \\\bottomrule
\end{tabular}
\caption{Spearman's rho correlation between model trend predictions and compositionality $\Delta$s for each $\langle$compound, constituent, inflection-point$\rangle$ triple. Sorted by rho, from negative (the desired direction of correlation) to positive --- German results.}\label{tab:ch8:corr-de}
\end{table}

\begin{table}
  \centering\footnotesize
\begin{tabular}{lllrr}\toprule
  Representation & Schedule     & Measure & Spearman's $\rho$ & p-value  \\ \midrule
2nd-order        & decades      & prt     & \textbf{-0.4163}           & 0.0046 \\
2nd-order        & decades      & apd     & \textbf{-0.3997}           & 0.0064 \\
skipgram         & sliding-full & prt     & \textbf{-0.3871}           & 0.0067 \\
2nd-order        & sliding-full & prt     & \textbf{-0.3613}           & 0.0139 \\
modern-bert             & decades      & apd     & -0.2465           & 0.0678 \\
modern-bert             & decades      & prt     & -0.2450           & 0.0690 \\
modern-bert             & sliding-full & prt     & -0.2343           & 0.0813 \\
modern-bert             & full         & apd     & -0.2192           & 0.1063 \\
modern-bert             & full         & prt     & -0.2114           & 0.1149 \\
2nd-order        & full         & jsd     & -0.2030           & 0.1246 \\
modern-bert             & sliding-full & apd     & -0.1920           & 0.1273 \\
2nd-order        & full         & prt     & -0.1905           & 0.1401 \\
2nd-order        & sliding-full & apd     & -0.1900           & 0.1299 \\
2nd-order        & full         & apd     & -0.1563           & 0.1886  \\
modern-bert             & full         & jsd     & -0.1561           & 0.1888 \\
2nd-order        & sliding-full & jsd     & -0.1184           & 0.2425 \\
cbow             & sliding-full & prt     & 0.0255            & 0.5621 \\
modern-bert             & decades      & jsd     & 0.0347            & 0.5821 \\
cbow             & decades      & prt     & 0.0440            & 0.6064 \\
skipgram         & decades      & prt     & 0.0718            & 0.6702  \\
2nd-order        & decades      & jsd     & 0.1264            & 0.7752 \\
modern-bert             & sliding-full & jsd     & 0.1499            & 0.8120 \\

 \bottomrule
\end{tabular}
\caption{Spearman's rho correlation between model trend predictions and compositionality $\Delta$s for the top/bottom most changed 20 $\langle$compound, constituent, inflection-point$\rangle$ triples, as determined by annotation compositionality $\Delta$s. Correlations sorted by rho, from negative (the desired direction of correlation) to positive --- German results.}\label{tab:ch8:corr-extremes-de}
\end{table}

\paragraph{English Results}
The results of the correlation analysis in Table~\ref{tab:ch8:corr-en} provide us with little material to make substantive recommendations for any model configuration over another.
No Spearman's $\rho$ correlation was significant (p-value $< 0.05$).
The configurations closest to this threshold of statistical significance used the \texttt{2nd-order} representations.

We also performed a correlation analysis on the triples with the twenty highest or twenty lowest compositionality $\Delta$s in Table~\ref{tab:en-anno-trends}, which we report in Table~\ref{tab:ch8:corr-extremes-en}.
Here we can see several configurations whose compositionality trend predictions correlate at a significant level with the annotations.
The strongest of these are \texttt{modern-bert} configurations with \texttt{decades} schedules and \texttt{apd} and \texttt{prt} measures, with \texttt{2nd-order}, \texttt{decades}, \texttt{apd} close behind.
The stronger correlations of several \texttt{modern-bert} configurations on the triples at the extreme ends of $\Delta$ compositionality, which were not significantly correlated in the German experiments, provide further support for the hypothesis that the overall larger size of the English corpus makes  the \texttt{modern-bert} models more viable.

\begin{table}
  \centering\footnotesize
  \begin{tabular}{lllrr}\toprule
    Representation & Schedule     & Measure & Spearman's $\rho$ & p-value \\ \midrule
2nd-order          & decades      & apd     & -0.1728           & 0.0625  \\
2nd-order          & decades      & prt     & -0.1652           & 0.0714  \\
2nd-order          & full         & prt     & -0.1581           & 0.0861  \\
modern-bert               & decades      & apd     & -0.1481           & 0.0947  \\
2nd-order          & full         & apd     & -0.1343           & 0.1235  \\
modern-bert               & decades      & prt     & -0.1252           & 0.1341  \\
modern-bert               & sliding-full & prt     & -0.1005           & 0.1938  \\
modern-bert               & decades      & jsd     & -0.0848           & 0.2270  \\
modern-bert               & sliding-full & apd     & -0.0733           & 0.2645  \\
modern-bert               & sliding-full & jsd     & -0.0615           & 0.2986  \\
skipgram           & sliding-full & prt     & -0.0417           & 0.3564  \\
cbow               & decades      & prt     & -0.0300           & 0.3957  \\
skipgram           & decades      & prt     & -0.0148           & 0.4481  \\
cbow               & sliding-full & prt     & 0.0015            & 0.5055  \\
modern-bert               & full         & apd     & 0.0180            & 0.5613  \\
modern-bert               & full         & jsd     & 0.0203            & 0.5690  \\
2nd-order          & decades      & jsd     & 0.0318            & 0.6103  \\
2nd-order          & sliding-full & apd     & 0.0432            & 0.6448  \\
2nd-order          & sliding-full & jsd     & 0.0538            & 0.6780  \\
2nd-order          & sliding-full & prt     & 0.0551            & 0.6820  \\
modern-bert               & full         & prt     & 0.0820            & 0.7594  \\
2nd-order          & full         & jsd     & 0.1157            & 0.8403  \\
\bottomrule
  \end{tabular}
\caption{Spearman's rho correlation between model trend predictions and compositionality $\Delta$s for each $\langle$compound, constituent, inflection-point$\rangle$ triple. Sorted by rho, from negative (the desired direction of correlation) to positive --- English results.}\label{tab:ch8:corr-en}
\end{table}

\begin{table}
  \centering\footnotesize
\begin{tabular}{lllrr}\toprule
  Representation & Schedule     & Measure & Spearman's $\rho$ & p-value \\ \midrule
modern-bert             & decades      & apd     & \textbf{-0.3766}           & 0.0082  \\
modern-bert             & decades      & prt     & \textbf{-0.3714}           & 0.0091  \\
2nd-order        & decades      & apd     & \textbf{-0.3642}           & 0.0104  \\
modern-bert             & sliding-full & prt     & \textbf{-0.2971}           & 0.0392  \\
2nd-order        & decades      & prt     & \textbf{-0.2653}           & 0.0489  \\
2nd-order        & full         & prt     & -0.2287           & 0.0898  \\
2nd-order        & full         & apd     & -0.2043           & 0.1160  \\
modern-bert             & sliding-full & apd     & -0.2027           & 0.1178  \\
modern-bert             & decades      & jsd     & -0.1175           & 0.2350  \\
2nd-order        & sliding-full & prt     & -0.0701           & 0.3421  \\
2nd-order        & sliding-full & apd     & -0.0612           & 0.3612  \\
modern-bert             & sliding-full & jsd     & -0.0517           & 0.3822  \\
cbow             & decades      & prt     & -0.0355           & 0.4137  \\
modern-bert             & full         & jsd     & -0.0043           & 0.4900  \\
modern-bert             & full         & apd     & 0.0185            & 0.5427  \\
2nd-order        & decades      & jsd     & 0.0198            & 0.5484  \\
2nd-order        & sliding-full & jsd     & 0.0221            & 0.5509  \\
modern-bert             & full         & prt     & 0.0359            & 0.5823  \\
skipgram         & decades      & prt     & 0.0439            & 0.6060  \\
2nd-order        & full         & jsd     & 0.0720            & 0.6619  \\
cbow             & sliding-full & prt     & 0.1114            & 0.7532  \\
skipgram         & sliding-full & prt     & 0.1551            & 0.8304  \\

 \bottomrule
\end{tabular}
\caption{Spearman's rho correlation between model trend predictions and compositionality $\Delta$s for the top/bottom most changed 20 $\langle$compound, constituent, inflection-point$\rangle$ triples, as determined by annotation compositionality $\Delta$s. Correlations sorted by rho, from negative (the desired direction of correlation) to positive --- English results.}\label{tab:ch8:corr-extremes-en}
\end{table}

\paragraph{Overall}
On the whole, this evaluation perspective favors temporal window schedules that use incremental slices of the data, particularly \texttt{decades}, with \texttt{sliding-full} close behind.
In no case are \texttt{jsd} measures preferred.
The representation type that performed best in this evaluation depended on the language: \texttt{skipgram} and \texttt{2nd-order} were best in the lower-resourced German evaluation, and \texttt{modern-bert} and \texttt{2nd-order} were best in the English evaluation (though only the English evaluation against the most extreme $\langle$compound, constituent, inflection point$\rangle$ triples had any configuration show a significant correlation between its predictions and the annotated compositionality $\Delta$s).
In both languages (meaning also two different magnitudes of data), a simpler representation, \texttt{2nd-order} was the best or competitive with the best configuration.
Referring to \textbf{RQ2}, this evaluation supports the use of fine-grained temporal window schedules, but it is less conclusive on supporting a recommendation on the use of contextualized representations, since a \texttt{skipgram} configuration performed well against the German compositionality $\Delta$s.

\subsubsection{Effect of Frequency}
It is worth considering whether either the evaluation in terms of $R^2$ scores (\ref{ch8:lr-scores}) or in terms of the overall trend predictions (\ref{ch8:sec:trend-prediction}) could be explained as an effect of the overall corpus frequencies of the target compounds, which can be viewed in Tables~\ref{tab:de-target-freq-1}, and~\ref{tab:de-target-freq-2} (German) and Tables~\ref{tab:en-target-freq-1} and~\ref{tab:en-target-freq-2} (English).
We therefore fit linear models (again, making use of the \texttt{lm()} method in R) to the $R^2$ scores with respect to the $\log(\text{total-frequency})$ of the target compounds, as well as model fit to the trend prediction scores with respect to the $\log(\text{total-frequency})$ of the target compounds.

The model fit on the English $R^2$ scores with respect to the $\log$ total frequencies was highly negatively correlated, with an F-statistic of 17.07 and a p-value of \num{7.78e-05}.
The residuals of the fit regression were however very right-skewed.
It should also be noted that the overall range of total frequencies for the English target compounds is not a dramatic range: with a minimum of 199, a mean of 320, and a maximum of 504.

The model fit on the English trend prediction values with respect to the $\log$ total frequencies also showed a significant, negative correlation, with an F-statistic of 4.57 and a p-value of 0.032.
The residuals of the fit regression were normally-distributed.
This says that target compounds predicted to have become less compositional were more frequent, and compounds predicted to have become more compositional were less frequent.

The model fit on the German $R^2$ scores with respect to the $\log$ total frequencies did not show a significant correlation between the two variables. The F-statistic was 1.84 and the p-value was 0.175.
This was also the case with the model fit to the German trend prediction values with respect to the $\log$ total frequencies, where the F-statistic was 0.59 and the p-value 0.442.
The overall frequency distribution of the German target compounds was larger than the English target compounds, with a minimum frequency of 388, a mean of 822, and a maximum of 2612.

Overall, this analysis weakens the English results, and strengthens the German results.

\subsubsection{PMI Associations}\label{ch8:sec:pmi-results}
Finally, we can turn to the word lists generated by PMI models to provide a validation both of the predictions of compositionality change sourced from the models, but also the largest changes that we obtained from the annotation (from Section~\ref{ch8:subsec:annotation-results}).
These tables provide us with a more qualitative way of checking whether semantic changes can be observed in the contexts of the target compounds over several decades.
The word lists are presented in pairs: a table with $\langle$target, constituent, inflection-point$\rangle$ triples that should (according to annotation or model prediction) be increasing in compositionality, and a table with triples which should be decreasing in compositionality.
The PMI-ranked word lists do not, however, model an interrelationship between the compound as a whole and a constituent --- they are only presenting changing context terms for the compound.
As such, the interpretive task is to say whether the changes in PMI-associated terms is indicative of a semantic change towards (positive) or away from (negative) the meaning of the indicated constituent.
Observations from the tables are detailed below, the full tables are given in Appendix~C.

%


%
\paragraph{German Results From Annotation}
The top annotated rating of increased compositionality was \textit{Wechselbrief} (bill of exchange), centered around 1720 (Table~\ref{tab:de-pmi-anno-most-pos}).
We can see a great number of references in the early decades to \textit{akzeptieren}, \textit{indossieren} (accepting, endorsing), and other terms having to do with the mechanics of a bill of exchange (a \textit{Wechselbrief}), whereas in the last two decades, the term is more associated with numbers, units of currency (\textit{Taler}, \textit{Dukaten}), in addition to a few abbreviations that probably also refer directly to units of currency.
This would appear to indicate a progressively more transparent relationship between the letter (\textit{Brief}) and the exchange of money (\textit{Wechsel}).

The top annotated rating of decreased compositionality (and one that we had prior knowledge of) was \textit{Donnerwetter} (stormy weather/damn), centered around 1740 (Table~\ref{tab:de-pmi-anno-most-neg}).
In the early decades we can see several references to literal thunder and lightning and other bad weather (\textit{Blitzen, Blitz, Donnerschlag, Donner-Schlag, Regen, Sturm-Winde} (lightning, lightning, thunder-clap, thunder-clap, rain, storm-wind)).
By the final decade, we still see some reference to weather phenomena (\textit{Sturm} (storm)), but we also see terms that could be more closely associated with the newer sense of \textit{Donnerwetter} as an exclamation: \textit{Teufel} (devil), \textit{Fluch} (curse).
In \textit{Mittelstand} (middle class) we see changes related to the modifier \textit{Mittel} (middle) centered on 1900, but limited to 1880--1910.
Here, we see some basic confirmation of the PMI method by finding terms in the first decade like \textit{fleißig} (diligent), \textit{sebstbewußt} (confident), and \textit{Reichtum} (prosperity), as stereotypical (if flattering) traits of this social group. In the final decade, however, we see terms less associated with describing the \textit{Mittelstand}, but perhaps more about its role in politics: \textit{Wohnungsfürsorge} (housing assistance), \textit{Steuerpolitik} (tax policy), \textit{Wähler} (voter).
Curiously, \textit{Wechselbrief} (bill of exchange) appears among the top compounds with decreased compositionality as well, stretching from the previous interval's end of 1740 up until 1770.
The units of currency listed in the 1740s give way to a more diverse set of contexts through these decades --- by the end, we see words like \textit{Papiergeld} (paper money), \textit{Verstorbene} (deceased), and \textit{Taschenuhr} (pocket watch), and \textit{Erfindung} (invention).
Whether this change has to do more with the ways that \textit{Wechselbrief} was used, or the changing set of documents across this range of time is unclear.

\paragraph{German Results From Best Model ($R^2$ Score)}

Among the top rated triples that increased in compositionality according to the model with the best $R^2$ score (Table~\ref{tab:de-pmi-LR-most-pos}) is \textit{Krankenhaus} (hospital). It is difficult to see exactly why it should be more compositional (in terms of its head) by the 1920s. In the word associations, we see terms in the 1910s that perhaps allude to a more professionalized kind of hospital, with specialized roles like \textit{Sekundararzt} and \textit{Hilfesarzt} (both are a kind of junior doctor), \textit{Verwaltungsdirektor} (administrative director).
\textit{Wechselbrief} appears again in this table (shared with Table~\ref{tab:de-pmi-anno-most-pos}).

Looking over to the triples predicted to have decreased in compositionality according to the model with the best $R^2$ score (Table~\ref{tab:de-pmi-LR-most-neg}), we have two ranges of \textit{Wortspiel} (wordplay) (1730, head), and (1890, modifier).
In the long run, we can see a stronger association with humor emerge by the 1890s and 1900s, where we see associated terms like \textit{Humor} (humor), \textit{Witz} (joke), \textit{witzig} (funny).
Earlier, in the \nth{18} century time-span, we see possible references to kinds of verbal creativity that are potentially distinct from humor: \textit{Sinnbild} (symbol), \textit{Zweideutigkeit} (double meaning), \textit{Dichtkunst} (poetry), but we do still see some reference to humor, with the top association from the 1720s being \textit{Burlesque}.
Turning to \textit{Bauernstand} (peasantry), we can make a similar observation as with \textit{Mittelstand} (middle class), that there is a noticeable increase in political terminology at the beginning of the \nth{20} century.

%

\paragraph{German Results From Best Model (Correlation)}

In Table~\ref{tab:de-pmi-CORR-most-pos} we see the triples predicted as most increasing in compositionality by the best model in terms of the correlation of its trend predictions and annotated compositionality $\Delta$s (for the most extreme of the annotated $\Delta$s).
With \textit{Ehefrau} (wife) and \textit{Handelsmann} (merchant) we see a great deal of given names (female and male respectively), but otherwise it is difficult to interpret any changes.

In Table~\ref{tab:de-pmi-CORR-most-neg}, we see the triples predicted as most decreasing in compositionality.
Here we see the later period of \textit{Wechselbrief} repeated, as well as \textit{Gasthaus} (shared with Table~\ref{tab:de-pmi-anno-most-neg}).
\textit{Marktpreis} has also been seen previously in Table~\ref{tab:de-pmi-LR-most-pos}, albeit with the opposite predicted polarity.
This leaves us with \textit{Kartenspiel} (card game), which is associated throughout with many terms related to gambling and games \textit{Würfel-} (dice-), \textit{Einsatz} (wager), \textit{trumpfen} (to trump) in the early decades, and \textit{Karo} (diamonds --- playing card suit), \textit{Lottospiel} (lottery game), \textit{Schachspiel} (chess) occurring in the last decades.
Any semantic change attested here may be evidenced by terms in the final decade like \textit{verjubeln} (wasting all of one's money), \textit{verflucht} (cursed), and \textit{schrecklich} (awful).
Perhaps these could be attributed to changing attitudes about gambling.
%



\paragraph{English Results From Annotation}

In Table~\ref{tab:en-pmi-anno-most-pos}, we see the triples with the highest positive $\Delta$ compositionality.
Some clear discursive shifts are visible in the entry for \textit{baby boy}, where the 1990s sees \textit{circumcision} (a widespread practice in the United States\index{United States} becoming more polemical at that time), and \textit{autistic} (probably due to the \textit{extremely} polemical and highly discredited\footnote{See for example this 1998 article: \url{https://www.thelancet.com/journals/lancet/article/PIIS0140-6736(97)11096-0/fulltext} --- note the word `retracted' in large red text.} idea that various childhood vaccinations cause autism).
Very few descriptive terms are in the association list for the 1970s, or for the 2000s, while the 1980s has several: \textit{chubby, plump, splendid, fat, healthy}.
With the compound \textit{leisure time} (centered on 1990), another set of emerging discourses is visible, particularly between the 1970s and 1990s.
The associated terms in the 1970s are mainly pleasant: \textit{enrichment, enjoyable, campground, prosperity, luxury}.
Those associated with the 1990s take on a tone that is anxious about or otherwise antithetical to leisure, with associated terms like \textit{redefinition, self-improvement, sedentary, overworked, addiction}, while other terms like \textit{salaryman, tiananmen, chopsticks} could be due to discourses of economic rivalry between the United States\index{United States} and Japan or China.
The 2000s has a mix of terms that could be positive or negative (\textit{bountiful, relaxation, recreation} against \textit{time-consuming, constraint, selfish, commute}).

In Table~\ref{tab:en-pmi-anno-most-neg}, we see the triples with the highest negative $\Delta$ compositionality.
It is difficult to see much of a development over time for \textit{school day}, \textit{lady friend}, or \textit{iron door}.
This may be due to the low number of examples to draw on in these decades (see Table~\ref{tab:en-target-freq-1}).
The last example, however, is more frequent than the other three.
\textit{Baby boy} (1900) is annotated here as having had a decline in compositionality (becoming less related to \textit{baby}). Throughout the range, we see clear association terms for \textit{baby}, like \textit{cradle, bath} (1880s), \textit{adoptive, toddle, cute, chubby, patter} (1890s), \textit{cute, christen, mamma} (1900s), \textit{two-year-old, christening, breast} (1910s), \textit{suck, carriage, weigh, baby, born} (1920s).
In the early decades, we see association terms relating to famous biblical babies (\textit{moses}), and Jesus (via his birthplace \textit{bethlehem}).
In the final decade, association with \textit{half-breed} is perhaps indicative of the eugenics movement of that era.
%
%

\paragraph{English Results From Best Model ($R^2$ Score and Correlation)}
 Table~\ref{tab:en-pmi-LR-most-pos} shows the triples with the highest predicted increase in compositionality, per the model configuration with the best performance in the evaluations from Sections~\ref{ch8:lr-scores} and~\ref{ch8:sec:trend-prediction}.
The first example is \textit{frame house}. We can see some terms associated with military buildings in the 1860s (the decade in which the American Civil War occurred) such as \textit{garrison} and \textit{fort}.
Throughout the range of decades, we see humble terminology: \textit{cabin, shanty, modest-looking, primitive}, alongside more positive terms like \textit{respectable-looking, showy, roomy, substantial, fashionable, neat}. It may be the case that the contexts in which \textit{frame house} occurred in changed during this time period of increasing urbanization, but it is difficult to detect from these association terms.
The example of \textit{school day} (1990) shows more evocative associated terms. The \textit{school day} is associated with some negative terms, possibly related to its structure: \textit{repetitive, unbearable, tiring, frenetic}.
In the 1990s, the term \textit{ad/hd} occurs as an associated term, the acronym for Attention Deficit Hyperactivity Disorder, whose diagnosis and treatment is a source of some amount of controversy, particularly as related to scholastic environments.
In the 2000s, we can see \textit{heterogeneity, compulsory, and bullying} as possible signs of a changing political environment relating to education --- \textit{heterogeneity} could be framed positively or negatively, \textit{cumpulsory} could be applied to a politicization of subjects taught in public schools or to the compulsory nature of childhood education itself (e.g.\ by the home-schooling movement), and  \textit{bullying} could reflect a changing outlook of the responsibility of educators to discourage this kind of behavior among students.
\textit{Leisure time} (1990) and \textit{baby boy} (1990) are repeated from Table~\ref{tab:en-pmi-anno-most-pos}.

Table~\ref{tab:en-pmi-LR-most-neg} shows the triples with the highest predicted decrease in compositionality.
The compound predicted to have decreased in compositionality with respect to its modifier is \textit{rebel army}, centered around 1890.
In every decade other than the 1870s we can see some possible reference to the American Civil War (1861--1865).
\textit{Perryville, murfreesboro} (1880s), \textit{manassas} (1890s), \textit{monterey, petersburg}\footnote{Monterey could be in Mexico (with an extra `r'), or `Monterey Pass' in Pennsylvania. Petersburg could refer to a city in Virginia or to the Russian city.} (1910s) all likely refer to locations were battles were fought.
We also see cities in the former Confederacy (whose army in the war could be described as a \textit{rebel army}), like \textit{nashville} (1880s) and the former capital \textit{richmond} (1890s).
Many negative associated terms are seen up until the last decade: \textit{undisciplined, dishonorable} (1870s), \textit{disorganized} (1880s), \textit{treasonable} (1900s), while the 1910s has \textit{friendliness} as an associated term.
Particularly the presence of \textit{nicaragua} in the 1910s (the site of a US military occupation in that decade), in addition to \textit{monterey} (if indeed associated with the Mexican Revolution), and \textit{petersburg} (if associated with the February or October Revolutions in Russia) may show an expanding scope for the concept of \textit{rebel army}.
For the compound \textit{silk hat} (1850), we can observe that more descriptive adjectives occur in the last decades, like \textit{immaculate} (1900s) and \textit{dandified, dapper} (1910s).
This could reflect a change toward a less-compositional association between the article of clothing itself and its wearers.
Prior to this, there are more associated terms with other types of luxury clothing, e.g.\ \textit{jewellery, overcoat, neckcloth, frock-coat, necktie, coattails, spat}  as well as materials, like \textit{fur-trimmed, beaver, satin}.
Finally, \textit{lady friend} (1850) is repeated from Table~\ref{tab:en-pmi-anno-most-neg}.

\paragraph{Overall}

We see differences over time in the contexts of the target compounds that we examined in this evaluation, particularly those ranked as having changed the most per the annotation.
This helps to validate the task that the models were put to, that there are appreciable differences to be found in the temporal distribution of the contexts of at least some of the target compounds.

This evaluation perspective is less suitable to distinguish between the model configuration types that were used to provide top predictions of $\langle$compound, constituent, inflection point$\rangle$ triples that had changed the most. Still, we were able to point out where the model predictions overlapped with the annotated top predictions, and where model predictions seemed incongruous with the qualitative sense of semantic change in the association lists.

Examining the PMI association terms of only the target compounds (i.e.\ not also looking at the relevant constituents) prevents this evaluation perspective from helping us to directly interpret changes in compositionality, but presenting and examining additional association lists for the constituents would have made the analysis even more unwieldy.

\section{Conclusion}\label{ch8:discussion}

In this study, we created a dataset of fine-grained in-context compositionality ratings of German and English noun-noun compounds, and used them to investigate a proposed general rule of lexical semantic change. As a companion to this investigation, we defined the Compositionality Trend Prediction (CTP) task, and tested several types of semantic representations, each trained using different granularities or `schedules' of using the diachronic corpus data.
Our task presents a challenging scenario for computational semantic modeling, involving not only the complex inter-dependencies of noun compound semantics, but also the temporal specificity of its examples selected across a window of several decades.

Our answer to the first research question, \textbf{RQ1}, whether compounds maintain their degree of compositionality or become less compositional over time, is negative. 
We did not find a decisive trend describing developments in German or English noun compound compositionality.
That said, we identified challenges arising from our annotations, most of which originate in the temporal complexity of the task, where it cannot be assumed that either the compound or the constituent being compared has a stable set of meanings.

In the experimental component of this study, we successfully demonstrated a technique for predicting or characterizing changes in noun compound compositionality across incremental time-slices of diachronic corpora.
We investigated several types of semantic representations, compositionality metrics, and temporal window schedules to test the hypothesis of \textbf{RQ2}, asking whether contextual models using fine-grained temporal data would perform better than static representations using coarse-grained temporal data.
Generally, contextual models (\texttt{modern-bert} and \texttt{2nd-order}) were only ever marginally better than the static models (\texttt{cbow} and \texttt{skipgram}).  
The performance of the \texttt{modern-bert} models in this experimental scenario did not offset the sheer amount of time that it took to train the models (particularly the larger English models).
These representations should not be seen as a default option for modeling semantic change over time. Simpler representations can allow for greater flexibility in examining fine-grained temporal increments of datasets.

Measures that aggregate representations into a single difference measure (i.e.\ PRT and APD) performed better than clustering-oriented measures. This is not a surprising result, as we were comparing against compositionality ratings that are themselves taken in the aggregate.

Most importantly, we found that the higher performance of configurations using the narrower-grained \texttt{decades} and \texttt{sliding-full} schedules over the coarse-grained \texttt{full} schedule shows that we have created a task that encodes a temporal dependency, and that it was a successful demonstration of the technique for evaluating each target term in its own temporal window, i.e., on its own journey through the ages.

Finally, an analyses of targets rated or predicted as having changed the most are both a confirmation that the methods used in this study have some basic functionality for detecting or describing semantic change in these corpora (with some allowance for noise, variant spellings, and corpus effects), but they are also a cautionary tale: closely associated terms may not be easily attributable to a specific decade without more supporting context than a few sentences (see the example of \textit{rebel army} above), and that interpreting associated terms from too great a distance invites us to invoke modern meanings of those terms, which may not be justifiable from a close-reading perspective.
There were also instances where the changing salience of a meaning of a constituent may have contributed to annotation results that reflected more of a modern than a historical perspective.
An example that stands out is \textit{trust fund}, which had a mean compositionality rating with respect to \textit{trust} (across all decades) of 1.51 ($\pm 1.68$) --- \textit{trust} has senses referring to the legal instrument for managing a source of wealth, as well as to an older form of corporation called a \textit{trust}.
These (noun) senses are closely related to many uses of \textit{trust fund}, and many uses of \textit{trust fund} are more distantly related to the more common meaning of \textit{trust}, meaning confidence in someone or something.
Overall, the ratings for this compound, constituent pair were polarized.

\bibliographystyle{compling}
\bibliography{references,ssiwBib}

\appendix


\vspace{+5mm}
\appendixsection{Annotator Instructions}
\label{app:instructions}

\begin{center}
\begin{figure}
\includegraphics[width=0.9\linewidth]{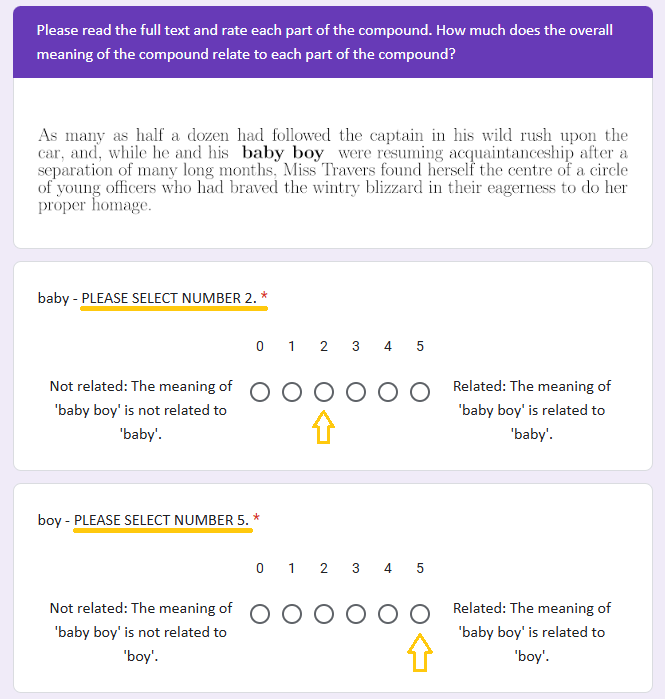}
\caption{`Attention check' question format for English annotation.}\label{fig:annotation-en-check-question}
\end{figure}
\end{center}

\begin{center}
\begin{figure}
\includegraphics[width=0.9\linewidth]{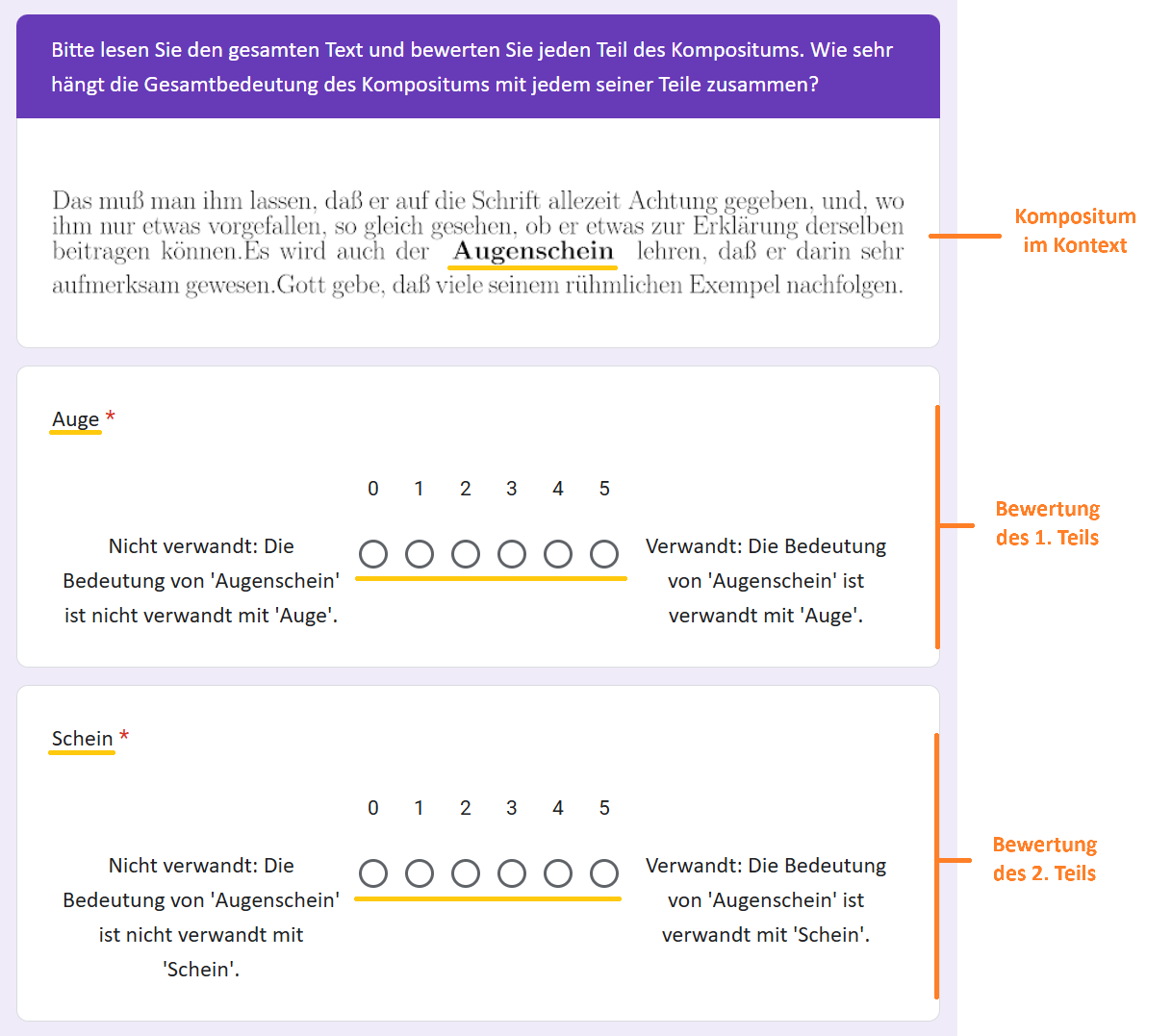}
\caption{Translated German instructions for compositionality-in-context annotators with a sample question.}\label{ch8:fig:example-question-de}
\end{figure}
\end{center}

\begin{center}
\begin{figure}
\includegraphics[width=0.9\linewidth]{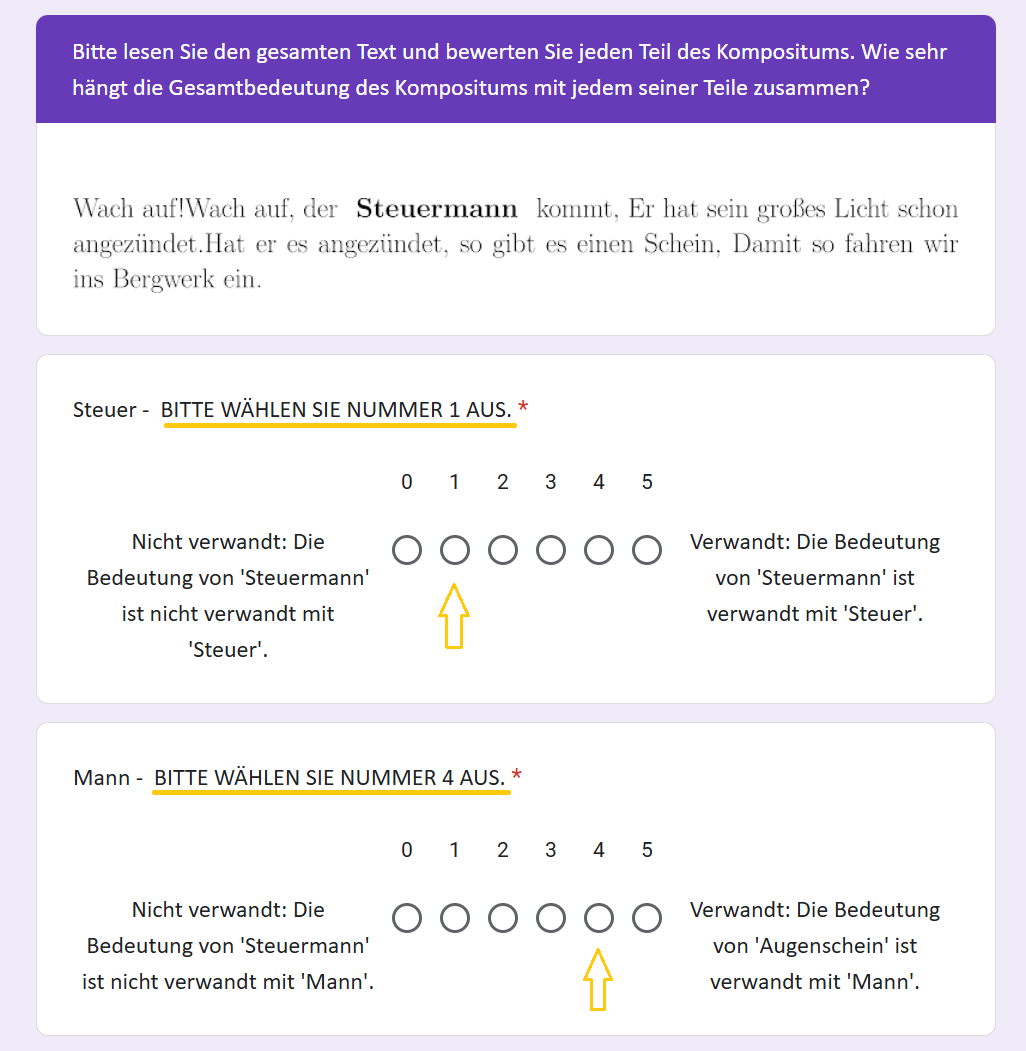}
\caption{`Attention check' question format for German annotation.}\label{ch8:fig:check-question-de}
\end{figure}
\end{center}

Here we show an additional part of the instructions presented to the annotators (as otherwise described in Section~\ref{ch8:annotation-procedure}, Figure~\ref{fig:annotation-en-standard-question}), which are instructions for the `attention check' questions in Figures~\ref{fig:annotation-en-check-question} (English) and~\ref{ch8:fig:check-question-de} (German), as well as the overall instructions for the German annotation in Figure~\ref{ch8:fig:example-question-de}.

\clearpage
\appendixsection{Modern-BERT Training}
\label{app:bert-training}
Each model was trained on a single NVIDIA RTX A6000 for 10 epochs.
The training data for any given time range was divided into train and evaluation sets, with a different `fold' of randomly determined training/evaluation data used in each epoch.
Models whose time slices contained more data, necessarily took longer to train in this setup.
In the (uncommon) case that the model loss and perplexity increased during an evaluation step between epochs, the model state was restored to the checkpoint at the end of the previous epoch, and training continued from there, since the next epoch would have a slightly different train/eval split.
Generally speaking, we achieved a training throughput of around 3.3 iterations (i.e.\ batches of examples) per second, which we can extrapolate from (per Equation~\ref{eq:training-time}) to calculate an estimate of the training time for a single run (discounting the time needed for evaluation after each epoch and the time taken to load the training data at program start).

\begin{equation}\label{eq:training-time}
  \text{train time} = (\frac{n}{\text{batch size}} \times \frac{1}{\text{iter/s}}) \times |\text{epochs}|
\end{equation}

At this scale of the number of models trained in total, it was not feasible or desired to adjust the hyperparameters for any individual model, but rather to pick reasonable defaults and to use them in every case.
To give a couple of concrete examples: the model trained on the COHA data from 1970--2009 took approximately 50 hours to train (on $\approx$102M tokens), while a smaller model like the one trained on DTA data from 1700--1709 took only approximately 2 hours to train (on $\approx$7M tokens).

Default training parameters were: learning rate: \num{5e-5}, max sequence length: 128, batch size: 128, adam epsilon: \num{1e-8}, mask percentage: 0.15. 


\clearpage
\appendixsection{Word Association Lists}
\label{app:word-lists}

We present the following pairs of PMI-ranked word lists:
\begin{itemize}
  \item $\langle$target, constituent, inflection-point$\rangle$ triples with the highest \textbf{annotated} (positive | negative) $\Delta$ compositionality.
 \vspace{+1mm}\\
 German: Table~\ref{tab:de-pmi-anno-most-pos} (+) and Table~\ref{tab:de-pmi-anno-most-neg} (-).
 \vspace{+1mm}\\
 English: Table~\ref{tab:en-pmi-anno-most-pos} (+) and Table~\ref{tab:en-pmi-anno-most-neg} (-).
 \item $\langle$target, constituent, inflection-point$\rangle$ triples predicted by the model with the highest $R^2$ score to have the highest (positive | negative) compositionality trend (from Tables~\ref{tab:ch8:de-avg-lr-repr-schedule} and~\ref{tab:ch8:en-avg-lr-repr-schedule} in Section~\ref{ch8:lr-scores}).
 \vspace{+1mm}\\
 German: Table~\ref{tab:de-pmi-LR-most-pos} (+) and Table~\ref{tab:de-pmi-LR-most-neg} (-)
\vspace{+1mm}\\
English: Table~\ref{tab:en-pmi-LR-most-pos} (+) and Table~\ref{tab:en-pmi-LR-most-neg} (-)

  \item $\langle$target, constituent, inflection-point$\rangle$ triples predicted by the model with the highest Spearman's $\rho$ correlation between its compositionality trend predictions, and the top/bottom 20 most extreme annotated $\Delta$ compositionality ratings (from Tables~\ref{tab:ch8:corr-extremes-de} and~\ref{tab:ch8:corr-extremes-en} in Section~\ref{ch8:sec:trend-prediction}).
 \vspace{+1mm}\\
 German: Table~\ref{tab:de-pmi-CORR-most-pos} (+) and Table~\ref{tab:de-pmi-CORR-most-neg} (-).
 \vspace{+1mm}\\
 English: The best model for this evaluation is the same as for the $R^2$ score evaluation above.
    \end{itemize}
Note: several $\langle$target, constituent, inflection-point$\rangle$ triples appear in more than one table.

\input{08_composition_tendency-wordlists}

\end{document}

%% file: 08_composition_tendency-wordlists.tex
\begin{landscape}
\begin{table}
  \footnotesize

  \resizebox*{!}{0.9\textwidth}{%
  \begin{tabular}{cll|ll|ll|ll|ll}
    Target & \multicolumn{2}{c}{$t_1$}  & \multicolumn{2}{c}{$t_2$} & \multicolumn{2}{c}{$t_3$} & \multicolumn{2}{c}{$t_4$} & \multicolumn{2}{c}{$t_5$} \\ \midrule

 & \multicolumn{2}{c}{1700}  & \multicolumn{2}{c}{1710}  & \multicolumn{2}{c}{1720}  & \multicolumn{2}{c}{1730}  & \multicolumn{2}{c}{1740} \\
Wechselbrief & indossiert & girieren & Welker & Payement & cambio & Homerum & inliegend & Hamburg & Frfl & Dukaten \\
1720 & geacceptirten & Andeme & Qvaestionis & bezahlend & Fiscus & Assignation & 5000 & à & 1000. & Barbar \\
mod/head & geindossirten & Protest & akzeptiert & tauff- & Creditor & einzeichnen & Geld-Summe & Thlr. & Obligation & listig \\
& Meßzeit & akzeptieren & acceptirter & giriert & Aviso-Brief & sothanen & Pilgrim & W. & Thl. & a \\
& Geacceptirten & Bellin & Wechsel-Schuld & direkte & Acceptant & Ben & 10000 & mehrenteils & Spec. & hierbei \\
& Aushändigung & gekehrt & Skontrierung & Literae & versendet & fertigen & übersenden & gewaltig & 150. & Leipzig \\
& Secunda & geacceptirte & Vista & Documentorum & Uso & Secunda & Brauch & Taler & 2000 & folgen \\
& geprotestirten & einhaltend & Sola & Acceptationes & 50000. & akzeptieren & hungrig & G. & 300. & dergleichen \\
& geindossirter & Partimeno & einzukassierend & Abschreibung & Virgilium & Verzögerung & Thl. & womit & Paß & unsere \\
& furnieren & indossieren & Akzeptieren & girieren & Verkehr & respektieren & 100. & antreffen & Fl. & lassen \\
\midrule

 & \multicolumn{2}{c}{1730}  & \multicolumn{2}{c}{1740}  & \multicolumn{2}{c}{1750}  & \multicolumn{2}{c}{1760}  & \multicolumn{2}{c}{1770} \\
Sinnbild & lehre-reich & malen & Ehrenreich & strahlend & Inschrift & Neubegierde & Wohltun & überhäufen & Verunreinigung & Inschrift \\
1750 & belehrend & schrecken & Viole & aufgestellt & Warze & jugendlich & dea & Sturz & Erdenrund & allegorisch \\
head & ausmustern & Große & Kaudelka & ritter & Fußstapfen & erbaulich & ausstaffieren & Kommerz & horazischen & Überschrift \\
& versilbern & Reinigung & Wappen-Schild & Nichtigkeit & Schriftstelle & verlaufen & Thalia & Gratien & aufspringen & Satyr \\
& lehrend & Geiz & belehrend & nichtig & Platte & Rock & Schutzgottin & einladen & Allegorie & Deutung \\
& Gnaden-Stand & Spruch & Wankelmut & lehrend & Deckel & Spruch & Münzengeschmack & Abbildung & Budsdo & entlehnen \\
& lehr- & Sievers & Friedenstein & Dreinest & Nummer & Betrübnis & roma & Aufschrift & Sinnbild & Wappen \\
& land- & vorstellen & geschlingen & irgendwo & abzeichnen & überaus & Allwissenheit & abstrakt & Portrait & umkehren \\
& sogar & preisen & fest- & gemalt & anmutig & stillschweigend & unförmlich & Tapferkeit & 620 & Besondere \\
& Erleuchtung & Blume & Vergänglichkeit & Devise & Merkmal & Verzeichnis & gebildet & Sprichwort & Errettung & Meisterstück \\

\midrule

 & \multicolumn{2}{c}{1870}  & \multicolumn{2}{c}{1880}  & \multicolumn{2}{c}{1890}  & \multicolumn{2}{c}{1900}  & \multicolumn{2}{c}{1910} \\
Edelmann & kurländischer & florentinisch & vond & kunstliebend & Octavius & Reisebegleiter & herabgekommen & Geiz & tiefverschuldet & 1888-92 \\
1890 & simpeln & Lastträger & Polenkönig & aufrechtstehend & bonum & Luigi & geziemen & Sohle & hinleiten & respektabel \\
head & neugebacken & *378 & Pelzmantel & abgetragen & Königliche & Strada & bretonisch & Gutsbesitzer & Leibeigner & aufgreifen \\
& Scharwerk & Westphal & Kiszka & Bettelmann & Zollen & Ehrenmann & Reformationszeit & Latude & Cancan & imponierend \\
& Oelsen & gleichmütig & Gastold & hochgebären & Sales & begütern & huschen & flüchtig & ekeln & Scheideweg \\
& Bassompierre & feingebildet & Witold & graf & Andrian & beansprucht & Röte & allmählich & Speis & befleißen \\
& Kohlenwerk & Rodenstein & genuesisch & doe & Bürokrat & Gerichtsherr & Kavalier & Erziehung & Mordlust & 1886-87 \\
& Fähndrich & unwirsch & abgünstig & Soter & Demidoff & Geldpreis & Scheitel & bürgerlich & lumpig & Totschlag \\
& sechzigst & beikommen & toen & Monseigneur & Mylius & Edelmann & Eitelkeit & echt & Gewalttätigkeit & verschmelzen \\
& Kindesbein & Zeder & ruhmredig & Hauptbeschäftigung & Wilddieb & Kroatien & ritterlich & lernen & Dönniges & Beschämung \\
\midrule

 & \multicolumn{2}{c}{1810}  & \multicolumn{2}{c}{1820}  & \multicolumn{2}{c}{1830}  & \multicolumn{2}{c}{1840}  & \multicolumn{2}{c}{1850} \\
Steuermann & mühselig & hinaus & ebendaher & absonderlich & Tiphys & Hieron & todverachtend & Holzm & Tutuma & Steuerruder \\
1830 & Fahrt & Arbeit & Schiffs-Kapitaine & Abrede & Ancäus & verhängt & Meergott & Frankenturm & Palinurus & Nathanael \\
head & gelassen & fahren & Richteramt & Übelstand & scharfblickend & entsinken & Palinurus & durchschiffen & Damis & Fletcher \\
& Klippe & kennen & Deep & wohlbehalten & Soli & wachsam & Raumann & Mammon & Schiffswand & Parteiung \\
& abstoßen & Bewegung & befehligen & schossen & Ruderer & Luftchen & Fritzem & Cäcilienstr & -ahsch & Ruderer \\
& ans & gerade & Staatsruder & Seeleute & Steuerruder & unverwandt & Delmenhorst & Kostgasse & gelenkt & Argo \\
& Schiff & heißen & Schiffsmannschaft & Abfahrt & zusammengepreßt & erkranken & Veronica & Back & Erblicken & indianisch \\
& verlangen & zurück & ober- & verdingen & eisgrau & aussehend & kopfüber & Tapezierer & Lysander & siebzehn \\
& Richtung & Land & hinterdrein & Peters & Lynceus & Ostern & Köter & Ration & Gubernium & Lotse \\
& glücklich & bleiben & gestreng & Dolmetscher & erlahmen & Mägdlein & Bartholomäus & Salzgasse & durchschiffen & Hermon \\
    \bottomrule
  \end{tabular}
  }%
  \caption{Per-decade top 20 PMI-associated words (German) for $\langle$compound, constituent, inflection point$\rangle$ with highest positive Delta compositionality per annotation (from Table~\ref{tab:de-anno-trends}).}\label{tab:de-pmi-anno-most-pos}
\end{table}
\end{landscape}

\begin{landscape}
\begin{table}
  \footnotesize

  \resizebox*{!}{0.8\textwidth}{%
  \begin{tabular}{cll|ll|ll|ll|ll}
    Target & \multicolumn{2}{c}{$t_1$}  & \multicolumn{2}{c}{$t_2$} & \multicolumn{2}{c}{$t_3$} & \multicolumn{2}{c}{$t_4$} & \multicolumn{2}{c}{$t_5$} \\ \midrule

 & \multicolumn{2}{c}{1720}  & \multicolumn{2}{c}{1730}  & \multicolumn{2}{c}{1740}  & \multicolumn{2}{c}{1750}  & \multicolumn{2}{c}{1760} \\
Donnerwetter & Nord-Winde & einschlagen & observiert & Platzregen & schloßen & überfällen & Wetterleuchten & brausen & nachgerade & Teufel \\
1740 & Blitzen & ausdehnend & Donnerschlag & zerstreut & Nebukadnezar & gebieten & Nebukadnezar & vorüber & herankommen & Flur \\
mod/head & wahren & Hagel & Pfingst & grausam & Platz-Regen & benachbart & Verurteilung & benachbart & fluchten & Sturm \\
& variabel & vermehrt & gottlob & Wasserflut & Stieg & 21. & Allerheilig & Regen & Ist & Arm \\
& 318. & schwanger & Regenguß & Schloße & Morgenland & fürchterlich & Stieg & warten & Raubsucht & schrecklich \\
& 325 & Wirkung & Nüssers & Hirschfeld & umwenden & grausam & Schaffen & dunkel & Nordhausen & sieben \\
& präsupponieren & hierunter & erschrecklich & Feuers-Brunst & spalten & Mai & zusprechen & Wiesbaden & Digression & Fürst \\
& Blitz & Regen & Donner-Schlag & Sturm-Winde & brausen & erregen & spalten & eröffnen & Lenz & Meer \\
& heiter & aufzeichnen & Mittel-Herwigsdorff & 1673. & Jul. & ziemlich & einschlagen & Schloß & täuschen & bestätigen \\
& vergesellschaften & dreierlei & nachmittags & Lausitz & vorüber & warten & 308 & Abend & Fluch & wild \\

\midrule

 & \multicolumn{2}{c}{1880}  & \multicolumn{2}{c}{1890}  & \multicolumn{2}{c}{1900}  & \multicolumn{2}{c}{1910} \\
Gasthaus & Yadoya & Kramladen & Schöll & Tribuswinkel & Kerschbaum & Kegelbahn & Holansky & Tscherne \\
1900 & schmidt'schen & Bauerngut & Grausam & Kreuzgasse & Manner-Gesang-Verein & Frühlingsliedertafel & mader & Pukl \\
mod & zusammenrotten & Badegeld & Stubbenkammer & Kottingbrunn & Hausball & Drobnitsch & kippen & Meichenitsch \\
& Theehäuser & Vincenz & Tanzunterhaltung & Hussowitz & Liederkranz & Blaschitz & asinnig & Jahresversammlung \\
& Pus & Neustift & Breitenseerstraße & Hausball & Wehrhonig & Linde & Wodenig & Gittertor \\
& Lusthaus & Müggelbude & Wassergasse & Theehäuser & Verbandsabend & Kottingbrunn & Vorlegeplatte & Fiden \\
& Bockenheim & Krawall & Eder & Teesdorf & Tombola & Sooß & Brudermann & Custer \\
& angerufen & makt & rim & Feuerwehrkapelle & Steueramtsoffizial & Police & Bahnbau & Roßwein \\
& Osterode & freiw. & Werkeltag & Bauernball & Laschitsch & Neuhold & 1/411 & Unwesen \\
& Michaeli & dekorieren & Vinicssay & überschattet & Kumeric & Lettemüller & zechen & Taverne \\

\midrule

 & \multicolumn{2}{c}{1880}  & \multicolumn{2}{c}{1890}  & \multicolumn{2}{c}{1900}  & \multicolumn{2}{c}{1910}  & \multicolumn{2}{c}{1920} \\
Mittelstand & Schleswigholsteins & Groll & Zerbröckelung & agrarisch & gutgesinnt & gesinken & Wohnungsfürsorge & ahmen &   &   \\
1900 & hineintragen & Rate & handeltreibend & Arbeiterstand & Ladengeschäft & Verspätung & Steuerpolitik & Favorit &   &   \\
mod & kantisch & fleißig & emporsteigend & Arbeitsgebiet & Familienzucht & hervorgegehen & Versorgungsanstalt & gewerbe- &   &   \\
& selbstbewußt & Reichtum & abnützen & Minderzahl & Erzeugungsart & Unfreie & Speisung & Flitter &   &   \\
& arglos & wachsend & Lebenskampf & Kochkunst & Ehrbarkeit & Aristokratie & arbeiter- & ausrücken &   &   \\
& Gleichartigkeit & jugendlich & bauern- & zerstörend & verkümmert & Wirtschaftlichkeit & Hauptziel & Schichte &   &   \\
& wohlhabend & sozial & Cook & Vergewaltigung & pedantisch & sukzessiv & Claudia & Wähler &   &   \\
& vorherrschen & Armut & Mittelstand & rapid & angelsächsisch & nachstehen & Kinderhort & Abnahme &   &   \\
& altdeutsch & Überzeugung & hinübergehen & chronisch & Hauptgruppe & rekrutieren & Junktim & ledig &   &   \\
& physisch & Bevölkerung & Berufsstand & Augustus & Gemütsleben & Kleinbauer & Eheschließungsziffer & kräftigen &   &   \\

\midrule

 & \multicolumn{2}{c}{1740}  & \multicolumn{2}{c}{1750}  & \multicolumn{2}{c}{1760}  & \multicolumn{2}{c}{1770} \\
Wechselbrief & Frfl & Dukaten & 4636 & 656 & ausgestellt & Formul & Papiergeld & kaufen \\
1760 & 1000. & Barbar & Wechselbillet & Kurrentgeld & Tauschbrief & fünfhundert & Wechselhandel & zerreissen \\
mod & Obligation & listig & trassiert & Verhandlung & Habe. & Rudolph & Karolin & verkaufen \\
& Thl. & a & akzeptiert & verjähren & Ausgeber & Foderung & Taschenuhr & Summe \\
& Spec. & hierbei & ausgestellt & Trassanten & verjährt & uneigentlich & sechshundert & A \\
& 150. & Leipzig & ausstellen & wechsel-gläubig & Beschihet & wechselrecht & mitgeben & bezahlen \\
& 2000 & folgen & Ausgeber & tertia- & Sola & Unterpfand & Paß & Erfindung \\
& 300. & dergleichen & verjährt & saldieren & Reichstaler & vergüten & Verstorbene & entgegen \\
& Paß & unsere & Aval & ficta & niederschreiben & be- & Verkaufe & zweien \\
& Fl. & lassen & Negotiant & exakt & October & ausstellen & Lord & tausend \\

    \bottomrule
  \end{tabular}
  }%
  \caption{Per-decade top 20 PMI-associated words (German) for $\langle$compound, constituent, inflection point$\rangle$ with highest negative Delta compositionality per annotation (from Table~\ref{tab:de-anno-trends}).}\label{tab:de-pmi-anno-most-neg}
\end{table}
\end{landscape}

\begin{landscape}
\begin{table}
  \footnotesize

  \resizebox*{!}{0.8\textwidth}{%
  \begin{tabular}{cll|ll|ll|ll|ll}
    Target & \multicolumn{2}{c}{$t_1$}  & \multicolumn{2}{c}{$t_2$} & \multicolumn{2}{c}{$t_3$} & \multicolumn{2}{c}{$t_4$} & \multicolumn{2}{c}{$t_5$} \\ \midrule

 & \multicolumn{2}{c}{1880}  & \multicolumn{2}{c}{1890}  & \multicolumn{2}{c}{1900} \\
Marktpreis & Depreciation & Mk. & regulierend & Produktionspreis & Guinee & 1900 \\
1900 & Konjunktur & desjenig & Blattel & oszillatorisch & 1779 & Juni \\
head & herabgehen & Viertel & Selbigkeit & Steigen & abkaufen & Gewicht \\
& Zollsatz & Weizen & Marktwert & Oszillation & juristisch & Juli \\
& Sachwert & Quantität & oszillieren & uniform & 1848 & entwickeln \\
& wechsel- & Erzeugnis & Repulsion & Ergiebigkeit & anerkennt & sogar \\
& Preiswechsel & beispielsweise & Konkurrenzverhältnis & Bodenprodukt & angeben & Begriff \\
& Transportkosten & abnehmen & Extraprofit & Zuschußkapital & 1901 & Wert \\
& Landweg & entschließen & Verkaufsgeschäft & Durchschnittszahl & Verfügung & schließlich \\
& Durchschnittspreis & Berechnung & Bebauer & Einkaufspreis & 7. & 20 \\

\midrule

 & \multicolumn{2}{c}{1880}  & \multicolumn{2}{c}{1890}  & \multicolumn{2}{c}{1900}  & \multicolumn{2}{c}{1910}  & \multicolumn{2}{c}{1920} \\
Kaffeehaus & gast- & Besitzer & Loupians & Restaurant & Rockdieb & Lesehalle & angebracht & verschwenderisch & Terrasse & Nacht \\
1900 & verheern & jetzig & Trauerkleid & neuartig & Grog & S' & Gallipoli & Hehl & Rang & treten \\
mod & besucht & sonstig & Montmartre & Warenmagazin & Versicherungsagent & Palme & Tischgesellschaft & Ab. & verdächtig & ein \\
& plündern & verbieten & Lucher & Bureaux & Punsch & Herrin & Feldbach & hieraus & drüben & sitzen \\
& Lokalität & Frankfurt & Teatro & Kaufladen & gast- & Na & Billard & Pyrmont & Frühstück & bringen \\
& kraus & Fenster & change & Prinzipal & Verpachtung & Sagl & Pancsova & mustern & nähern & dort \\
& Hotel & Hund & Loupian & weilend & Parade & zertrümmert & verlohnen & Jurist & gestern & gerade \\
& Zigarre & Anspruch & Picaut & Terrasse & Letztere & Rechtsanwalt & Regierungsbeamte & Hotel & nieder & aus \\
& zerschlagen & M. & gast- & Gasthof & Wanyek & Abwechselung & vergeuden & Mittwoch & beugen & klein \\
& *18 & zahlreich & Bogengang & Comptoir & Steinwurf & Fontana & Pathos & vernichten & irgendeine & tun \\

\midrule

 & \multicolumn{2}{c}{1700}  & \multicolumn{2}{c}{1710}  & \multicolumn{2}{c}{1720}  & \multicolumn{2}{c}{1730}  & \multicolumn{2}{c}{1740} \\
Wechselbrief & indossiert & girieren & Welker & Payement & cambio & Homerum & inliegend & Hamburg & Frfl & Dukaten \\
1720 & geacceptirten & Andeme & Qvaestionis & bezahlend & Fiscus & Assignation & 5000 & à & 1000. & Barbar \\
mod & geindossirten & Protest & akzeptiert & tauff- & Creditor & einzeichnen & Geld-Summe & Thlr. & Obligation & listig \\
& Meßzeit & akzeptieren & acceptirter & giriert & Aviso-Brief & sothanen & Pilgrim & W. & Thl. & a \\
& Geacceptirten & Bellin & Wechsel-Schuld & direkte & Acceptant & Ben & 10000 & mehrenteils & Spec. & hierbei \\
& Aushändigung & gekehrt & Skontrierung & Literae & versendet & fertigen & übersenden & gewaltig & 150. & Leipzig \\
& Secunda & geacceptirte & Vista & Documentorum & Uso & Secunda & Brauch & Taler & 2000 & folgen \\
& geprotestirten & einhaltend & Sola & Acceptationes & 50000. & akzeptieren & hungrig & G. & 300. & dergleichen \\
& geindossirter & Partimeno & einzukassierend & Abschreibung & Virgilium & Verzögerung & Thl. & womit & Paß & unsere \\
& furnieren & indossieren & Akzeptieren & girieren & Verkehr & respektieren & 100. & antreffen & Fl. & lassen \\

\midrule

 & \multicolumn{2}{c}{1880}  & \multicolumn{2}{c}{1890}  & \multicolumn{2}{c}{1900}  & \multicolumn{2}{c}{1910}  & \multicolumn{2}{c}{1920} \\
Krankenhaus & armen- & befreunden & Ratsche & Kirchlichkeit & Ratsch & Kapokmatratze & Ratsche & Abteilungsvorstand & endelos & sterben \\
1900 & Neujahrsgeschenk & auswärts & Heimatsland & Wallenstadt & 10--11 & verführt & überstellen & 1829-31 & Spaziergang & alt \\
head & Bettstelle & polizeilich & Gewalttäter & Einsame & Ratsche & Verkehrsanstalten & Sekundararzt & schwerverletzt & besuchen & ins \\
& 25.000 & Brünn & Öffentl. & Moabit & Assistenzarzt & Siechenhaus & Rettungswagen & Schädelbruch & Wagen & Fridolin \\
& 30,000 & zufallen & Versorgungshaus & Misericordia & Zeppelin & Prachtbau & Irrenanstalten & Hilfesarzt & nahe & Zeit \\
& Legat & überführen & Primararzt & Allgem & Rapport & Maria-Enzersdorf & Verwaltungsdirektor & Blutvergiftung & Straße & lassen \\
& vermachen & Anstoß & Deutsch-Afrika & Casa & Allgem & Maller & Kottnik & allernächst & allgemein & ihr \\
& behandelnd & bewährt & Jakobikirche & Verpflegung & Stadtfriedhof & Arznei & Hagener & verbauen & schaffen & zum \\
& Vorsteher & bestellen & Fliedner & Staatsbeitrag & Schweinestall & Aloisia & Ballonhülle & suczawaer & kaum & dann \\
& Lebzeiten & amtlich & Diakonisse & Schlachthaus & Rahner & Schulküche & Assistenzzeit & gast- & fahren & nach \\

    \bottomrule
  \end{tabular}
  }%
  \caption{Per-decade top 20 PMI-associated words (German) for $\langle$compound, constituent, inflection point$\rangle$ predicted as having the highest positive compositionality trend by top configuration in terms of $R^2$ score: $\langle$\texttt{2nd-order}, \texttt{decades}, \texttt{jsd}$\rangle$ (ranking in Table~\ref{tab:ch8:de-avg-lr-repr-schedule}).}\label{tab:de-pmi-LR-most-pos}
\end{table}
\end{landscape}

\begin{landscape}
\begin{table}
  \footnotesize

  \resizebox*{!}{0.8\textwidth}{%
  \begin{tabular}{cll|ll|ll|ll|ll}
    Target & \multicolumn{2}{c}{$t_1$}  & \multicolumn{2}{c}{$t_2$} & \multicolumn{2}{c}{$t_3$} & \multicolumn{2}{c}{$t_4$} & \multicolumn{2}{c}{$t_5$} \\ \midrule

 & \multicolumn{2}{c}{1710}  & \multicolumn{2}{c}{1720}  & \multicolumn{2}{c}{1730}  & \multicolumn{2}{c}{1740}  & \multicolumn{2}{c}{1750} \\
Wortspiel &   &   & Burlesque & 43 & Plinii & altfränkisch & dasmal & ausdrucken & verblümt & Ausschweifung \\
1730 &   &   & Junktur & Kompliment & Pickelhering & Schelle & läppisch & zweideutig & Folgende & Scherz \\
head &   &   & Formulen & anbringen & ausdienen & unzüchtig & schwülstig & anbringen & leichtfertig & verflucht \\
&   &   & pedantisch & 40 & Spitzfindigkeit & vorzeiten & Metaforum & kahl & schmutzig & Gedicht \\
&   &   & Sinnbild & häufig & Senecae & Wiederholung & Galimathias & Vortrefflichkeit & begabt & leer \\
&   &   & 317 & Sprache & Blümen & enfin & frostig & Gespött & ersinnen & Vetter \\
&   &   & 334 & etc. & Ex & Dichtkunst & triftig & gezwingen & Zweideutigkeit & edel \\
&   &   & 210 & verschieden & Zweideutigkeit & halber & verdrehen & possierlich & albern & 7. \\
&   &   & sqq. & darin & Weltweißheit & ausdrücklich & Phöbus & geistreich & Hauptwerk & unglücklich \\
&   &   & ungeheuer & allerhand & verwerflich & raison & Gallus & Übersetzer & Robert & 6. \\

\midrule

 & \multicolumn{2}{c}{1870}  & \multicolumn{2}{c}{1880}  & \multicolumn{2}{c}{1890}  & \multicolumn{2}{c}{1900} \\
Wortspiel & Huste & Lib & synonymisch & Säuger & plätschernd & Pfeifer & Gedankenwitz & erwecken \\
1890 & Radlein & Laborant & Trutzlied & expansiv & abpassen & Geheimrat & gedanken- & schätzen \\
mod & Ortsname & Kreuzberg & Meto & böhmerwald & Hebräische & Humor & witzig & ausgehen \\
& Feister & Greifenberger & Kurat & ausschelten & Wunderbare & Witz & Gefallen & Goethe \\
& Lindenholz & 5755 & klauben & Seiger & Klangassoziation & beliebt & Fechner & hervor \\
& massweise & 4879 & Sägespan & Liebesgedicht & Antithese & lächerlich & 218 & Sprache \\
& knickern & 4599 & Colt & Kniphausen & fad & 49 & 219 & Umstand \\
& antithetisch & 4014 & nolo & Fatto & Ironie & Vorliebe & Witz & davon \\
& Volksname & Salbei & hysterisch & Nebeneinanderstellung & boshaft & Beifall & Vergleich & legen \\
& Räpes & Verfallen & Zugvieh & Mank & Gen. & 29 & Keim & Entwicklung \\

\midrule

 & \multicolumn{2}{c}{1810}  & \multicolumn{2}{c}{1820}  & \multicolumn{2}{c}{1830}  & \multicolumn{2}{c}{1840}  & \multicolumn{2}{c}{1850} \\
Steuermann & mühselig & hinaus & ebendaher & absonderlich & Tiphys & Hieron & todverachtend & Holzm & Tutuma & Steuerruder \\
1830 & Fahrt & Arbeit & Schiffs-Kapitaine & Abrede & Ancäus & verhängt & Meergott & Frankenturm & Palinurus & Nathanael \\
head & gelassen & fahren & Richteramt & Übelstand & scharfblickend & entsinken & Palinurus & durchschiffen & Damis & Fletcher \\
& Klippe & kennen & Deep & wohlbehalten & Soli & wachsam & Raumann & Mammon & Schiffswand & Parteiung \\
& abstoßen & Bewegung & befehligen & schossen & Ruderer & Luftchen & Fritzem & Cäcilienstr & -ahsch & Ruderer \\
& ans & gerade & Staatsruder & Seeleute & Steuerruder & unverwandt & Delmenhorst & Kostgasse & gelenkt & Argo \\
& Schiff & heißen & Schiffsmannschaft & Abfahrt & zusammengepreßt & erkranken & Veronica & Back & Erblicken & indianisch \\
& verlangen & zurück & ober- & verdingen & eisgrau & aussehend & kopfüber & Tapezierer & Lysander & siebzehn \\
& Richtung & Land & hinterdrein & Peters & Lynceus & Ostern & Köter & Ration & Gubernium & Lotse \\
& glücklich & bleiben & gestreng & Dolmetscher & erlahmen & Mägdlein & Bartholomäus & Salzgasse & durchschiffen & Hermon \\

\midrule

 & \multicolumn{2}{c}{1870}  & \multicolumn{2}{c}{1880}  & \multicolumn{2}{c}{1890}  & \multicolumn{2}{c}{1900}  & \multicolumn{2}{c}{1910} \\
Bauernstand & festhaltend & xv. & bürger- & Indifferentismus & Treiben\_Es & Bauernkrieg & Agrarpolitik & Agrarierpolitik & Grundforderung & 121 \\
1890 & Köllmern & neugeschaffen & Pfarrgeistlichkeit & Kudlich & Berufsorganisation & lastend & scheel & Handwerkerstand & Familienangehörige & Amme \\
mod & Kleinwirtschaft & dekretieren & Bodenwert & Apostrophe & Schweiz. & Hörige & bürger- & niederträchtig & Aufsteigen & Alpenland \\
& lohn- & Sanct-Galle & anhänglich & kernhaft & bürger- & Wohlhabenheit & NordEuropa & Schimpf & Bauernstand & berücksichtigen \\
& Offiziersstelle & Zurückgehen & Hiecke & Vertretung & wehrhaft & Vollmar & Fernhaltung & Pluralität & Bauernhaus & Existenz \\
& gewohnheitsmässig & selbstgewählt & auskaufen & Kurmark & gewerbe- & Fortbestand & Ablösungsgesetz & verschlechtern & kleinrussisch & Erhaltung \\
& herren- & Lebensbeschreibung & Emporkommen & Fronde & eingebringt & Bauernhof & tiefstehend & entsprießen & Verarmung & mittel \\
& Gewerbestand & herabsehen & Kolonat & Irland & Erlaßung & mürrisch & handwerker- & anschwellen & B & Linie \\
& Karsthanse & Täufling & kurzgefaßt & Bauernstand & Notlage & Grundbedingung & durchbringen & Kräftigung & bürger- & gleichfalls \\
& bürger- & betrüglich & Kreisversammlung & Erweckung & Frondienst & Gerichtssaal & Bürgerstand & Notlage & beteiligt & wirtschaftlich \\

    \bottomrule
  \end{tabular}
  }%
  \caption{Per-decade top 20 PMI-associated words (German) for $\langle$compound, constituent, inflection point$\rangle$ predicted as having the highest negative compositionality trend by top configuration in terms of $R^2$ score: $\langle$\texttt{2nd-order}, \texttt{decades}, \texttt{jsd}$\rangle$ (ranking in Table~\ref{tab:ch8:de-avg-lr-repr-schedule})}\label{tab:de-pmi-LR-most-neg}
\end{table}
\end{landscape}

\begin{landscape}
\begin{table}
  \footnotesize

  \resizebox*{!}{0.8\textwidth}{%
  \begin{tabular}{cll|ll|ll|ll|ll}
    Target & \multicolumn{2}{c}{$t_1$}  & \multicolumn{2}{c}{$t_2$} & \multicolumn{2}{c}{$t_3$} & \multicolumn{2}{c}{$t_4$} & \multicolumn{2}{c}{$t_5$} \\ \midrule

 & \multicolumn{2}{c}{1730}  & \multicolumn{2}{c}{1740}  & \multicolumn{2}{c}{1750}  & \multicolumn{2}{c}{1760}  & \multicolumn{2}{c}{1770} \\
Briefwechsel & Dichterin & bereits & Rechtmäßigkeit & Hauptsache & Gerichtsplatz & Vernehmen & angelegentlich & englisch & ertappt & hingeben \\
1750 & führend & angenehm & Stadtssache & fortsetzen & höchstschädlich & Absetzung & abreissen & Lampert & Liebesverständnis & Auszug \\
mod & vertraut & er & Inspirierte & untersagen & einlassen & verschicken & Straßburg & geheim & Maximo & Stauzius \\
& ofenentelich & machen & Briefträger & Trog & vieljährig & angefangen & Unterhaltung & aufhören & Ossian & Aller \\
& einlassen & durch & abbrechen & unerlaubt & überhäuft & unterbleiben & britisch & Herrschaft & Kriegsgefangene & vervielfältigen \\
& sinnreich & die & hineinziehen & geheim & verräterisch & angewendet & unterhalten & Magister & nichtig & Begegnung \\
& unterhalten & eine & verbieten & Würtemberg & vorgeschlagen & Wannenher & Dion & gnädig & Einkleidung & Psychologie \\
& belieben & zu & unterhalten & Veranlaßung & unterhalten & Einkauf & einlassen & Freund & Zauberei & vormalig \\
& einrichten & als & widerraten & aufgeben & 192 & Zeitvertreib & London & wodurch & unerlaubt & 156 \\
& beschaffen & sich & verstohlen & bedrängt & Hauptabsicht & Vertraulichkeit & würdigen & Materie & schwärmerisch & Rambold \\

\midrule

 & \multicolumn{2}{c}{1880}  & \multicolumn{2}{c}{1890}  & \multicolumn{2}{c}{1900}  & \multicolumn{2}{c}{1910}  & \multicolumn{2}{c}{1920} \\
Kaffeehaus & gast- & Besitzer & Loupians & Restaurant & Rockdieb & Lesehalle & angebracht & verschwenderisch & Terrasse & Nacht \\
1900 & verheern & jetzig & Trauerkleid & neuartig & Grog & S' & Gallipoli & Hehl & Rang & treten \\
mod & besucht & sonstig & Montmartre & Warenmagazin & Versicherungsagent & Palme & Tischgesellschaft & Ab. & verdächtig & ein \\
& plündern & verbieten & Lucher & Bureaux & Punsch & Herrin & Feldbach & hieraus & drüben & sitzen \\
& Lokalität & Frankfurt & Teatro & Kaufladen & gast- & Na & Billard & Pyrmont & Frühstück & bringen \\
& kraus & Fenster & change & Prinzipal & Verpachtung & Sagl & Pancsova & mustern & nähern & dort \\
& Hotel & Hund & Loupian & weilend & Parade & zertrümmert & verlohnen & Jurist & gestern & gerade \\
& Zigarre & Anspruch & Picaut & Terrasse & Letztere & Rechtsanwalt & Regierungsbeamte & Hotel & nieder & aus \\
& zerschlagen & M. & gast- & Gasthof & Wanyek & Abwechselung & vergeuden & Mittwoch & beugen & klein \\
& *18 & zahlreich & Bogengang & Comptoir & Steinwurf & Fontana & Pathos & vernichten & irgendeine & tun \\

\midrule

 & \multicolumn{2}{c}{1830}  & \multicolumn{2}{c}{1840}  & \multicolumn{2}{c}{1850}  & \multicolumn{2}{c}{1860}  & \multicolumn{2}{c}{1870} \\
Ehefrau & Settegast & Einbildungskraft & Wipf & Manu & mariti & Confarreatio & durchstampfen & bas & Vorzugsrechte & uth \\
1850 & Setgast & Pfahl & Spranthal & Waldinger & Verwandtengericht & Ehemann & enterben & Ursula & pfand- & Vorzugsrecht \\
head & komfortabel & 1831 & Nuber'schen & Rosenfelder & Schaffhauser & Lea & Böglein & ausschweifend & Steinkrug & wohnhaft \\
& Kreuzfahrer & Juli & Nuber'sche & Grundvermögen & Krönerin & necis & sug & Schneckenhaus & Familiengenosse & Telegraphenbeamte \\
& jegliche & Wilhelm & Mönchweiler & wohnend & tugendsam & Fiscus & *341 & huren & Vormünderin & gescheiden \\
& geb. & Friedrich & Amtschirurg & Dingwall & angetraut & Militairpersonen & verschließend & Mariner & Schent & minderjährig \\
& Dorothea & Maria & Eheband & Witwenpension & Sacktuch & rügen & Genossein & Manus & unversorgt & Vergiftung \\
& 26sten & echt & Bierbräuer & falliert & unangesehen & verargen & Gertrud & Hausarbeit & Patronin & böswillig \\
& Anna & angenehm & hoffmansch & Wiederverheiratung & Manu & zurückgelaßen & Stirnband & Frühe & Pati & Entmündigung \\
& Katharina & Gegensatz & Controlirung & Vorgenannte & Bemächtigung & vielgestaltig & Ehefrau & Ehebruch & Prozeßfähigkeit & Brautschatz \\

\midrule

 & \multicolumn{2}{c}{1740}  & \multicolumn{2}{c}{1750}  & \multicolumn{2}{c}{1760}  & \multicolumn{2}{c}{1770}  & \multicolumn{2}{c}{1780} \\
Handelsmann & kauf- & Großmutter & um- & 506 & Görlitz & Solchemnach & Ratsverwandte & üppig & Kyrie & unterwerfen \\
1760 & kauff- & wohlbekannt & Amts-Verrichtung & Fuhrleute & Wolff & verzeihen & Rall & einkaufen & Aichach & wahrlich \\
head & Pegau & Schmidts & gallisch & 484 & Hanse & rechtmässig & Kall & Kredit & Kämmerer & wiederum \\
& Möllers & Georgius & neuangehend & rühmlich & kauf- & Leipzig & kauf- & verführen & Schneeberg & übel \\
& Rats-Verwandte & wohnhaft & kauf- & Abbruch & 386. & ehemalig & Hofmedikus & unterdrücken & Süden & gebären \\
& Hampel & nachgelaßen & Savary & 245 & Bernhard & Ware & Diez & handelnd & währen & Reiche \\
& Simons & Lucas & Abweg & unerfahren & Samuel & Berlin & Jägermeister & Geld & überlegen & Ohr \\
& Lauben & Kauf & 494. & 238 & Magdeburg & erfahren & Mslle & unternehmen & Hans & spielen \\
& Kirchhof & Hoyerswerda & \_ & Handelsmann & derohalben & aufnehmen & Antiochien & Baumeister & Rolle & Krieg \\
& Zipper & Valentin & verstohlen & verständig & 1746 & allhier & Abraham & all & Tür & Jude \\

    \bottomrule
  \end{tabular}
  }%
  \caption{Per-decade top 20 PMI-associated words (German) for $\langle$compound, constituent, inflection point$\rangle$ predicted as having the highest positive compositionality trend by top configuration in terms of correlation with (top/bottom 20 extreme) annotated Delta compositionality ratings: $\langle$\texttt{2nd-order}, \texttt{decades}, \texttt{prt}$\rangle$ (ranking in Table~\ref{tab:ch8:corr-extremes-de})}\label{tab:de-pmi-CORR-most-pos}
\end{table}
\end{landscape}

\begin{landscape}
\begin{table}
  \footnotesize

  \resizebox*{!}{0.8\textwidth}{%
  \begin{tabular}{cll|ll|ll|llll}
    Target & \multicolumn{2}{c}{$t_1$}  & \multicolumn{2}{c}{$t_2$} & \multicolumn{2}{c}{$t_3$} & \multicolumn{2}{c}{$t_4$} & \multicolumn{2}{c}{$t_5$} \\ \midrule

 & \multicolumn{2}{c}{1750}  & \multicolumn{2}{c}{1760}  & \multicolumn{2}{c}{1770} \\
Wechselbrief & 4636 & 656 & ausgestellt & Formul & Papiergeld & kaufen \\
1760 & Wechselbillet & Kurrentgeld & Tauschbrief & fünfhundert & Wechselhandel & zerreissen \\
mod/head & trassiert & Verhandlung & Habe. & Rudolph & Karolin & verkaufen \\
& akzeptiert & verjähren & Ausgeber & Foderung & Taschenuhr & Summe \\
& ausgestellt & Trassanten & verjährt & uneigentlich & sechshundert & A \\
& ausstellen & wechsel-gläubig & Beschihet & wechselrecht & mitgeben & bezahlen \\
& Ausgeber & tertia- & Sola & Unterpfand & Paß & Erfindung \\
& verjährt & saldieren & Reichstaler & vergüten & Verstorbene & entgegen \\
& Aval & ficta & niederschreiben & be- & Verkaufe & zweien \\
& Negotiant & exakt & October & ausstellen & Lord & tausend \\

\midrule

 & \multicolumn{2}{c}{1880}  & \multicolumn{2}{c}{1890}  & \multicolumn{2}{c}{1900} \\
Marktpreis & Depreciation & Mk. & regulierend & Produktionspreis & Guinee & 1900 \\
1900 & Konjunktur & desjenig & Blattel & oszillatorisch & 1779 & Juni \\
mod/head & herabgehen & Viertel & Selbigkeit & Steigen & abkaufen & Gewicht \\
& Zollsatz & Weizen & Marktwert & Oszillation & juristisch & Juli \\
& Sachwert & Quantität & oszillieren & uniform & 1848 & entwickeln \\
& wechsel- & Erzeugnis & Repulsion & Ergiebigkeit & anerkennt & sogar \\
& Preiswechsel & beispielsweise & Konkurrenzverhältnis & Bodenprodukt & angeben & Begriff \\
& Transportkosten & abnehmen & Extraprofit & Zuschußkapital & 1901 & Wert \\
& Landweg & entschließen & Verkaufsgeschäft & Durchschnittszahl & Verfügung & schließlich \\
& Durchschnittspreis & Berechnung & Bebauer & Einkaufspreis & 7. & 20 \\

\midrule

 & \multicolumn{2}{c}{1880}  & \multicolumn{2}{c}{1890}  & \multicolumn{2}{c}{1900}  & \multicolumn{2}{c}{1910} \\
Gasthaus & Yadoya & Kramladen & Schöll & Tribuswinkel & Kerschbaum & Kegelbahn & Holansky & Tscherne \\
1900 & schmidt'schen & Bauerngut & Grausam & Kreuzgasse & Manner-Gesang-Verein & Frühlingsliedertafel & mader & Pukl \\
mod & zusammenrotten & Badegeld & Stubbenkammer & Kottingbrunn & Hausball & Drobnitsch & kippen & Meichenitsch \\
& Theehäuser & Vincenz & Tanzunterhaltung & Hussowitz & Liederkranz & Blaschitz & asinnig & Jahresversammlung \\
& Pus & Neustift & Breitenseerstraße & Hausball & Wehrhonig & Linde & Wodenig & Gittertor \\
& Lusthaus & Müggelbude & Wassergasse & Theehäuser & Verbandsabend & Kottingbrunn & Vorlegeplatte & Fiden \\
& Bockenheim & Krawall & Eder & Teesdorf & Tombola & Sooß & Brudermann & Custer \\
& angerufen & makt & rim & Feuerwehrkapelle & Steueramtsoffizial & Police & Bahnbau & Roßwein \\
& Osterode & freiw. & Werkeltag & Bauernball & Laschitsch & Neuhold & 1/411 & Unwesen \\
& Michaeli & dekorieren & Vinicssay & überschattet & Kumeric & Lettemüller & zechen & Taverne \\

\midrule

 & \multicolumn{2}{c}{1860}  & \multicolumn{2}{c}{1870}  & \multicolumn{2}{c}{1880}  & \multicolumn{2}{c}{1890} \\
Kartenspiel & Schippenkönigin & hadern & würfel- & Coeur & Fürstenkongreß & Schwören & verklopfen & verflucht \\
1880 & Aide & Einsatz & Verlierende & verlierend & Wachmannschaft & Ausspielen & Lottospiel & Jubiläum \\
head & würfel- & Hader & trumpfen & nummerieren & Gewinner & Karo & verjubeln & Wirtshaus \\
& hochfahren & Macher & kart' & Sincerire & Dreiblatt & Collin & deklamatorisch & Sold \\
& wahrsagend & Compagnon & billard- & Hundebett & Blutzer & Zugeben & Schachspiel & leichtsinnig \\
& bret- & Macker & Seichen & 3629 & *992 & 1362 & vorurteilsfrei & Erwachsene \\
& Falschspieler & überbauen & Saunickeln & 3326 & Hausmittel & verrecken & Louisabeth & sitzend \\
& abmahnen & aufputzen & Kurrhahn & 3149 & 0,30 & Angreifer & Ehrengericht & wahren \\
& Hadern & Fiedel & 4578 & 2041 & Bekennen & steiermark. & Jour & schrecklich \\
& Chaise & Geplauder & *895 & *769 & *198 & Niederlausitz & Trumpf & Franken \\

    \bottomrule
  \end{tabular}
  }%
  \caption{Per-decade top 20 PMI-associated words (German) for $\langle$compound, constituent, inflection point$\rangle$ predicted as having the highest negative compositionality trend by top configuration in terms of correlation with (top/bottom 20 extreme) annotated Delta compositionality ratings: $\langle$\texttt{2nd-order}, \texttt{decades}, \texttt{prt}$\rangle$ (ranking in Table~\ref{tab:ch8:corr-extremes-de})}\label{tab:de-pmi-CORR-most-neg}
\end{table}
\end{landscape}

\begin{landscape}
\begin{table}
  \footnotesize

  \resizebox*{!}{0.9\textwidth}{%
  \begin{tabular}{cll|ll|ll|ll|ll}
    Target & \multicolumn{2}{c}{$t_1$}  & \multicolumn{2}{c}{$t_2$} & \multicolumn{2}{c}{$t_3$} & \multicolumn{2}{c}{$t_4$} & \multicolumn{2}{c}{$t_5$} \\ \midrule

 & \multicolumn{2}{c}{1810}  & \multicolumn{2}{c}{1820}  & \multicolumn{2}{c}{1830}  & \multicolumn{2}{c}{1840}  & \multicolumn{2}{c}{1850} \\
leisure hour & assiduously & employ & sisterly & accomplishment & oui & dreamy & tedium & instructive & law-office & amuse \\
1830 & winter & study & forty-eight & yourselves & nnd & amuse & war-whoop & humbly & caligula & engrave \\
head & intellectual & desire & melville & theme & beguiling & squirrel & thou'lt & amusing & belles-lettres & carrington \\
& cultivate & fill & bethink & amusement & poesy & recreation & beguile & apprentice & nnd & devote \\
& spend & attend & aura & spend & sundays & bis & cadet & employer & mclane & purchaser \\
& disguise & purpose & revel & leisure & beguile & toy & relaxation & occupation & foreman & blacksmith \\
& flower & fire & industrious & appropriate & divided & lounge & sometime & deficiency & anticipated & reading \\
& profession & thus & mature & employ & profitably & nut & devote & amuse & relaxation & geological \\
& evening & while & amuse & ye & pore & ruthven & recreation & elsewhere & spend & hideous \\
& sex & up & snatch & fit & sydney & anaxagoras & twill & belt & chart & recreation \\

\midrule

 & \multicolumn{2}{c}{1810}  & \multicolumn{2}{c}{1820}  & \multicolumn{2}{c}{1830}  & \multicolumn{2}{c}{1840} \\
country girl & hogg & allow & well-dressed & rude & phebe & credulity & rosy-cheeked & elinor \\
1820 & elope & i.- & unperverted & original & nic & construe & ungenerous & clementina \\
head & milk & <temp> & dimpled & attend & tidy & snug & chat & yonder \\
& black & feeling & housekeeper & town & peggy & clara & silly & satisfaction \\
& charming & eye & rosy & condition & batch & apparition & rustic & blind \\
& text & wife & bespeak & sit & unpractised & lawn & breed & lean \\
& despise & mrs.- & breed & suppose & romp & merton & confident & frequently \\
& main & act & group & feeling & lusty & fisher & cunning & victim \\
& contain & little & unworthy & history & enamoured & meek & supreme & born \\
& humble & with & paint & open & minion & seventeen & ignorant & edward \\

\midrule

 & \multicolumn{2}{c}{1970}  & \multicolumn{2}{c}{1980}  & \multicolumn{2}{c}{1990}  & \multicolumn{2}{c}{2000} \\
baby boy & rorvik & messenger & eight-pound & promising & seidel & doing & wallwho & septimus \\
1990 & lucious & aging & langdon & satisfied & yeh & tyrant & ofthem & senhor \\
mod & six-foot & hampton & cesarean & splendid & arie & dimple & langstrom & abhay \\
& kay & tactic & rectory & infant & wernick & lucinda & mpe & cuisinart \\
& sourly & curb & chubby & fat & beady & \# & priebes & pram \\
& eccentric & infant & cox & lung & circumcision & carole & backus & abacus \\
& dewey & foster & clemens & healthy & autistic & yoshiko & mejia & emmitt \\
& saloon & healthy & plump & naked & 1849 & befall & six-foot-two & foreskin \\
& adoption & adopt & hormone & float & meningitis & motoda & peete & raul \\
& smuggle & proud & inherit & passage & symptomatic & complexion & cross-fade & masterwork \\

\midrule

 & \multicolumn{2}{c}{1970}  & \multicolumn{2}{c}{1980}  & \multicolumn{2}{c}{1990}  & \multicolumn{2}{c}{2000} \\
leisure time & tivities & consumption & bellyaching & companion & salaryman & chopsticks & svendsen & retirees \\
1990 & underworked & unfold & -e & plenty & pre-1973 & leisure & kilgore & relaxation \\
head & enrichment & meaningful & assateague & reading & self-creation & indulgence & karachi & selective \\
& enjoyable & psychiatric & mcguire & heavily & redefinition & 2010 & bountiful & constraint \\
& campground & profitable & gnp & spend & self-improvement & vr & casolo & mobility \\
& intercom & prosperity & pimp & data & affluence & erosion & keg & ga \\
& vastly & dispose & athletic & village & sedentary & addiction & cai & spending \\
& relaxation & luxury & compensation & politics & overworked & expanding & time-consuming & selfish \\
& respectively & worthy & tan & activity & tiananmen & pursuit & monique & commute \\
& constructive & verse & augusta & value & discretionary & 50\% & equate & recreation \\

    \bottomrule
  \end{tabular}
  }%
   \caption{Per-decade top 20 PMI-associated words (English) for $\langle$compound, constituent, inflection point$\rangle$ with highest positive Delta compositionality per annotation (from Table~\ref{tab:en-anno-trends}).}\label{tab:en-pmi-anno-most-pos}
\end{table}
\end{landscape}

\begin{landscape}
\begin{table}
  \footnotesize

  \resizebox*{!}{0.9\textwidth}{%
  \begin{tabular}{cll|ll|ll|ll|ll}
    Target & \multicolumn{2}{c}{$t_1$}  & \multicolumn{2}{c}{$t_2$} & \multicolumn{2}{c}{$t_3$} & \multicolumn{2}{c}{$t_4$} & \multicolumn{2}{c}{$t_5$} \\ \midrule

 & \multicolumn{2}{c}{1810}  & \multicolumn{2}{c}{1820}  & \multicolumn{2}{c}{1830}  & \multicolumn{2}{c}{1840}  & \multicolumn{2}{c}{1850} \\
school day &   &   &   &   & cabaret & emerge & innate & close & hankering & holiday \\
1830 &   &   &   &   & crescent & joke & upward & even & ida & flora \\
head &   &   &   &   & 180 & 9 & amongst & end & systematically & bayard \\
&   &   &   &   & reminiscence & text & married & soon & inwardly & blount \\
&   &   &   &   & cathedral & childhood & praise & seem & vividly & strictly \\
&   &   &   &   & gul & slumber & assume & old & one-half & beloved \\
&   &   &   &   & gulberth & i.- & remove & over & reminiscence & agreeable \\
&   &   &   &   & healthful & chapter & equal & most & ann & proposition \\
&   &   &   &   & aloft & beloved & dress & before & entreaty & alike \\
&   &   &   &   & soar & discipline & during & their & interruption & empty \\

\midrule

 & \multicolumn{2}{c}{1850}  & \multicolumn{2}{c}{1860}  & \multicolumn{2}{c}{1870} \\
lady friend & dowager & reception & belcher & slightly & arsenical & berth \\
1850 & guillemot & abode & disparaging & charming & ossipee & miriam \\
mod/head & redburn & thanks & colchester & assistance & charlottesville & soothing \\
& statistics & clayton & mortification & hospital & apprehensively & quaker \\
& lilias & tie & rational & agreeable & ducat & cough \\
& foolish & favorite & dine & numerous & chum & startling \\
& velvet & explain & instruct & fashion & toleration & forrest \\
& cloak & inform & gratify & edward & boarding-house & testify \\
& monsieur & kiss & sensible & virginia & willoughby & orleans \\
& nod & introduce & elegant & enable & spruce & poison \\

\midrule

 & \multicolumn{2}{c}{1830}  & \multicolumn{2}{c}{1840}  & \multicolumn{2}{c}{1850}  & \multicolumn{2}{c}{1860} \\
iron door & unconscionable & safe & grim-looking & unlock & kiln & reappear & blab & fit \\
1840 & dilapidated & build & padlock & sunken & clashing & vase & xlvi & heavy \\
mod & cell & stone & tiled & massive & grated & accumulation & countless & free \\
& brick & bill & key-holes & fireman & ponderous & massive & sheriff & close \\
& necessity & sort & burial-ground & partition & steadfastly & abide & lock & till \\
& marble & establish & rusty & nero & dungeon & fiend & apartment & side \\
& indicate & except & overgrow & cleft & illumine & milton & behind & between \\
& <temp> & wall & guerne & hinge & rust & bolt & treasure & once \\
& finally & strong & ponderous & vault & rend & vault & bar & part \\
& aim & also & warehouse & clang & unlock & key & beat & they \\

\midrule

 & \multicolumn{2}{c}{1880}  & \multicolumn{2}{c}{1890}  & \multicolumn{2}{c}{1900}  & \multicolumn{2}{c}{1910}  & \multicolumn{2}{c}{1920} \\
baby boy & acquaintanceship & destine & beranger & silvery & cute & beside & bouncin & ** & sanitarium & jersey \\
1900 & annis & bath & adoptive & bravely & christen & shoulder & tomlins & lonesome & half-breed & 8 \\
mod & reprobate & moses & winnipeg & hebrew & parker & wife & oh-h & andy & suck & splendid \\
& blooming & jerrie & toddle & crow & stagger & send & two-year-old & conductor & switch & alive \\
& geoffrey & clasp & cute & walter & mamma & talk & christening & linger & carriage & couple \\
& holland & resume & chubby & violet & print & become & bright-eyed & widow & 1923 & marriage \\
& 49 & household & methuen & gloom & nice & home & powdered & household & weigh & baby \\
& cluster & slip & prattle & queen & lot & another & pore & beast & buddy & born \\
& cradle & circle & bethlehem & wherever & arthur & word & hen & breast & angel & claim \\
& designate & arrive & patter & embrace & sudden & again & lovin & truly & pleased & spend \\

    \bottomrule
  \end{tabular}
  }%
   \caption{Per-decade top 20 PMI-associated words (English) for $\langle$compound, constituent, inflection point$\rangle$ with highest negative Delta compositionality per annotation (from Table~\ref{tab:en-anno-trends}).}\label{tab:en-pmi-anno-most-neg}
\end{table}
\end{landscape}

\begin{landscape}
\begin{table}
  \footnotesize

  \resizebox*{!}{0.9\textwidth}{%
  \begin{tabular}{cll|ll|ll|ll|ll}
    Target & \multicolumn{2}{c}{$t_1$}  & \multicolumn{2}{c}{$t_2$} & \multicolumn{2}{c}{$t_3$} & \multicolumn{2}{c}{$t_4$} & \multicolumn{2}{c}{$t_5$} \\ \midrule

 & \multicolumn{2}{c}{1860}  & \multicolumn{2}{c}{1870}  & \multicolumn{2}{c}{1880}  & \multicolumn{2}{c}{1890}  & \multicolumn{2}{c}{1900} \\
frame house & sacramento & appearance & two-story & showy & door-yards & hopelessly & modest-looking & pontiac & two-story & adjoin \\
1880 & garrison & company & respectable-looking & unoccupied & 733 & decorate & land-seekers & bluff & twenty-first & quiver \\
mod/head & log & walk & bighorn & knoll & 731 & prohibit & gray-white & old-fashioned & rambling & wherever \\
& 9 & small & glaring & perry & three-story & solitary & nottingham & situate & northeastern & machinery \\
& addition & room & abortive & thereon & two-story & substantial & two-story & crazy & jailer & fairly \\
& cabin & also & smouldering & ornamental & cistern & olive & verandah & lace & a- & opposite \\
& tax & two & schuyler & roadside & roomy & fashionable & herman & edge & settin & main \\
& meaning & better & morse & unwonted & fatally & replace & sift & primitive & casey & window \\
& fort & stand & stockton & basement & stead & emerson & neat & erect & hallway & white \\
& inquire & here & greasy & shanty & neat & block & shanty & application & shutter & edge \\

\midrule

 & \multicolumn{2}{c}{1970}  & \multicolumn{2}{c}{1980}  & \multicolumn{2}{c}{1990}  & \multicolumn{2}{c}{2000} \\
school day & well-populated & anecdote & olmsted & recess & zombied & robards & 14.7 & frenetic \\
1990 & nurse-teacher & classroom & workday & deepen & pre-kindergarten & 1857 & beadle & undue \\
head & nakedly & unbearable & hacker & berry & bandy-legged & fad & heterogeneity & compulsory \\
& wittily & feverish & circa & annually & kipp & keller & findlay & midsection \\
& antedate & schoolteacher & parochial & 1950 & parent-teacher & chiefly & bari & recess \\
& clio & garland & p144 & closing & ad/hd & concord & naesp & innumerable \\
& glorify & 1939 & goode & strictly & rugby & idly & woodstove & marlon \\
& repetitive & shorten & tattered & learning & bray & dwight & first-person & bullying \\
& scribner & math & sneakers & sock & workday & rhyme & born-again & semblance \\
& defiant & stairway & diana & johnny & tiring & prep & \#4 & wilmington \\

\midrule

 & \multicolumn{2}{c}{1970}  & \multicolumn{2}{c}{1980}  & \multicolumn{2}{c}{1990}  & \multicolumn{2}{c}{2000} \\
leisure time & tivities & consumption & bellyaching & companion & salaryman & chopsticks & svendsen & retirees \\
1990 & underworked & unfold & -e & plenty & pre-1973 & leisure & kilgore & relaxation \\
mod & enrichment & meaningful & assateague & reading & self-creation & indulgence & karachi & selective \\
& enjoyable & psychiatric & mcguire & heavily & redefinition & 2010 & bountiful & constraint \\
& campground & profitable & gnp & spend & self-improvement & vr & casolo & mobility \\
& intercom & prosperity & pimp & data & affluence & erosion & keg & ga \\
& vastly & dispose & athletic & village & sedentary & addiction & cai & spending \\
& relaxation & luxury & compensation & politics & overworked & expanding & time-consuming & selfish \\
& respectively & worthy & tan & activity & tiananmen & pursuit & monique & commute \\
& constructive & verse & augusta & value & discretionary & 50\% & equate & recreation \\

\midrule

 & \multicolumn{2}{c}{1970}  & \multicolumn{2}{c}{1980}  & \multicolumn{2}{c}{1990}  & \multicolumn{2}{c}{2000} \\
baby boy & rorvik & messenger & eight-pound & promising & seidel & doing & wallwho & septimus \\
1990 & lucious & aging & langdon & satisfied & yeh & tyrant & ofthem & senhor \\
mod & six-foot & hampton & cesarean & splendid & arie & dimple & langstrom & abhay \\
& kay & tactic & rectory & infant & wernick & lucinda & mpe & cuisinart \\
& sourly & curb & chubby & fat & beady & \# & priebes & pram \\
& eccentric & infant & cox & lung & circumcision & carole & backus & abacus \\
& dewey & foster & clemens & healthy & autistic & yoshiko & mejia & emmitt \\
& saloon & healthy & plump & naked & 1849 & befall & six-foot-two & foreskin \\
& adoption & adopt & hormone & float & meningitis & motoda & peete & raul \\
& smuggle & proud & inherit & passage & symptomatic & complexion & cross-fade & masterwork \\

    \bottomrule
  \end{tabular}
  }%
\caption{Per-decade top 20 PMI-associated words (English) for $\langle$compound, constituent, inflection point$\rangle$ predicted as having the highest positive compositionality trend by top configuration in terms of $R^2$ score: $\langle$\texttt{bert}, \texttt{decades}, \texttt{apd}$\rangle$ (ranking in Table~\ref{tab:ch8:en-avg-lr-repr-schedule})}\label{tab:en-pmi-LR-most-pos}
\end{table}
\end{landscape}

\begin{landscape}
\begin{table}
  \footnotesize

  \resizebox*{!}{0.9\textwidth}{%
  \begin{tabular}{cll|ll|ll|ll|ll}
    Target & \multicolumn{2}{c}{$t_1$}  & \multicolumn{2}{c}{$t_2$} & \multicolumn{2}{c}{$t_3$} & \multicolumn{2}{c}{$t_4$} & \multicolumn{2}{c}{$t_5$} \\ \midrule

 & \multicolumn{2}{c}{1870}  & \multicolumn{2}{c}{1880}  & \multicolumn{2}{c}{1890}  & \multicolumn{2}{c}{1900}  & \multicolumn{2}{c}{1910} \\
rebel army & wagon-train & 19th & ter-night & grapple & i8th & apprehend & treasonable & ned & 1685 & destruction \\
1890 & fifty-nine & twenty-six & perryville & concentration & manassas & disperse & burgoyne & northern & monmouth & responsible \\
mod & gadsden & subsistence & murfreesboro & intimation & unconditional & hind & deserter & spite & monterey & wing \\
& undisciplined & saw & pearce & tennessee & 200,000 & mcclellan & unearth & proceed & nicaragua & broad \\
& golly & marietta & disorganized & cripple & re-cross & recruit & repute & presence & conquering & strength \\
& enlistment & 1863 & virtual & nashville & harass & potomac & willingness & affair & friendliness & shoulder \\
& civilian & alternative & straggler & johnston & hem & overwhelm & canada & demand & petersburg & catch \\
& enlist & joseph & provisional & front & rout & surrender & complain & wonder & advancing & rest \\
& forty-eight & aggregate & 30th & duly & victorious & richmond & hungry & toward & grove & show \\
& dishonorable & surgeon & encamp & rosy & peninsula & kentucky & plot & york & observer & good \\

\midrule

 & \multicolumn{2}{c}{1950}  & \multicolumn{2}{c}{1960}  & \multicolumn{2}{c}{1970}  & \multicolumn{2}{c}{1980}  & \multicolumn{2}{c}{1990} \\
family physician & internally & freely & unselfishly & old-time & massie & 10,000 & categorize & surgeon & rheumatologist & nj \\
1970 & ead & poll & torrence & staffs & lungren & prescribe & sorely & consult & board-certified & obstetrician \\
mod/head & depressing & panel & abortionist & perceptive & goldstein & residence & coroner & embarrassed & lozano & fresno \\
& 6,000 & harold & panelist & specialization & 5000 & penelope & elated & academy & noe & midwife \\
& subscriber & surgeon & gynecologist & mac & trusted & alert & uncertainly & courage & umpteenth & md \\
& d.c.-the & anxious & harried & cole & hays & thumb & akin & 19 & gillen & 14\% \\
& ascertain & foundation & humphreys & diagnosis & concur & physician & diagnose & volume & primary-care & 145 \\
& meek & tongue & downgrade & confide & detection & ralph & two-thirds & reader & internist & gallagher \\
& unnecessary & regular & shopper & 81 & phillips & possess & 64 & regular & fuhrman & exasperate \\
& nelson & ability & dwindling & dentist & frantically & considerable & accumulate & suggest & 1815 & nutritional \\

\midrule

 & \multicolumn{2}{c}{1870}  & \multicolumn{2}{c}{1880}  & \multicolumn{2}{c}{1890}  & \multicolumn{2}{c}{1900}  & \multicolumn{2}{c}{1910} \\
silk hat & fur-trimmed & ruffle & pollytishuns & decorum & long-tailed & broadcloth & fourth-of-july & frock & dandified & spat \\
1890 & jewellery & overcoat & neckcloth & glove & alta & leer & coattails & crape & gold-headed & regalia \\
head & ulster & beaver & watch-chain & patter & rectory & chaste & floresville & doff & frock-coats & patent-leather \\
& dens & array & adjieu & raven & frock-coat & derby & frock-coat & topeka & cutaway & longears \\
& tilt & outer & glossy & skeezicks & slick & necktie & five-pound & overcoat & incontinently & fur-lined \\
& gaily & unlike & shaven & clerical & deferential & unsteady & spat & hopper & gardenia & broadcloth \\
& conspirator & garment & gloved & albert & glossy & thet & starched & kindergarten & wide-brimmed & dapper \\
& railing & thrust & latterly & desmit & flowered & accidentally & rumpled & boarding-house & frock-coat & tall \\
& trouser & iii & broadcloth & grab & trouser & wear & pale-faced & immaculate & desperadoes & vintage \\
& jacket & straight & shiny & trouser & shining & satin & paddock & landlady & distressingly & clean-shaven \\

\midrule

 & \multicolumn{2}{c}{1850}  & \multicolumn{2}{c}{1860}  & \multicolumn{2}{c}{1870} \\
lady friend & dowager & reception & belcher & slightly & arsenical & berth \\
1850 & guillemot & abode & disparaging & charming & ossipee & miriam \\
mod & redburn & thanks & colchester & assistance & charlottesville & soothing \\
& statistics & clayton & mortification & hospital & apprehensively & quaker \\
& lilias & tie & rational & agreeable & ducat & cough \\
& foolish & favorite & dine & numerous & chum & startling \\
& velvet & explain & instruct & fashion & toleration & forrest \\
& cloak & inform & gratify & edward & boarding-house & testify \\
& monsieur & kiss & sensible & virginia & willoughby & orleans \\
& nod & introduce & elegant & enable & spruce & poison \\

    \bottomrule
  \end{tabular}
  }%
\caption{Per-decade top 20 PMI-associated words (English) for $\langle$compound, constituent, inflection point$\rangle$ predicted as having the highest negative compositionality trend by top configuration in terms of $R^2$ score: $\langle$\texttt{bert}, \texttt{decades}, \texttt{apd}$\rangle$ (ranking in Table~\ref{tab:ch8:en-avg-lr-repr-schedule})}\label{tab:en-pmi-LR-most-neg}
\end{table}
\end{landscape}